\documentclass{article}

\PassOptionsToPackage{numbers, compress}{natbib}

\usepackage[preprint]{neurips_2024}

\newcommand{\ours}{{Drantal}}
\usepackage[utf8]{inputenc} 
\usepackage[T1]{fontenc}    
\usepackage{hyperref}       
\usepackage{url}            
\usepackage{booktabs}       
\usepackage{amsfonts}       
\usepackage{nicefrac}       
\usepackage{microtype}      
\usepackage{xcolor}         
\usepackage{graphicx}
\usepackage{multirow}
\usepackage{amsmath}
\usepackage{wrapfig, lipsum, booktabs}
\usepackage{algorithm}  
\usepackage{algpseudocode}  
\usepackage{amsmath} 
\usepackage{wrapfig}
\title{Drantal-NeRF: Diffusion-Based Restoration for Anti-aliasing Neural Radiance Field}

\author{Ganlin Yang \textsuperscript{\rm 1}\thanks{This work was done when Ganlin Yang was an intern at Microsoft Research Asia, from August, 2023 to April, 2024.}\quad
    Kaidong Zhang \textsuperscript{\rm 1}\quad
    Jingjing Fu \textsuperscript{\rm 2 $\dag$}\quad
    Dong Liu \textsuperscript{\rm 1}\thanks{Corresponding authors: J. Fu and D. Liu.} \\
\textsuperscript{\rm 1} University of Science and Technology of China 
\textsuperscript{\rm 2} Microsoft Research Asia\\
{\tt\small ygl666@mail.ustc.edu.cn\quad
richu@mail.ustc.edu.cn
} \\
{\tt\small 
jifu@microsoft.com
\quad dongeliu@ustc.edu.cn
}
}

\begin{document}

\maketitle

\begin{abstract}
   Aliasing artifacts in renderings produced by Neural Radiance Field (NeRF) is a long-standing but complex issue in the field of 3D implicit representation, which arises from a multitude of intricate causes and was mitigated by designing more advanced but complex scene parameterization methods before. In this paper, we present a \textbf{D}iffusion-based \textbf{r}estoration method for \textbf{ant}i-\textbf{al}iasing \textbf{Ne}ural \textbf{R}adiance \textbf{F}ield
  (\textbf{Drantal-NeRF}). We
  consider the anti-aliasing issue from a low-level restoration perspective by viewing aliasing artifacts as a kind of degradation model added to clean ground truths. By leveraging the powerful prior knowledge encapsulated in diffusion model, we could restore the high-realism anti-aliasing renderings conditioned on aliased low-quality counterparts. We further employ a feature-wrapping operation to ensure multi-view restoration consistency and finetune the VAE decoder to better adapt to the scene-specific data distribution. 
  Our proposed method is easy to implement and agnostic to various NeRF backbones. We conduct extensive experiments on challenging large-scale urban scenes as well as unbounded 360-degree scenes and achieve substantial 
 qualitative and quantitative improvements. 
\end{abstract}

\section{Introduction}
\label{sec:intro}
Neural Radiance Field (NeRF)~\cite{nerf} has been demonstrated to be a powerful tool for a wide range of 3D vision tasks, such as novel view synthesis~\cite{nerf, mipnerf, pixelnerf, instantngp, tensorf}, 3D explicit modeling~\cite{pointnerf, nerf2mesh}, 3D content generation~\cite{zero123, dreamfusion, magic3d, makeit3d, prolificdreamer}, robotics~\cite{nerfrobotic}, computational photography~\cite{nerfinthedark} and so on. Recently, to accelerate the training and rendering process for NeRF, voxel-grid based NeRFs~\cite{tensorf, instantngp, nsvf, dvgo, zipnerf, plenoxels} have been dominated compared with previous purely implicit MLP-based NeRFs for faster convergence as well as the advantages of explicit scene modeling.

Although NeRF exhibits impressive performance in the novel-view synthesis for small-scale real-world scenes, such as object-centric and forward-facing scenes, it still suffers from aliasing artifacts and blurry renderings, especially when applied for 1) Unbounded 360-degree scenes, where the camera may point in any direction and the content may exist at any remote distance, and 2) Large-scale complex scenes, such as city blocks which may extend over a scale of square kilometers. We attribute the occurrence of these aliasing rendering artifacts to the following factors:
\begin{enumerate}
    \item \textbf{Limited Representation Capacity.} The learnable parameters for a 3D scene, whether they are implicit MLPs or explicit voxel girds, are inherently limited. Specifically, the memory usage increases cubically with voxel-grid resolutions, making it impractical to increase voxel numbers to a very large value. When utilized for a complicated large-scale scene abundant with intricate details, the representation capacity saturates and therefore it produces blurs at fine details.

    \item \textbf{Inappropriate Scene Parameterization.} For instance, an infinitesimal sampled 3D point along the ray cannot reason about the scale information of the scene. To address this, MipNeRF~\cite{mipnerf} casts conical frustums instead of rays and encodes scale information into integrated positional encoding (IPE). Furthermore, unbounded scenes necessitate a well-designed parameterization trick for sampling strategy, aiming to balance the nearby and remote objects in the scene. MipNeRF-360~\cite{miperf360} and ZipNeRF~\cite{zipnerf} have proposed specially-tailored parameterization methods for 360-degree unbounded scenes.

    \item \textbf{Inherent Ill-posed Optimization.} Certain regions of the large-scale scene may only be observed within a limited batch of rays in training sets, thereby lacking robust multi-view supervision. Previous works mitigate this issue by incorporating regularization terms that promote surface smoothness~\cite{nerfactor, unisurf} and discourage semi-transparent floaters~\cite{freenerf}. 
\end{enumerate}

Although previous studies have offered several solutions to the aforementioned aliasing issue, they not only tend to be mathematically sophisticated and challenging to implement, but also have not entirely addressed the issue. The renderings from these methods still suffer from aliasing artifacts, such as blurs and unclear details in some challenging scene regions. This prompts us to consider the anti-aliasing issue from a fresh perspective: \textbf{The aliasing rendering images can be viewed as a specific degradation model added to clear images. Although this degradation form is difficult to express precisely in mathematical terms, it adheres to specific probabilistic distributions, allowing it to be implicitly learned by neural networks.}

We draw inspirations from blind image restoration (BIR) task in low-level vision, whose ultimate objective is to reconstruct a high-quality image from its low-quality counterpart, taking into account general degradation distributions. Recently, diffusion models~\cite{ddim, ddpm, stablediff} have demonstrated significant advancements in generating high-quality images with fine details~\cite{stablediff, controlnet, t2iadapter}, and have been proposed as solutions to BIR task~\cite{diffbir, stablesr, pasd}. We consider that the rich prior knowledge encapsulated in the large-scale pretrained diffusion model~\cite{stablediff}, could be extremely beneficial for anti-aliasing Neural Radiance Field renderings due to its exceptional ability to generate high-definition texture details. Given this, we propose a \textbf{D}iffusion-based \textbf{r}estoration method for \textbf{ant}i-\textbf{al}iasing \textbf{Ne}ural \textbf{R}adiance \textbf{F}ield (\textbf{Drantal-NeRF}). Specifically, we design a two-stage training paradigm for our Drantal-NeRF. In the first stage, we finetune a pretrained diffusion model while training Neural Radiance Field. We treat the rendering patches produced by NeRF and their corresponding ground truth patches as the low-quality and high-quality pairs. Diffusion model is trained to generate anti-aliased high-quality renderings \textit{conditioned on low-quality inputs}. Inspired by CodeFormer~\cite{codeformer}, to further suppress randomness and enhance the fidelity of the generated images (\textit{i.e.}, to avoid generating content that does not exist in real captured scenes), in the second stage we incorporate an additional network module to fuse features from low-quality inputs and high-quality diffusion model outputs, which is optimized in an adversarial training manner. After the two-stage training process, the aliased low-quality renderings by NeRF can be enhanced \textit{by a large margin} both qualitatively and quantitatively, as measured by distortion metric PSNR, structural metric SSIM and perceptual quality metric LPIPS. Our contributions can be summarized as follows:
\begin{itemize}
    \item We reveal that the aliasing artifacts issue appeared in NeRF, although resulting from various complex factors, could be alleviated to a great extent from a simple yet novel image restoration perspective. We further exploit the strong generation capacity encapsulated in pretrained diffusion model as the prior for the removal of aliasing artifacts and deblurring.
    \item Our proposed method is NeRF-agnostic and could be applied to various kinds of NeRFs with multiple degradation types. We implement our proposed methodology on two strong baselines GridNeRF~\cite{gridnerf} and ZipNeRF~\cite{zipnerf}, for large-scale complex urban scenes and unbounded 360-degree scenes, respectively, and achieve state-of-the-art performance. We also conduct extensive ablation studies to validate the efficacy of each component and the tendency of performance gains with the voxel resolutions.
\end{itemize}

\section{Related Work}

\subsection{Anti-aliasing Neural Radiance Field}
The aliasing artifact appeared in Neural Radiance Field means that the renderings do not accurately capture the detailed information of the scene. To address this issue, MipNeRF~\cite{mipnerf} proposes to sample and render the conical frustums for continuous scale rendering. MipNeRF-360~\cite{miperf360} further presents a series of regularization methods to address the aliasing problem in unbounded scenes. ZipNeRF~\cite{zipnerf} introduces the frustum hash encoding and improves the proposal MLP for anti-aliasing in voxel grid-based NeRFs. Currently, there are also some works that formulate NeRF super-resolution task to enhance the rendering quality of NeRFs. They devise explicit super-sampling strategy~\cite{wang2022nerf} or design guidance network~\cite{huang2023refsr,han2023super,yoon2023cross} for rendering quality improvement. What's more, some other works~\cite{Lee_2024_WACV,dai2023hybrid,lee2023dp,lee2023exblurf,wang2023bad} propose to deblur NeRF, which can also be regarded as a feasible routine to address aliasing artifacts. These methods introduce sharpness prior~\cite{Lee_2024_WACV}, explicit neural point cloud~\cite{dai2023hybrid}, 6-DOF motion blur formation~\cite{lee2023dp}, physical scene priors~\cite{lee2023exblurf} or bundle adjustment~\cite{wang2023bad}. Different from previous works, our method firstly exploits the pretrained diffusion knowledge for anti-aliasing NeRF. Without formulating the degradation type heuristically, we adopt the diffusion prior as the general solution to address various kinds of degradations occurred in NeRF.

\subsection{Generative Model for Blind Image Restoration}
Generative models have emerged as a powerful tool in the field of blind image restoration(BIR), leveraging their ability to learn the complex distribution of image data and generate high-quality(HQ) restorations from low-quality(LQ) inputs, especially for the real-world degraded images with complex and unknown degradation. Most of existing real-world image super-resolution(SR) methods~\cite{ji2020real,ledig2017photo,liang2021swinir} are based on Generative Adversarial Networks(GANs)~\cite{goodfellow2014generative}. BSRGAN~\cite{zhang2021designing} and Real-ESRGAN~\cite{wang2021real} introduce innovative strategies for emulating real-world degradations. These methods are capable of eliminating most degradations, but they cannot generate realistic details due to limited generation capabilities. 
In response, recent studies extensively explore the generative facial priors from pretrained generators for blind face restoration~\cite{yang2021gan,wang2021gfpgan,chan2021glean}.
State-of-the-art studies~\cite{codeformer,wang2022restoreformer,gu2022vqfr} introduce the HQ codebook to generate astonishingly realistic face details. In a departure from explicit loss function design, diffusion-based image restoration methods~\cite{diffbir, stablesr} leverage prior knowledge encapsulated in pretrained text-to-image diffusion models~\cite{stablediff} for real-world image super-resolution. The adaptation of rich diffusion prior further enhances image restoration with sharp and realistic details. In this paper we treat the aliasing artifacts in NeRF as a kind of real-world degradation and exploit the potential of diffusion model in removing such artifacts.

\subsection{Diffusion Model Meets Neural Radiance Field}
Nowadays, a proliferation of works capitalize on the powerful photo-realistic image prior from diffusion models to empower the applications of Neural Radiance Fields. A series of works focus on lifting the 2D diffusion prior into 3D content generation, including text-to-3D generation~\cite{dreamfusion, magic3d, prolificdreamer, instant3d} and image-to-3D generation~\cite{zero123, zeronvs, wonder3d, makeit3d}.
As a seminal work, DreamFusion~\cite{dreamfusion} exploits the knowledge from the diffusion model to regularize the scene content of a NeRF with the novel Score Distillation Sampling (SDS) loss. Following DreamFusion, Magic3D~\cite{magic3d}, Dreambooth3D~\cite{dreambooth3d} and Wonder3D~\cite{wonder3d} extend the application to mesh-based 3D content generation, 3D generation content editing and image-to-3D generation. 
Another series of works~\cite{deng2023nerdi,ssdnerf,liu2023deceptive,gu2023nerfdiff, realfusion, wu2023reconfusion} utilize the diffusion models to enrich the quality of NeRFs under the few-shot sparse or even single view setting. 
Deceptive-NeRF~\cite{liu2023deceptive} and ReconFusion~\cite{wu2023reconfusion} both adopt the photo-realistic image generation capabilities from diffusion models to generate the high-realism pseudo-labels under the novel-sampled viewpoints, which are used to regularize NeRF's training under the few-shot observation views.
Different from the previous works, our proposed method does not concentrate on the text or image to 3D content generation or the few-shot novel-view synthesis task.
Instead, we target on exploiting the rich prior knowledge within the diffusion model in solving the anti-aliasing issue for NeRF's renderings in a unified approach.

\section{Methods}
\label{sec:methodoverall}
\begin{figure}[t]
    \centering
    \includegraphics[width=0.96\textwidth]{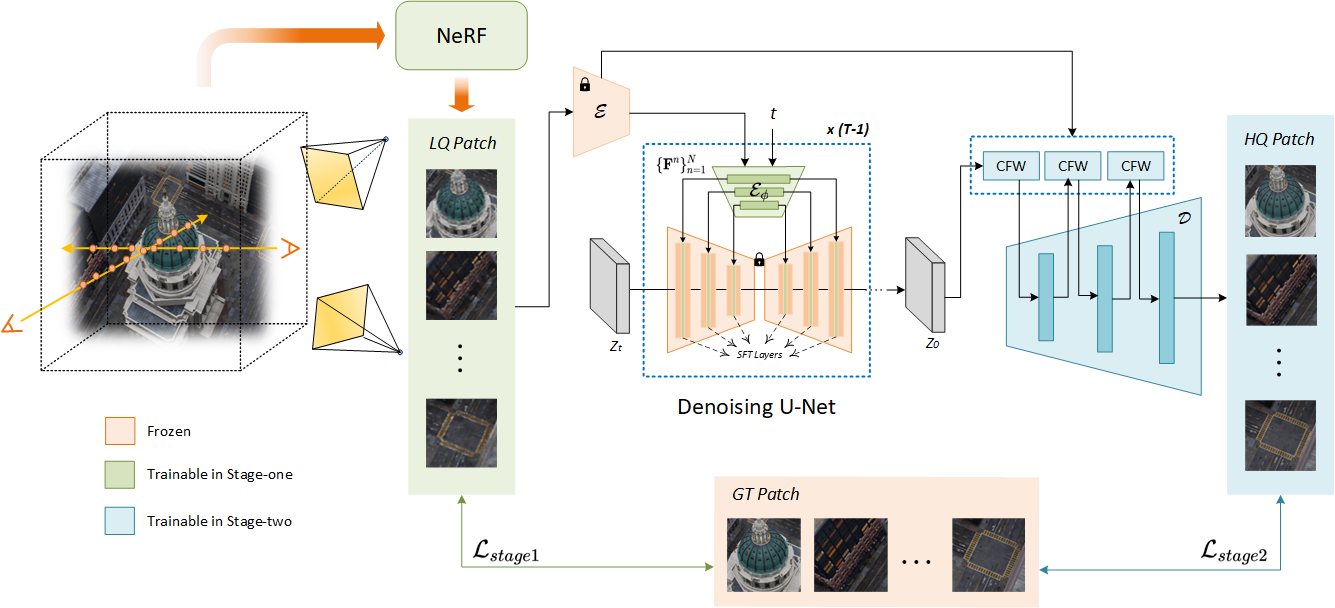}
    \caption{
   \textbf{Overview of our proposed method \ours, which involves two-stage training.} At the first stage, we optimize a Neural Radiance Field along with the time-aware encoder $\mathcal{E}_\phi$ and SFT layers in the diffusion model, marked as green. At the second stage, we optimize the controllable feature wrapping (CFW) module as well as the VAE decoder $\mathcal{D}$, distinguished in blue. The VAE encoder $\mathcal{E}$ and the rest of the parameters in the diffusion model are kept fixed, depicted in orange. After the whole training process, we are able to restore the high-quality anti-aliasing renderings from the degraded ones output by Neural Radiance Field.}
    \label{fig:pipeline}
\end{figure}

As shown in Figure \ref{fig:pipeline}, our NeRF-agnostic anti-aliasing method follows a two-stage training paradigm. In Section \ref{sec3.1} we describe the process of finetuning a Stable Diffusion model in a parameter-efficient manner simultaneously as we train Neural Radiance Field. To further enhance the generation fidelity, we employ a controllable feature wrapping module following CodeFormer~\cite{codeformer}, as elaborated in Section \ref{sec3.2}. We present the implementation details in Section \ref{sec3.3}.

\subsection{Aliasing-conditioned Diffusion Model Finetuning}
\label{sec3.1}
We firstly train a Neural Radiance Field as normal from multi-view images, supervised by a pixel-wise rendering loss:
\begin{equation}
\begin{split}
    \mathcal{L}_{nerf} =  \Vert \mathbf{C}(r)- \mathbf{C}^*(r) \Vert_2^2
    + \mathcal{L}_{reg}
\end{split}
\label{equation1}
\end{equation}
where $\mathbf{C}(r)$ and $\mathbf{C}^*(r)$ denote the rendered pixel value and the ground truth for the camera ray $r$ respectively. $\mathcal{L}_{reg}$ encompasses regularization terms such as the total variation loss~\cite{plenoxels}, $\mathcal{L}_1$ sparsity loss~\cite{tensorf}, anti-aliased interlevel loss~\cite{zipnerf} and so on, depending on different NeRF backbones. We leave more details to the Appendix \ref{additional training details}.

As we train the NeRF itself, it naturally produces amounts of low-quality aliased rendering patches $x_{lq}$ and corresponding high-quality ground truth patches $x_{hq}$, which could be transformed to the latent representation $z_{lq}$ and $z_{hq}$ easily using the VAE encoder $\mathcal{E}$. \textit{Our training objective is to learn a generation process restoring an anti-aliased high-quality image conditioned on the degraded aliased inputs.} The key is how to introduce the low-quality condition appropriately, preserving the high-frequency details from inputs while still not disrupting the generative prior within the original diffusion model. 
Following StableSR~\cite{stablesr}, we finetune the diffusion model in a parameter-efficient manner. The pretrained Stable Diffusion weights are frozen and we adopt an additional trainable time-aware encoder $\mathcal{E}_\phi$ to extract a multi-scale feature map $\{\mathbf{F}^n\}_{n=1}^N$ from the aliased degraded rendering patches. The additional time-aware encoder adopts the similar structure as the first half part of the Stable Diffusion U-Net for better intermediate features alignment. Given the intermediate feature maps output by original Stable Diffusion $\{\mathbf{F}_{dif}^n\}_{n=1}^N$, we use trainable spatial feature transformation (SFT) layers~\cite{sftlayer} for integrating degraded low-quality information into the diffusion model as the condition:
\begin{equation}
    \hat{\mathbf{F}}_{dif}^n = (1+\mathbf{\alpha}^n) \odot \mathbf{F}_{dif}^n + \mathbf{\beta}^n
    \qquad \qquad
    \mathbf{\alpha}^n, \mathbf{\beta}^n = \mathcal{M}_{\theta}^n(\mathbf{F}^n)
\end{equation}
Where $\mathcal{M}_{\theta}^n$ are learnable SFT layers to predict the affine parameters $\mathbf{\alpha}^n$ and $\mathbf{\beta}^n$, and $n$ denotes the scale indices in Stable Diffusion U-Net. The time-aware encoder and SFT layers are trained by approximating noise $\epsilon$ from the network prediction $\epsilon_\theta(z_t, t, z_{lq})$:
\begin{equation}
    \mathcal{L}_{diff} (\mathcal{E}_\phi, \mathcal{M}_{\theta}^n) = \mathbb{E}_{z,t,\epsilon} \Vert \epsilon-\epsilon_\theta(z_t, t, z_{lq}) \Vert _2^2
\end{equation}
The overall training loss in stage-one can be expressed in Equation \ref{loss1} while $\lambda$ is the coefficient to balance the two loss terms:
\begin{equation}
    \label{loss1}
    \mathcal{L}_{stage1} = \mathcal{L}_{nerf} + \lambda \mathcal{L}_{diff}
\end{equation}

An alternative choice is to separate the stage-one training into two cascaded processes: training the Neural Radiance Field alone at first, then generate a group of low-high quality pairs from the evaluation renderings and the ground truths, which are used to finetune the diffusion model subsequently. In practice we find that optimizing NeRF and the diffusion model jointly in stage-one not only streamlines the training process to be more elegant, but also provides a more diversified set of degradation types together with NeRF's training procedure, which endows the diffusion model with more robust and generalizable anti-aliasing capacity.

\subsection{Feature Wrapping for Improving Fidelity}
\label{sec3.2}
After stage-one training, our proposed method is already able to produce visually-compelling anti-aliasing renderings. However, since all the image renderings are enhanced individually under given camera viewpoints, it lacks an explicit mechanism to ensure multi-view consistent anti-aliasing enhancement. We seek the solutions from a fidelity-realism trade-off perspective. \textit{The low-quality NeRF's renderings, though degraded and aliased, are multi-view consistent inherently.} In order to assure better 3D consistency, the generated anti-aliased images should be more of fidelity to the low-quality inputs, suppressing the probability of generating multi-view inconsistent contents. 

We adopt the controllable feature wrapping (CFW) module inspired by CodeFormer~\cite{codeformer} and StableSR~\cite{stablesr}. As illustrated in Figure \ref{fig:pipeline}, the CFW module is learned to modulate the intermediate features in VAE decoder from the features output by corresponding VAE encoder layers. Let $\mathbf{F}_e$ and $\mathbf{F}_d$ be the original encoder and decoder features, the modulated decoder feature $\hat{\mathbf{F}}_d$ is controlled by an adjustable coefficient $w\in [0,1]$: 
\begin{equation}
    \hat{\mathbf{F}}_d = \mathbf{F}_d + \mathcal{C}(\mathbf{F}_e, \mathbf{F}_d; \Theta) \times w
\end{equation}
where $\mathcal{C}(\cdot; \Theta)$ is the learnable feature fusion module with convolution and RRDB~\cite{esrgan} layers. We leave the detailed structure of the CFW module to the Appendix \ref{cfw-structure}. During stage-two training, different from StableSR~\cite{stablesr} which only optimizes $\Theta$ and keeps all the other networks frozen, we additionally finetune the whole VAE decoder $\mathcal{D}$ to better fit the aliasing degradation distribution for a specific scene. We further adopt a patch-based discriminator $\mathcal{D}_{\Phi}$~\cite{isola2017image} (we omit in Figure \ref{fig:pipeline} for brevity) and optimize in an adversarial manner:
\begin{equation}
\begin{split}
    \Theta^{\star}, \mathcal{D}^{\star}, \mathcal{D}_{\Phi}^{\star} & = \arg \underset{\Theta, \mathcal{D}}{\min}
    \underset{\mathcal{D}_{\Phi}}{\max} \,
    \mathcal{L}_{stage2} \\
    & = \arg \underset{\Theta, \mathcal{D}}{\min}
    \underset{\mathcal{D}_{\Phi}}{\max} \,
    \Vert x - \hat{x} \Vert + \mathcal{L}_{perp}(x,\hat{x}) + \mathcal{L}_{GAN}(x, \hat{x})
    \end{split}
\end{equation}
Where $\hat{x}$ is the restored anti-aliased image patch output by $\mathcal{D}$ and 
$x$ is the ground truth. $\mathcal{L}_{perp}$ is the LPIPS~\cite{lpips} perceptual loss and $\mathcal{L}_{GAN} = \log \mathcal{D}_{\Phi}(x) + \log (1-\mathcal{D}_{\Phi}(\hat{x}))$. Such an adversarial training manner in stage-two further encourages the distribution of the anti-aliasing enhanced images to approximate the real distribution of the ground truth images, which significantly improves the fidelity of reconstruction quality thus promotes better multi-view consistency.

\subsection{Implementation Details}
\label{sec3.3}
In this section, we provide some key implementation details regarding our proposed method. More minor details could be found in the Appendix \ref{appendix: more imple details}.

\noindent \textbf{Color Correction.}
As noted by~\cite{choi2022perception}, diffusion models could occasionally cause color shifts compared with the original image, which hinders the multi-view illumination consistency in the Neural Radiance Field. We perform a color correction procedure after diffusion-enhancement to align the mean and variance between the aliased low-quality image $x_{lq}$ and the enhanced image $x_{enh}$ by the diffusion model:
\begin{equation}
    x_{enh}^c = \frac{x_{enh}-\mu_{x_{enh}}}{\sigma_{x_{enh}}} \cdot \sigma_{x_{lq}} + \mu_{x_{lq}}
\end{equation}
We use the color corrected image $x_{enh}^c$ for final evaluation.

\noindent \textbf{Altering Sampling Strategy.}
In most cases, as we train a NeRF, the cast rays are randomly sampled from all the pixels for all the training views. However, the diffusion model requires entire patches as inputs rather than independent pixels. What's more, the low-quality patches, if taken at the beginning of NeRF's optimization, are under too poor structure and semantic quality to guide the conditional generation process for diffusion model. To accommodate to the above issues, we alter the sampling strategy during stage-one training. We firstly cast rays at pixel-level as before and train a \textit{coarse} Neural Radiance Field to produce relatively acceptable low-quality renderings, then we alter the ray-sampling to patch-wise, continue NeRF's training and begin optimizing the diffusion model. 
For more details, please refer to Appendix \ref{additional training details}.


\noindent \textbf{Patch Aggregation Reverse Sampling.}
As mentioned above, the diffusion model is trained under resolution $512 \times 512$, however at inference time the overall aliased low-quality image $x_{lq}$ may exhibit an arbitrary resolution $h \times w$, probably larger than 512. This makes the latent feature map $z_{lq} \in \mathcal{R}^{(h/8) \times (w/8)}$ output by the VAE encoder larger than the resolution of $64 \times 64$, which is the preset resolution for the latent diffusion model~\cite{stablediff}. Inspired by~\cite{jimenez2023mixture, stablesr}, we use the patch aggregation operation at the reverse sampling stage, following the divide-and-conquer principle. During each timestep in the reverse sampling, $z_{lq}^{(t)}$ is split into overlapping $64 \times 64$ patches, processed by the diffusion model individually, then aggregated via a weighted sum by a Gaussian filter to produce $z_{lq}^{(t-1)}$. The pseudocode for this process is located in the Appendix \ref{patch reverse sampling pseducode}.

\noindent \textbf{Additional Details.}
During the stage-one training,
we train GridNeRF~\cite{gridnerf} with batch size 8192 for total 100k iterations. The voxel resolution increases from $128^3$ to $300^3$ with the training process. We sample rays at pixel-level for the first 50k iterations and at patch-level for the last 50k iterations. Following GridNeRF~\cite{gridnerf}, we optimize the grid branch at the beginning and add the NeRF branch at 40k iterations.
The learning rates for grid branch, NeRF branch and the diffusion branch are set to 0.02, $5e-4$ and $5e-5$, respectively. 
We train ZipNeRF~\cite{zipnerf} based on a pytorch-version reimplementation\footnote{https://github.com/SuLvXiangXin/zipnerf-pytorch} with batch size 65536 for total 70k iterations, where the voxel grid is implemented based on Instant-NGP~\cite{instantngp} with 10 scales. 
We sample rays at pixel-level for the first 50k iterations and at patch-level for the last 20k iterations.
The learning rate decays from 0.01 to 0.001 for NeRF branch and retains $5e-5$ for diffusion branch. $\lambda$ is set to 1 unless otherwise stated.
During the stage-two training we train the CFW module and finetune the VAE decoder together with learning rate $5e-5$, and train for 20k and 25k iterations for GridNeRF and ZipNeRF respectively. At inference time we use DDIM sampler~\cite{ddim} with 20 reversing sampling steps. Controllable adjustable coefficient $w$ is set to 1 at both training and inference time. More details could be found in the Appendix \ref{additional training details}.

\begin{table*}[t]

\begin{minipage}{0.47\textwidth}
\centering
\caption{Quantitative results for aerial scenes in MatrixCity~\cite{matrixcity} Dataset. We take average over five city block scenes (\textit{Block\_A} $\sim$ \textit{Block\_E}). Best in \textbf{bold}.
}
\vspace{+2mm}
\setlength{\tabcolsep}{8pt}
\scriptsize
\begin{tabular}{@{}lccc@{}}

\toprule 

 & PSNR$\uparrow$ & SSIM$\uparrow$ & LPIPS $\downarrow$ \\ 
\midrule
NeRF\cite{nerf} & 22.58	& 0.591	& 0.571 \\
BungeeNeRF\cite{bungeenerf} &23.56	& 0.686	& 0.500\\
MegaNeRF\cite{meganerf} & 23.58	& 0.689	& 0.499\\
SwitchNeRF\cite{switchnerf} & 24.12	& 0.715	& 0.461 \\
TensoRF\cite{tensorf} & 24.37 & 0.724 & 0.436 \\ 
GridNeRF\cite{zipnerf} & 24.88 & 0.749	& 0.385 \\
GridNeRF+NeRFLiX\cite{nerflix} & 25.01	& 0.767	& 0.365 \\
GridNeRF+\ours$_{gen}$ & 25.61 & 0.787 & 0.221 \\
GridNeRF+\ours$_{spe}$ & \textbf{25.88} & \textbf{0.797} & \textbf{0.201} \\

\bottomrule

\end{tabular}
\label{table:matrixcity}
\end{minipage}
\hspace{0mm}
\begin{minipage}{0.47\textwidth}
\centering
\caption{Quantitative results for scenes in MipNeRF-360~\cite{miperf360} Dataset. We take average over five outdoor scenes and four indoor scenes. Best in \textbf{bold}.
}
\vspace{+2mm}
\setlength{\tabcolsep}{8pt}
\scriptsize
\begin{tabular}{@{}lccc@{}}

\toprule 

 & PSNR$\uparrow$ & SSIM$\uparrow$ & LPIPS $\downarrow$ \\ 
\midrule
NeRF\cite{nerf} & 23.85 & 0.605 & 0.451 \\
MipNeRF\cite{mipnerf} & 24.04 & 0.616 & 0.441 \\
NeRF++\cite{nerf++} & 25.11 & 0.676 &  0.375\\ 
Deep Blending\cite{deepblend} & 23.70 & 0.666 & 0.318 \\
Instant-NGP\cite{instantngp} & 25.68 & 0.705 & 0.302\\
MipNeRF-360\cite{miperf360} & 27.57& 0.793 & 0.234\\
ZipNeRF\cite{zipnerf} & 28.57 & 0.833 & 0.224 \\
ZipNeRF+NeRFLiX\cite{nerflix} & 28.73 & 0.832 & 0.240 \\
ZipNeRF+\ours & \textbf{29.41} & \textbf{0.850} & \textbf{0.134} \\

\bottomrule

\end{tabular}
\label{table:360}
\end{minipage}
\vspace{-8mm}
\end{table*}

\begin{figure*}[t]
	\small
	\centering
        \begin{minipage}[t]{0.2\linewidth}
		\centering
		\centerline{\includegraphics[width=0.99\linewidth]{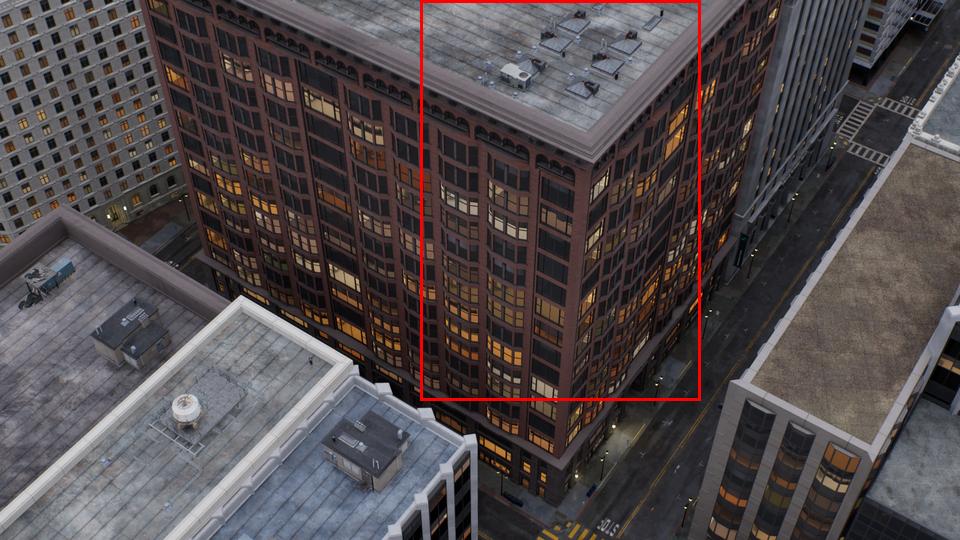}}
		\vspace{-0.3mm}
	\end{minipage}
	\begin{minipage}[t]{0.15\linewidth}
		\centering
		\centerline{\includegraphics[width=0.99\linewidth]{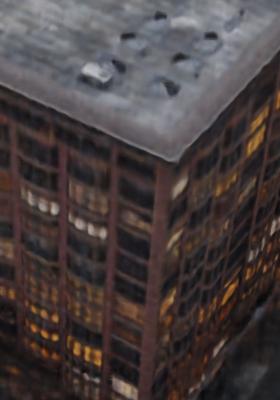}}
		\vspace{-0.3mm}
	\end{minipage}
	\begin{minipage}[t]{0.15\linewidth}
		\centering
		\centerline{\includegraphics[width=0.99\linewidth]{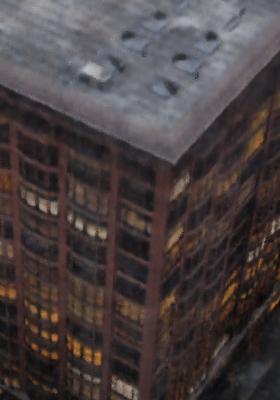}}
		\vspace{-0.3mm}
	\end{minipage}
	\begin{minipage}[t]{0.15\linewidth}
		\centering
		\centerline{\includegraphics[width=0.99\linewidth]{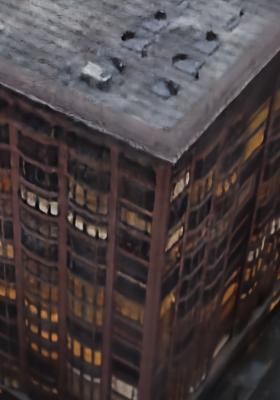}}
		\vspace{-0.3mm}
	\end{minipage}
    \begin{minipage}[t]{0.15\linewidth}
		\centering
		\centerline{\includegraphics[width=0.99\linewidth]{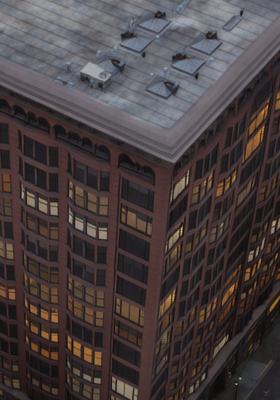}}
		\vspace{-0.3mm}
	\end{minipage}
	\begin{minipage}[t]{0.15\linewidth}
		\centering
		\centerline{\includegraphics[width=0.99\linewidth]{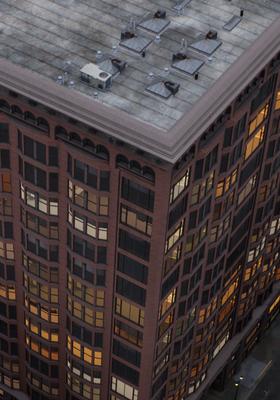}}
		\vspace{-0.3mm}
	\end{minipage}

    \begin{minipage}[t]{0.2\linewidth}
		\centering
		\centerline{\includegraphics[width=0.99\linewidth]{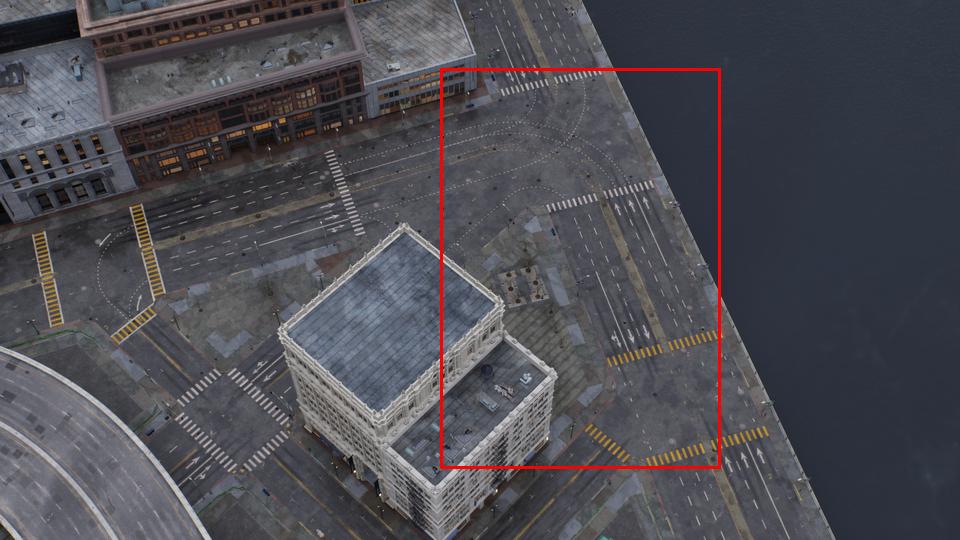}}
		\vspace{-0.3mm}
	\end{minipage}
	\begin{minipage}[t]{0.15\linewidth}
		\centering
		\centerline{\includegraphics[width=0.99\linewidth]{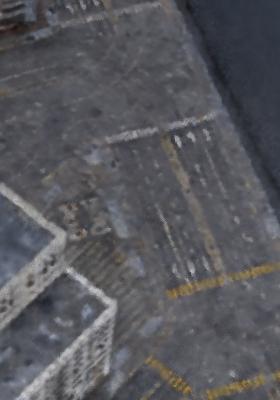}}
		\vspace{-0.3mm}
	\end{minipage}
	\begin{minipage}[t]{0.15\linewidth}
		\centering
		\centerline{\includegraphics[width=0.99\linewidth]{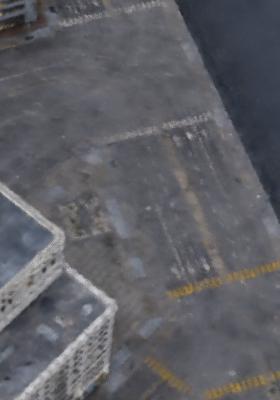}}
		\vspace{-0.3mm}
	\end{minipage}
	\begin{minipage}[t]{0.15\linewidth}
		\centering
		\centerline{\includegraphics[width=0.99\linewidth]{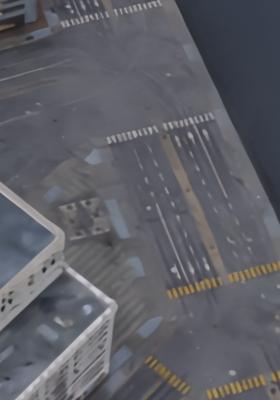}}
		\vspace{-0.3mm}
	\end{minipage}
    \begin{minipage}[t]{0.15\linewidth}
		\centering
		\centerline{\includegraphics[width=0.99\linewidth]{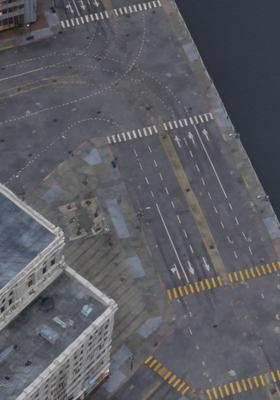}}
		\vspace{-0.3mm}
	\end{minipage}
	\begin{minipage}[t]{0.15\linewidth}
		\centering
		\centerline{\includegraphics[width=0.99\linewidth]{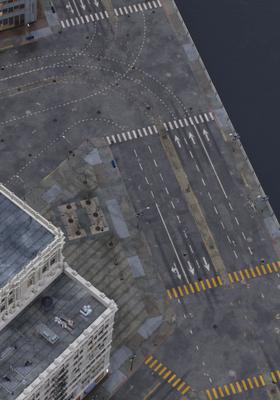}}
		\vspace{-0.3mm}
	\end{minipage}

    \begin{minipage}[t]{0.2\linewidth}
		\centering
		\centerline{\includegraphics[width=0.99\linewidth]{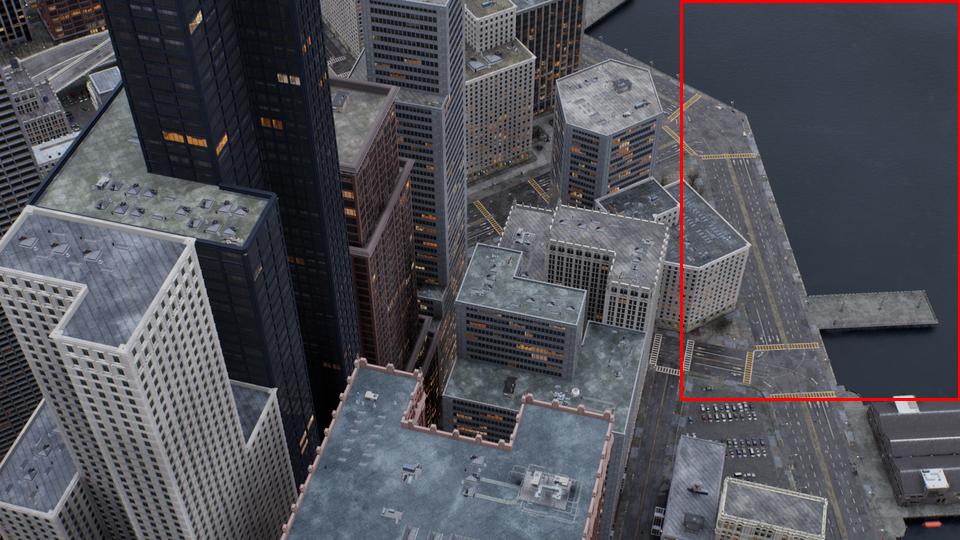}}
		\vspace{-0.3mm}
        \centerline{Full Image}
	\end{minipage}
	\begin{minipage}[t]{0.15\linewidth}
		\centering
		\centerline{\includegraphics[width=0.99\linewidth]{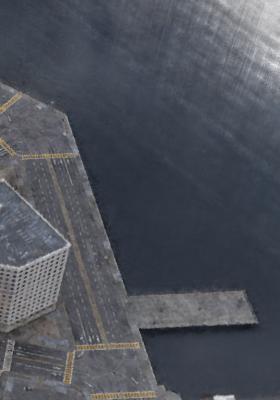}}
		\vspace{-0.3mm}
		\centerline{TensoRF}
	\end{minipage}
	\begin{minipage}[t]{0.15\linewidth}
		\centering
		\centerline{\includegraphics[width=0.99\linewidth]{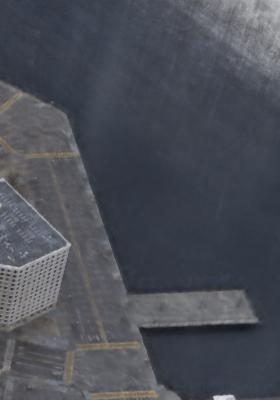}}
		\vspace{-0.3mm}
		\centerline{GridNeRF}
	\end{minipage}
    \begin{minipage}[t]{0.15\linewidth}
		\centering
		\centerline{\includegraphics[width=0.99\linewidth]{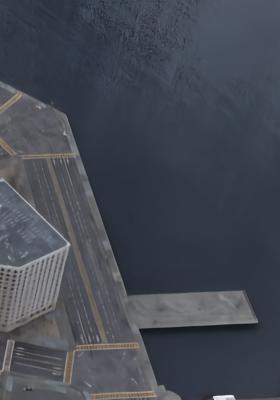}}
		\vspace{-0.3mm}
		\centerline{GridNeRF}
        \centerline{+NeRFLiX}
	\end{minipage}
	\begin{minipage}[t]{0.15\linewidth}
		\centering
		\centerline{\includegraphics[width=0.99\linewidth]{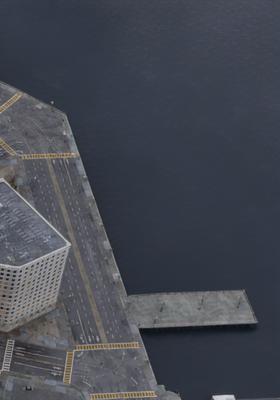}}
		\vspace{-0.3mm}
		\centerline{GridNeRF}
        \centerline{+\ours}
	\end{minipage}
	\begin{minipage}[t]{0.15\linewidth}
		\centering
		\centerline{\includegraphics[width=0.99\linewidth]{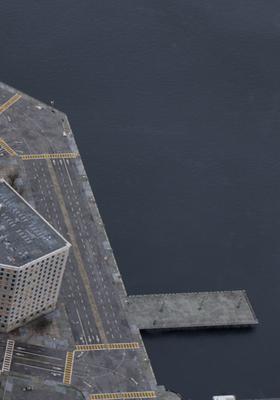}}
		\vspace{-0.3mm}
		\centerline{G.T.}
	\end{minipage}
	
	\vspace{-1mm}
	\caption{Visualizations on aerial scenes in MatrixCity dataset. Our proposed \ours~significantly produces renderings with more realistic texture details and much less aliasing artifacts, blurs and floaters. We crop a patch in each whole image and zoom in for more detailed comparison.}
	\label{fig:matrixcity}
	\vspace{-4mm}
\end{figure*}


\section{Experiments}
\label{sec:Exp}
To validate the effectiveness of our proposed method~\ours, 
we conduct experiments on two challenging datasets: the large-scale urban scene dataset MatrixCity~\cite{matrixcity}, and the unbounded 360-degree scene dataset MipNeRF-360~\cite{miperf360}, with two state-of-the-art methods GridNeRF~\cite{gridnerf} and ZipNeRF~\cite{zipnerf} as baselines respectively. We also use another state-of-the-art NeRF-based restoration framework NeRFLiX~\cite{nerflix} for a more comprehensive comparison.
The experimental results are presented in Section \ref{exp:urban} and Section \ref{exp:360}, while the ablation studies are presented in Section \ref{exp:abla}. We employ PSNR, SSIM~\cite{ssim} and LPIPS~\cite{lpips} metrics for quantitative evaluation. 

\subsection{Effectiveness on the Large-scale Urban Scene}
\label{exp:urban}
MatrixCity~\cite{matrixcity} is a large-scale, comprehensive, and high-quality city-block dataset containing 67k aerial images, covering an expansive area of 28 square kilometers. The aerial captures involve six blocks: \textit{Block\_A}, \textit{Block\_B}, \textit{Block\_C}, \textit{Block\_D}, \textit{Block\_E} and \textit{Block\_ALL}. The image renderings and camera poses in \textit{Block\_ALL} scene is actually the union set for all the other five block scenes.
Owing to the complexity of these scenes, which encompass a diverse array of buildings, streets, and water bodies, traditional NeRF methods~\cite{nerf, tensorf, meganerf} often exhibit severe rendering degradations, including aliasing artifacts, floaters and over-smoothed contents. 
To address these issues, GridNeRF~\cite{gridnerf} decomposes the large-scale scene into a multi-resolution pyramid of 
voxel grids. 
Nevertheless, due to the limited voxel resolutions, the renderings produced by GridNeRF still suffer from aliasing artifacts, especially for the fine texture details in street lines and building outlines.

We add our diffusion-based anti-aliasing method \textbf{\ours~}upon GridNeRF, and compare with NeRF~\cite{nerf}, BungeeNeRF~\cite{bungeenerf}, MegaNeRF~\cite{meganerf}, SwitchNeRF~\cite{switchnerf} and TensoRF~\cite{tensorf} as well. The numerical results of these comparisons are presented in Table \ref{table:matrixcity}. 
We report the average metrics across \textit{Block\_A} $\sim$ \textit{Block\_E} in Table \ref{table:matrixcity} as common practice, and the detailed breakdown results for each scene are put in Table \ref{table:matrix_per_scene}.
In accordance with GridNeRF, all the training and testing images are downsampled by a factor of $2 \times$ to the resolution of $540 \times 960$ for a fair comparison.
As for the city block scenes, since different blocks contain similar textural and structural contents, the degradation types for different blocks adhere to analogous distributions and could be modeled merely by a single diffusion model. In consideration of this, we either train the per-scene \textbf{spe}cific diffusion model for 
each block separately (denoted as \ours$_{spe}$), 
or we just train a cross-scene \textbf{gen}eralizable diffusion model for all the block scenes (denoted as
\ours$_{gen}$). In practice, we use the diffusion model trained along with the NeRF for the \textit{Block\_ALL} scene as the default \ours$_{gen}$. As evidenced in Table \ref{table:matrixcity}, even when only equipped with a single diffusion model \textit{generalized} across all the city blocks, \ours~already achieves state-of-the-art performance by a large margin, while the per-scene \textit{specific} diffusion models bring further improvements. It indicates that, equipped with the strong generation capability, the diffusion model can effectively restore the high-fidelity texture details from relatively low-quality aliased and blurred renderings, and performing the two-stage training procedure as described in Section \ref{sec:methodoverall} only once a time is sufficiently generalizable for a kind of scenes.

The visualizations are presented in Figure \ref{fig:matrixcity}, and more visual comparisons could be found in Appendix \ref{more results for matrixcity}.
We use \ours$_{spe}$ as the default version of \ours~to verify the ceiling of our proposed method. In the Appendix \ref{more results for matrixcity} we further show that \ours$_{gen}$ could perform as well as \ours$_{spe}$ in most of the scene cases.
It is clear that the diffusion-based restoration method can effectively eliminate most of the blurring and aliasing artifacts, and restore more high-realism texture details. This is particularly noticeable in the window contours, the outlines of the building and the lines on the roads. Additionally, \ours~is capable of removing the floaters and sharpening the fuzzy edges, especially in the under-constrained regions such as water surfaces and empty spaces. Clearly, our proposed \ours~yields a remarkable improvement in the quality of the visual representation.

\vspace{+3mm}
\begin{figure*}[t]
	\small
	\centering
    \begin{minipage}[t]{0.2\linewidth}
		\centering
		\centerline{\includegraphics[width=0.99\linewidth]{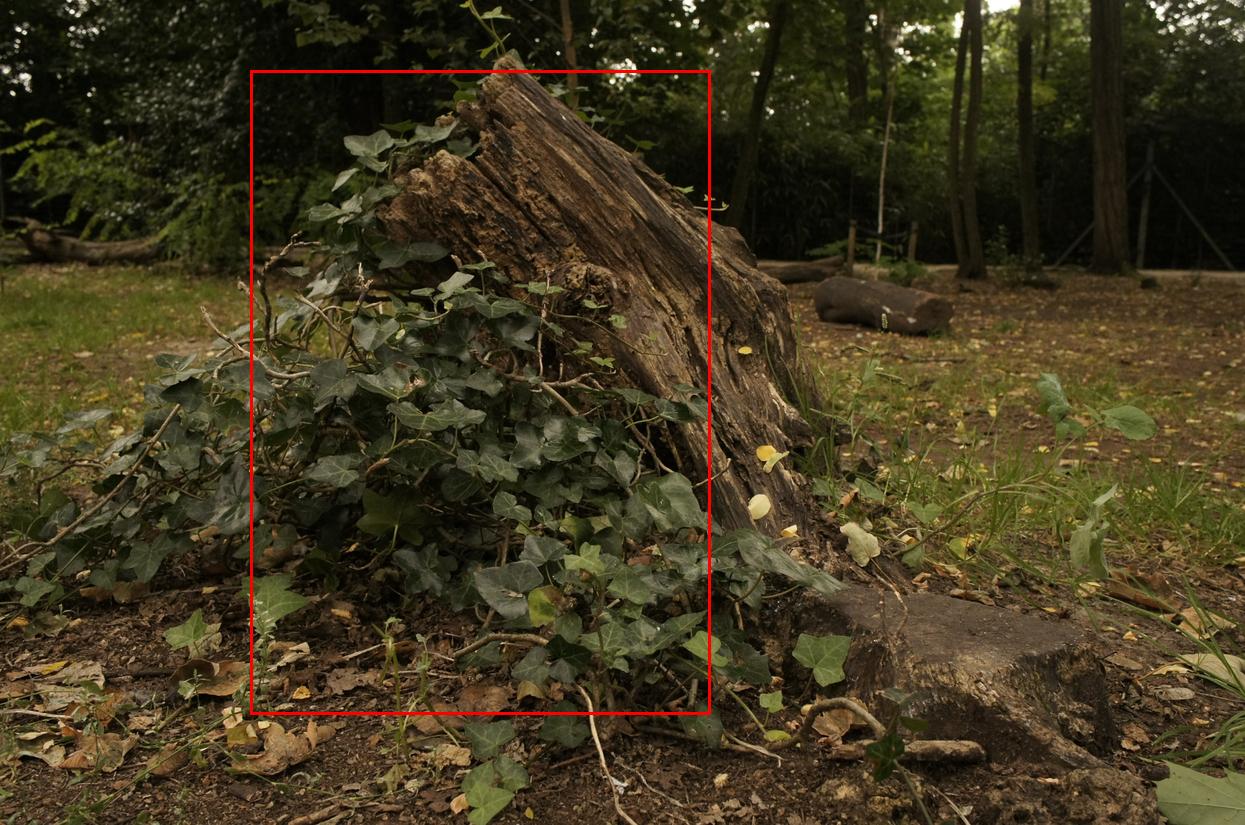}}
		\vspace{-0.3mm}
	\end{minipage}
	\begin{minipage}[t]{0.15\linewidth}
		\centering
		\centerline{\includegraphics[width=0.99\linewidth]{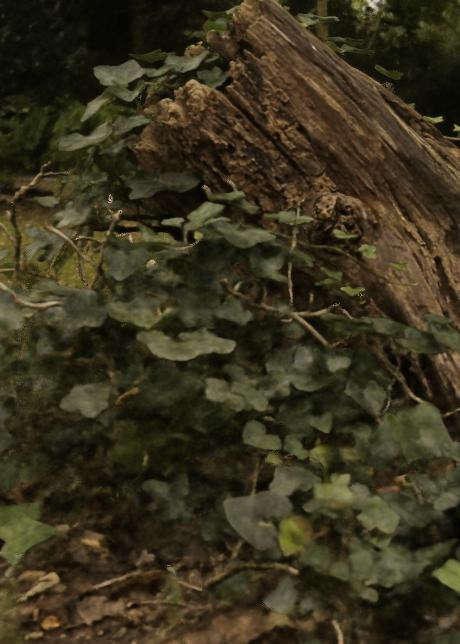}}
		\vspace{-0.3mm}
	\end{minipage}
	\begin{minipage}[t]{0.15\linewidth}
		\centering
		\centerline{\includegraphics[width=0.99\linewidth]{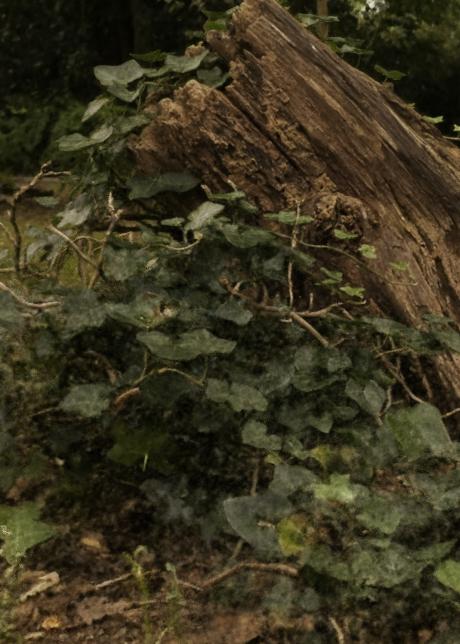}}
		\vspace{-0.3mm}
	\end{minipage}
	\begin{minipage}[t]{0.15\linewidth}
		\centering
		\centerline{\includegraphics[width=0.99\linewidth]{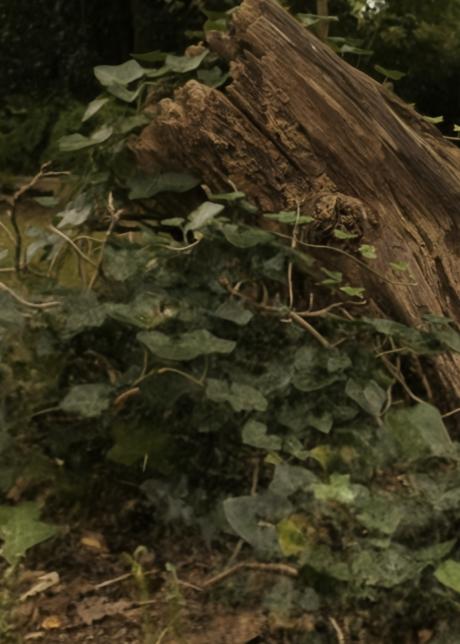}}
		\vspace{-0.3mm}
	\end{minipage}
    \begin{minipage}[t]{0.15\linewidth}
		\centering
		\centerline{\includegraphics[width=0.99\linewidth]{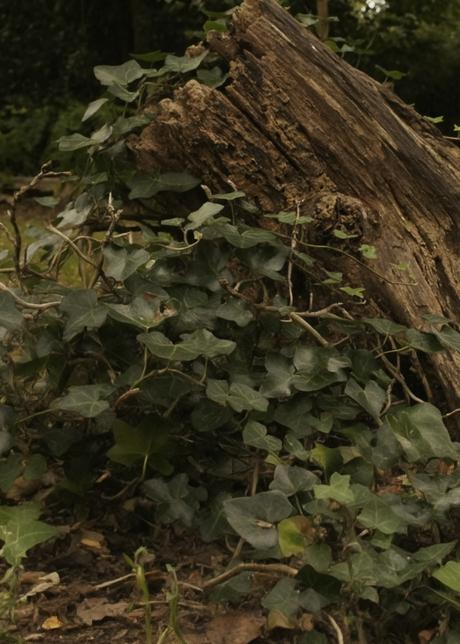}}
		\vspace{-0.3mm}
	\end{minipage}
	\begin{minipage}[t]{0.15\linewidth}
		\centering
		\centerline{\includegraphics[width=0.99\linewidth]{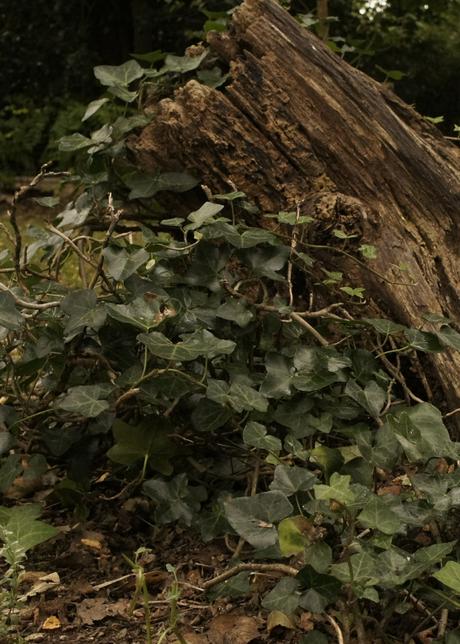}}
		\vspace{-0.3mm}
	\end{minipage}

    \begin{minipage}[t]{0.2\linewidth}
	\centering
		\centerline{\includegraphics[width=0.99\linewidth]{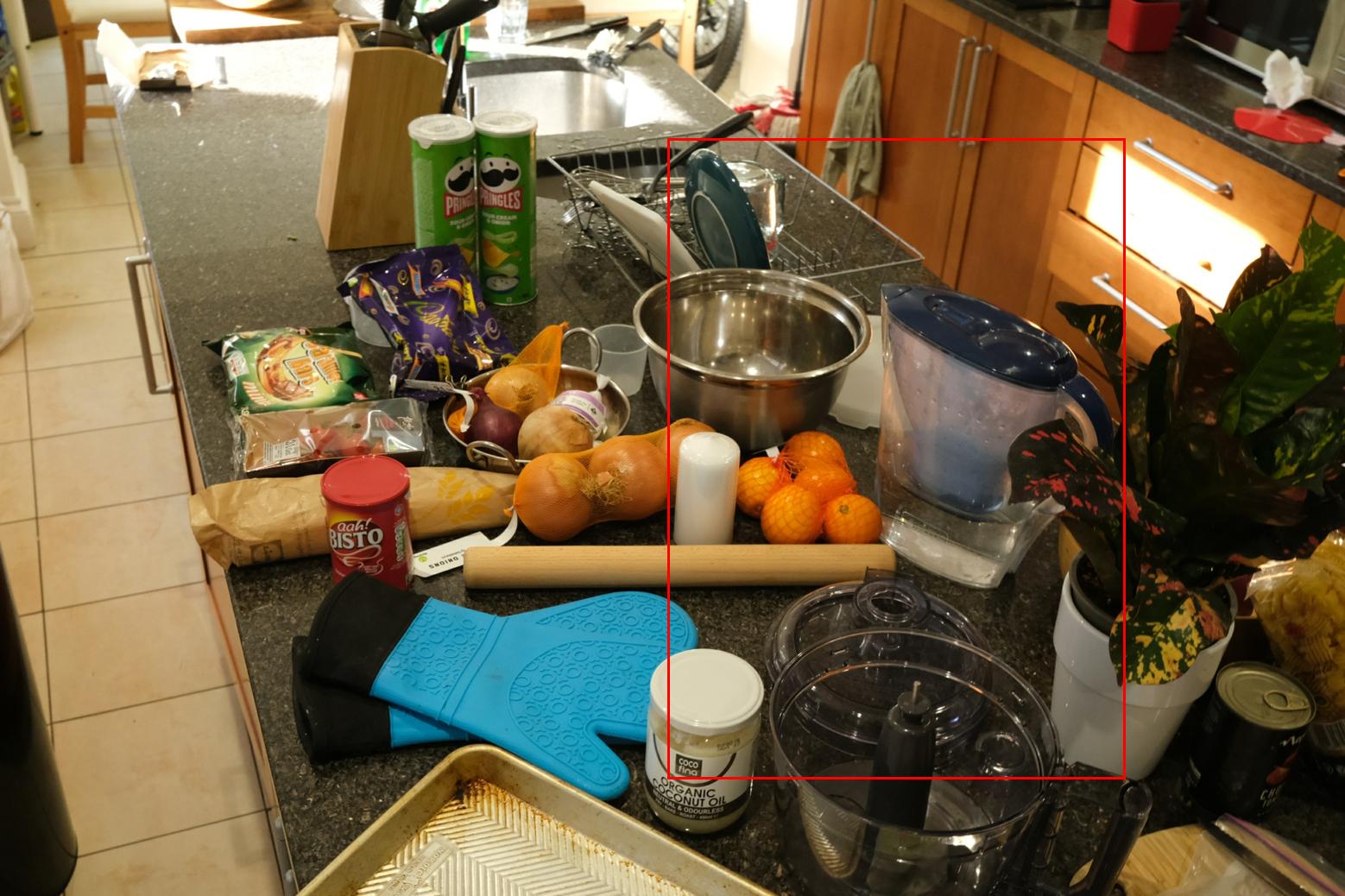}}
		\vspace{-0.3mm}
	\end{minipage}
	\begin{minipage}[t]{0.15\linewidth}
		\centering
		\centerline{\includegraphics[width=0.99\linewidth]{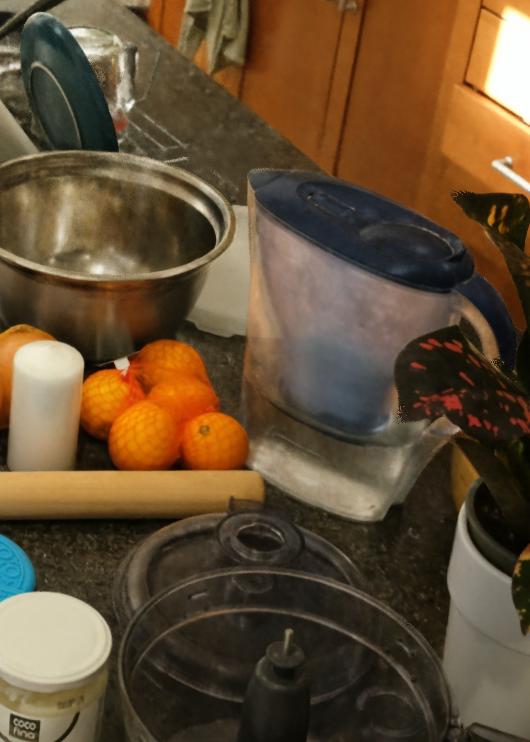}}
		\vspace{-0.3mm}
	\end{minipage}
	\begin{minipage}[t]{0.15\linewidth}
		\centering
		\centerline{\includegraphics[width=0.99\linewidth]{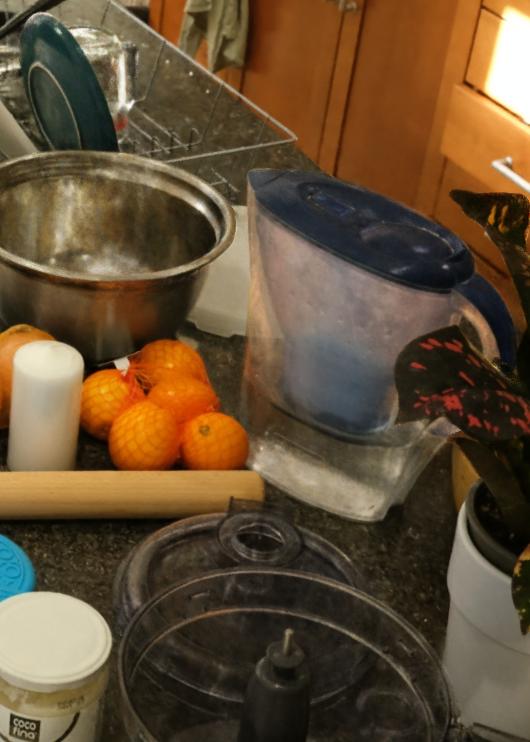}}
		\vspace{-0.3mm}
	\end{minipage}
	\begin{minipage}[t]{0.15\linewidth}
		\centering
		\centerline{\includegraphics[width=0.99\linewidth]{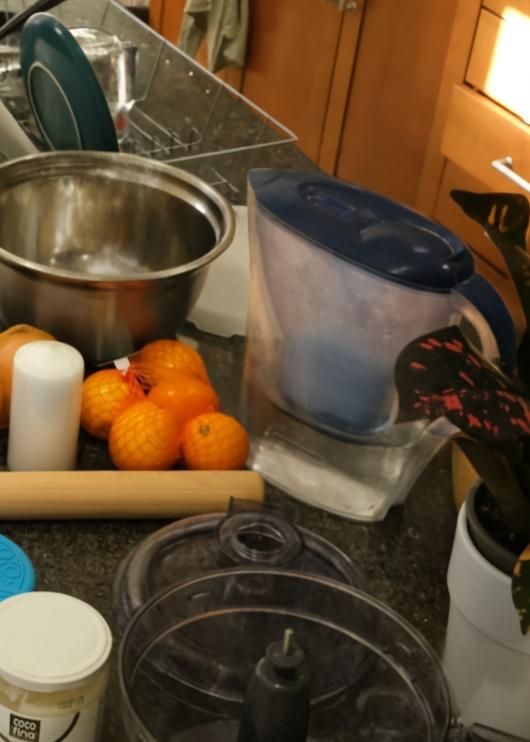}}
		\vspace{-0.3mm}
	\end{minipage}
    \begin{minipage}[t]{0.15\linewidth}
		\centering
		\centerline{\includegraphics[width=0.99\linewidth]{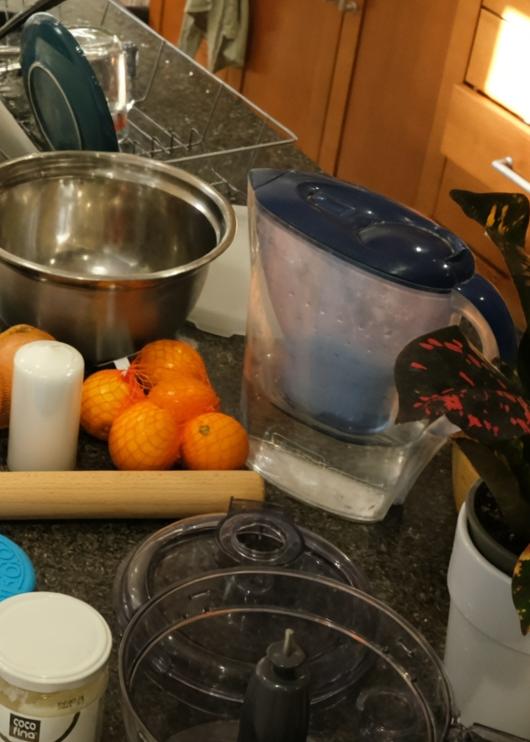}}
		\vspace{-0.3mm}
	\end{minipage}
	\begin{minipage}[t]{0.15\linewidth}
		\centering
		\centerline{\includegraphics[width=0.99\linewidth]{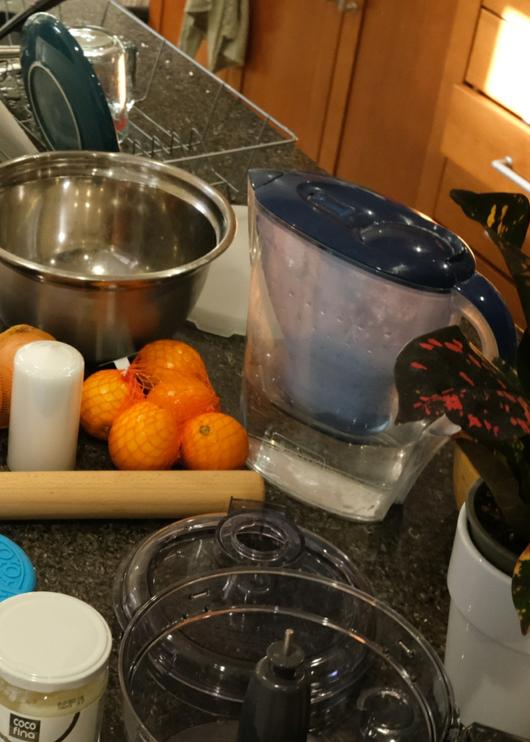}}
		\vspace{-0.3mm}
	\end{minipage}

    \begin{minipage}[t]{0.2\linewidth}
		\centering
		\centerline{\includegraphics[width=0.99\linewidth]{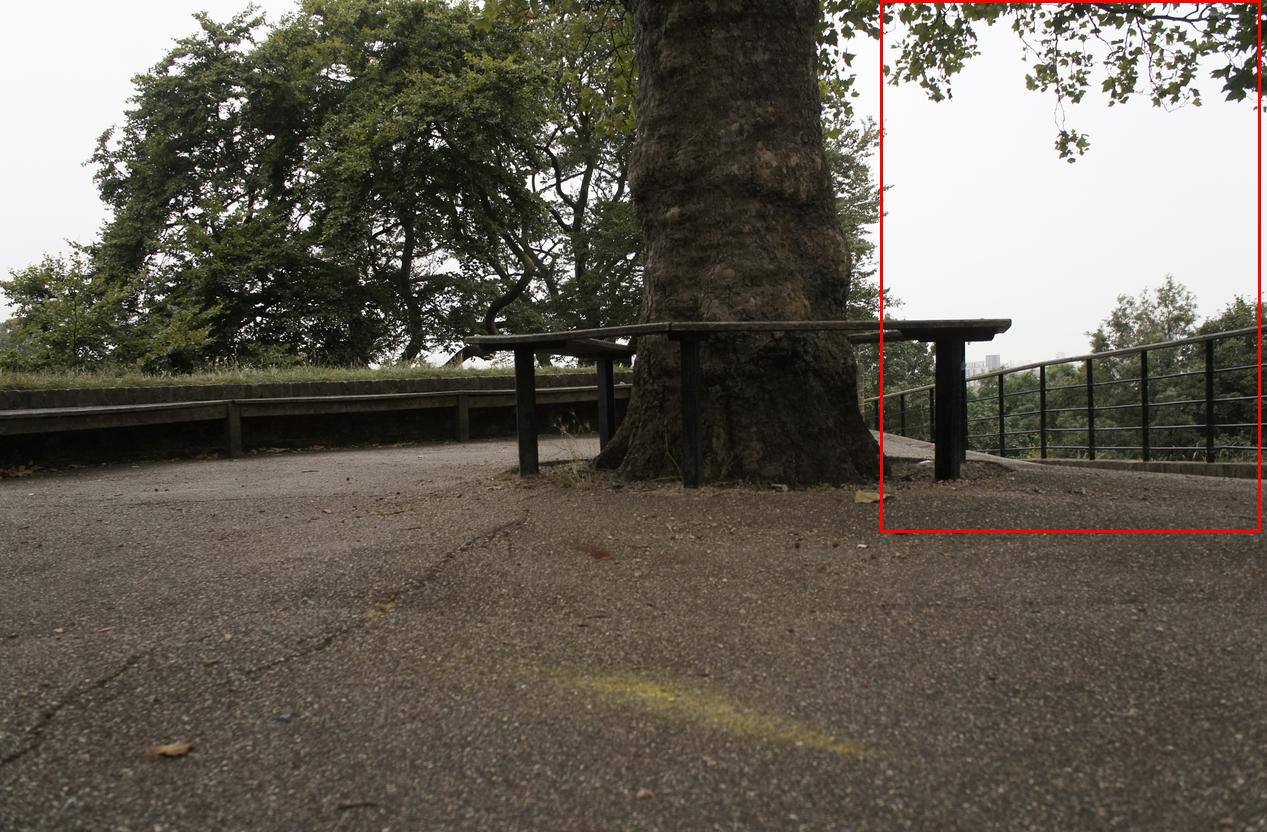}}
		\vspace{-0.3mm}
		\centerline{Full Image}
	\end{minipage}
	\begin{minipage}[t]{0.15\linewidth}
		\centering
		\centerline{\includegraphics[width=0.99\linewidth]{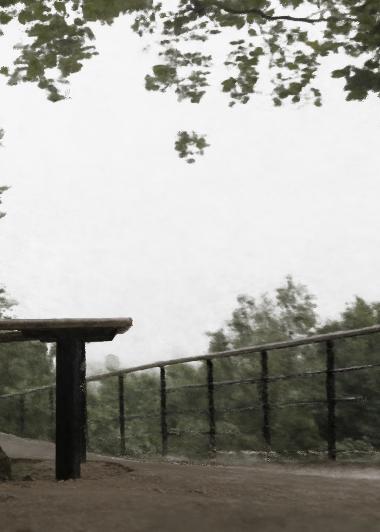}}
		\vspace{-0.3mm}
		\centerline{MipNeRF-360}
	\end{minipage}
	\begin{minipage}[t]{0.15\linewidth}
		\centering
		\centerline{\includegraphics[width=0.99\linewidth]{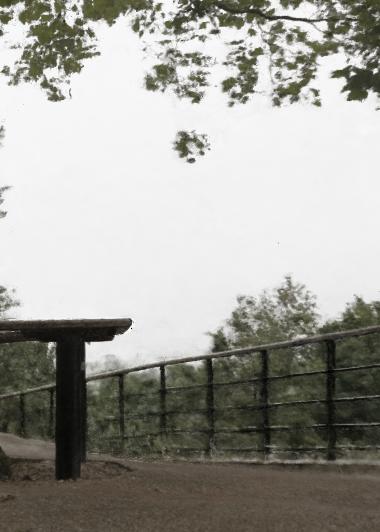}}
		\vspace{-0.3mm}
		\centerline{ZipNeRF}
	\end{minipage}
    \begin{minipage}[t]{0.15\linewidth}
		\centering
		\centerline{\includegraphics[width=0.99\linewidth]{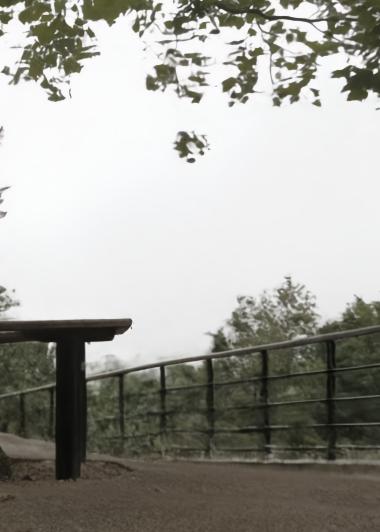}}
		\vspace{-0.3mm}
		\centerline{ZipNeRF}
        \centerline{+NeRFLiX}
	\end{minipage}
	\begin{minipage}[t]{0.15\linewidth}
		\centering
		\centerline{\includegraphics[width=0.99\linewidth]{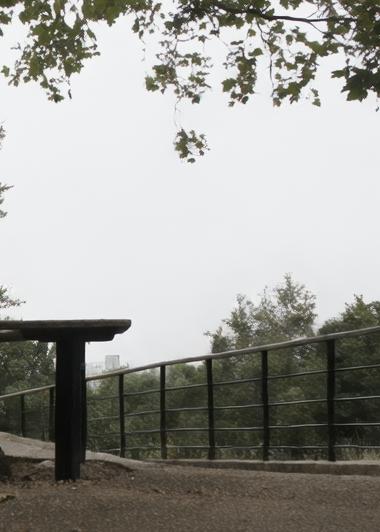}}
		\vspace{-0.3mm}
		\centerline{ZipNeRF}
        \centerline{+\ours}
	\end{minipage}
	\begin{minipage}[t]{0.15\linewidth}
		\centering
		\centerline{\includegraphics[width=0.99\linewidth]{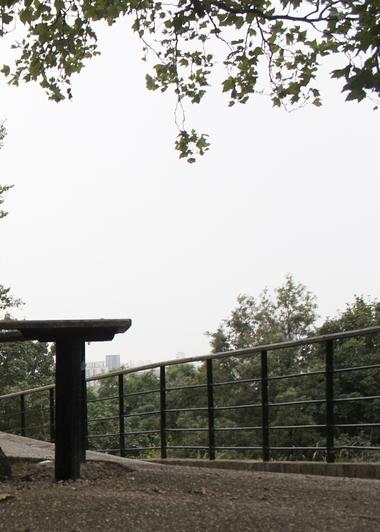}}
		\vspace{-0.3mm}
		\centerline{G.T.}
	\end{minipage}
 
	\caption{Visualizations for scene \textit{stump}, \textit{counter} and \textit{treehill} (from top to down) in MipNeRF-360 dataset. Drantal exhibits strong capability in restoring more realistic anti-aliasing renderings, with sharper edges and finer details. We crop a patch and zoom in for more detailed comparison.}
	\label{fig: 360}
	\vspace{-5mm}
\end{figure*}




\subsection{Effectiveness on the Unbounded 360-degree Scene}
\label{exp:360}
We also investigate the effectiveness of our proposed method on unbounded 360-degree scenes. MipNeRF-360 dataset~\cite{miperf360} contains five outdoor scenes and four indoor scenes, with the scene contents that may exist at arbitrary distance. These scenes contain a lot of fine-grained details, making them extremely challenging to be reconstructed perfectly. Previous representative works, such as MipNeRF-360~\cite{miperf360} and ZipNeRF~\cite{zipnerf} have achieved commendable results, but there is still scope for further improvements. 
Given that the degradation types vary across the nine disparate scenes, we independently optimize 
a diffusion model for each scene along with the  training process of NeRF. 
By integrating \ours~into the currently leading method, ZipNeRF, we achieve significant performance improvements across all the nine scenes. As presented in Table \ref{table:360}, we report the average score over five outdoor and four indoor scenes, which includes an increase of nearly $0.85$dB in PSNR, $0.02$ in SSIM and a decrease of $0.09$ in LPIPS. More detailed results for each scene are presented in Table \ref{table:360_per_scene}. 
The visual comparisons in Figure \ref{fig: 360} also demonstrate the superiority of our method. Please refer to the textures of the leaves on the \textit{stump} (the first row), the reflections of the metal bowl on the \textit{counter} (the second row), as well as the structures of the fences in the \textit{treehill} (the last row).
Due to page limits, more numerical results, visual comparisons and analyses could be found in Appendix \ref{more results for 360}.

\noindent \textbf{Discussion}
NeRFLiX~\cite{nerflix} is also a NeRF-based restoration framework, similar to ours. However, it uses several (mostly two) training views as the reference guidance for restoring the target view. Practically, the ground truths for training views are not always available at inference time. What's more, the lack of usage of diffusion model in NeRFLiX results in enhanced images that appear blurry and overly smoothed.
The quantitative and qualitative comparisons across the two datasets in Section \ref{exp:urban} and \ref{exp:360} better demonstrate the superiority of our proposed \ours.


\subsection{Ablation Studies}
\label{exp:abla}

\begin{wraptable}{r}{0.5\textwidth}
\vspace{-9mm}
\begin{center}
\caption{Ablation study on \textit{bicycle} scene in MipNeRF-360 dataset. We compare the rest of rows to the default setting in the second row.}
\label{table:ablation} 
\setlength{\tabcolsep}{5pt}
\scriptsize
\begin{tabular}{@{}lccc@{}}

\toprule 

 & PSNR$\uparrow$ & SSIM$\uparrow$ & LPIPS $\downarrow$ \\ 
\midrule
1) ZipNeRF & 25.90 & 0.780 & 0.231 \\
\midrule
2) ZipNeRF+\ours \, (Default) & 27.18 & 0.806 & 0.143 \\
\midrule
3) Only stage-one training & 22.17 & 0.605 & 0.229 \\
4) Separate stage-one training & 26.98 & 0.801 & 0.155 \\
\midrule
5) Only CFW module & 25.82 & 0.773 & 0.175 \\
5) CFW and decoder last layer & 26.34 & 0.791 & 0.163 \\
\midrule
6) $w=0.5$ & 25.83 & 0.751 & 0.198 \\
6) $w=0.75$ & 26.89 & 0.792 & 0.154 \\
\midrule
7) DDIM, 50 steps & 27.18 & 0.807 & 0.141 \\
7) DDPM, 200 steps & 27.22 & 0.810 & 0.137 \\

\bottomrule

\end{tabular}
\end{center}
\vspace{-2mm}
\vspace{-4mm}
\end{wraptable}

In this section, we conduct several ablation studies regarding the \textit{bicycle} scene in MipNeRF-360 dataset to verify the effectiveness of each component proposed in Section \ref{sec:methodoverall}. In Appendix \ref{appendix:ablation}, we 
study the trend of performance gains in relation to voxel resolutions, which are conducted on \textit{Block\_ALL} scene in the MatrixCity dataset.

As shown in Table \ref{table:ablation}, 1) ZipNeRF represents the performance of our baseline and 2) ZipNeRF+Drantal (Default) represents the default setting of our proposed method \ours, which corresponds to the overall result in Table \ref{table:360}. 3) Only stage-one training means that we do not conduct stage-two training of feature wrapping, and only adopts the finetuned diffusion model after stage-one training for anti-aliasing. Although the images produced by the diffusion model may appear visually appealing, the inherent generation capability adversely affects the fidelity of the output images, resulting in a notable decline in PSNR and SSIM. 4) Separate stage-one training indicates that we still adopt two-stage training, but in the first stage, we optimize NeRF and diffusion model successively rather than jointly. This leads to slight performance loss, as diffusion model could not encounter more diverse types of degradations while training the NeRF. 5) Either optimizing only the CFW module or the CFW module along with the last layer of the VAE decoder during the second stage training is sub-optimal compared to our default setting, as a learnable VAE decoder has a better capability to fit the degradation distribution of a specific scene. 6) For the stage-two training, we set controllable adjustable coefficient $w=1$. If $w$ is set to a smaller value at inference time to bias towards stronger generation rather than reconstruction, the quality of reconstruction (PSNR, SSIM and LPIPS) consequently weakens.
7) We utilize DDIM~\cite{ddim} sampler with 20 steps by default. Using DDIM~\cite{ddim} with 50 steps or DDPM~\cite{ddpm} with 200 steps slightly boost the performance. Nevertheless, both options increase the  inference time greatly, so we do not adopt them as the default setting.


\section{Limitations}
\label{sec:limit}
Although \ours~exhibits strong capability in restoring various NeRF-based degradations, the incorporation of the diffusion model brings extra overhead to the entire model, slowing down the rendering speed at inference time. This could be resolved potentially by utilizing more advanced diffusion acceleration techniques~\cite{diff-acc1, diff-acc2, diff-acc3, diff-acc4}. Moreover, it would be promising to introduce \ours~to other 3D reconstruction methods like the Gaussian Splatting~\cite{3dGS} as the future work. 

\section{Conclusion}
In this paper, we propose a diffusion-based restoration method for anti-aliasing Neural Radiance Field (\textbf{Drantal-NeRF}). We regard the aliasing artifacts in the renderings of NeRF as a type of unknown degradation and utilize the diffusion model to generate high-fidelity realistic renderings conditioned on the aliased inputs. Extensive experiments demonstrate the effectiveness of our proposed method in handling various kinds of degradations, for different instances of NeRF backbones, under both large-scale urban scenes and unbounded 360-degree scenes. We hope our proposed method could serve as a general framework in the post-processing stage for anti-aliasing Neural Radiance Field, and could further inspire viewing the anti-aliasing issue from another simple but novel low-level restoration perspective.

\section*{Ethics statement}

Drantal-NeRF is purely a research project. Currently, we have no plans to incorporate Drantal-NeRF into a product or expand access to the public. We will also put Microsoft AI principles into practice when further developing the models. All the datasets used in this paper are public and have been reviewed to ensure they do not contain any personally identifiable information or offensive content.



\bibliographystyle{ieeenat_fullname}
{\small
  \bibliography{ref}
}


\appendix
\newpage



\section{More Experimental Results}

\begin{table}[ht]
\centering
\caption{Quantitative results for each scene in MatrixCity~\cite{matrixcity} Dataset. Best in \textbf{bold}.
}
\setlength{\tabcolsep}{3pt}
\normalsize
\begin{tabular}{@{}lcccccc@{}}

\toprule 
PSNR $\uparrow$ &  Block\_A & Block\_B & Block\_C   
 &  Block\_D & Block\_E & Block\_ALL  \\
\midrule
NeRF\cite{nerf} & 23.15 & 22.04 & 21.98 & 23.09 & 22.64 & 22.07 \\
BungeeNeRF\cite{bungeenerf} & 24.55 & 23.02 & 22.53 & 23.59
& 24.09 & 22.52\\
MegaNeRF\cite{meganerf} & 24.71 & 22.93 & 22.47 & 23.84
& 23.96 & 22.55 \\
SwitchNeRF\cite{switchnerf} & 25.36 & 23.78 & 23.09 & 23.96
& 24.39 & 23.15 \\
TensoRF\cite{tensorf} & 25.51 & 23.92 & 23.21 & 24.37 & 24.84
 & 23.41 \\ 
GridNeRF\cite{zipnerf} & 26.01 & 24.59 & 23.76 & 24.86 & 25.19 & 24.79 \\
GridNeRF+NeRFLiX\cite{nerflix} & 26.07 & 24.85 & 23.93 & 24.90 & 25.29 & 24.84 \\
GridNeRF+\ours$_{gen}$ & 26.53 & 25.57 & 24.62 & 25.71 & 25.61
 & 25.69 \\
GridNeRF+\ours$_{spe}$ & \textbf{26.84} & \textbf{26.06} & \textbf{24.85} & \textbf{25.82} & \textbf{25.81}
 & \textbf{25.69} \\
\bottomrule
\end{tabular}
\vspace{+3mm}

\begin{tabular}{@{}lcccccc@{}}

\toprule 
SSIM $\uparrow$ &  Block\_A & Block\_B & Block\_C   
 &  Block\_D & Block\_E & Block\_ALL  \\
\midrule
NeRF\cite{nerf} & 0.561 & 0.613 & 0.595 & 0.588 & 0.598
 & 0.555 \\
BungeeNeRF\cite{bungeenerf} & 0.632 & 0.731 & 0.712 & 0.659
& 0.698 & 0.583 \\
MegaNeRF\cite{meganerf} & 0.662 & 0.714 & 0.700 & 0.688
& 0.681 & 0.588 \\
SwitchNeRF\cite{switchnerf} & 0.694 & 0.747 & 0.719 & 0.705
& 0.708 & 0.620 \\
TensoRF\cite{tensorf} & 0.712 & 0.756 & 0.722 & 0.720 & 0.712
 & 0.645 \\ 
GridNeRF\cite{zipnerf} & 0.743 & 0.773 & 0.747 & 0.735 & 0.750 & 0.722 \\
GridNeRF+NeRFLiX\cite{nerflix} & 0.763 & 0.786 & 0.763
& 0.755 & 0.770 & 0.747 \\
GridNeRF+\ours$_{gen}$ & 0.772 & 0.819 & 0.791 & 0.776 & 0.779
 & 0.770 \\
GridNeRF+\ours$_{spe}$ & \textbf{0.788} & \textbf{0.832} & \textbf{0.800} & \textbf{0.780} & \textbf{0.786}
 & \textbf{0.770} \\
\bottomrule
\end{tabular}
\vspace{+3mm}

\begin{tabular}{@{}lcccccc@{}}

\toprule 
LPIPS $\downarrow$ &  Block\_A & Block\_B & Block\_C  
 &  Block\_D & Block\_E & Block\_ALL  \\
\midrule
NeRF\cite{nerf} & 0.649 & 0.512 & 0.577 & 0.550 & 0.567
 & 0.622 \\
BungeeNeRF\cite{bungeenerf} & 0.556 & 0.430 & 0.467 & 0.542
& 0.503 & 0.597 \\
MegaNeRF\cite{meganerf} & 0.538 & 0.447 & 0.489 & 0.508
& 0.511 & 0.580 \\
SwitchNeRF\cite{switchnerf} & 0.512 & 0.409 & 0.421 & 0.488
& 0.476 & 0.569 \\
TensoRF\cite{tensorf} & 0.485 & 0.387 & 0.407 & 0.448
& 0.455 & 0.558 \\ 
GridNeRF\cite{zipnerf} & 0.428 & 0.331 & 0.371 & 0.393
& 0.404 & 0.425\\
GridNeRF+NeRFLiX\cite{nerflix} & 0.399 & 0.322 & 0.364
& 0.365 & 0.373 & 0.386 \\
GridNeRF+\ours$_{gen}$ & 0.222 & 0.184 & 0.245 & 0.213
& 0.238 & 0.206 \\
GridNeRF+\ours$_{spe}$ & \textbf{0.207} & \textbf{0.165} & \textbf{0.219} & \textbf{0.198} & \textbf{0.214}
& \textbf{0.206} \\
\bottomrule
\end{tabular}
\vspace{+2mm}
\label{table:matrix_per_scene}
\vspace{-3mm}
\end{table}

\begin{table}[ht]
\centering
\caption{Quantitative results for each scene in MipNeRF-360~\cite{miperf360} Dataset (five outdoor scenes and four indoor scenes). Best in \textbf{bold}.
}
\setlength{\tabcolsep}{3pt}
\normalsize
\begin{tabular}{@{}lccccc|cccc@{}}

\toprule 
\multirow{2}{*}{PSNR $\uparrow$} & \multicolumn{5}{c|}{Outdoor} & \multicolumn{4}{c}{Indoor}  \\
\cmidrule{2-6}
\cmidrule{7-10}
 &  bicycle & flowers & garden   
 &  stump & treehill & room & counter & kitchen & bonsai \\
\midrule
NeRF\cite{nerf} & 	21.76 & 19.40 & 23.11 & 21.73 & 21.28 & 28.56 & 25.67 & 26.31 & 26.81  \\
MipNeRF\cite{mipnerf} & 21.69 & 19.31 & 23.16 & 23.10 & 21.21 & 28.73 & 25.59 & 26.47 & 27.13	 \\
NeRF++\cite{nerf++} & 22.64 & 20.31 & 24.32 & 24.34 & 22.20 & 28.87 & 26.38 & 27.80 & 29.15  \\ 
Deep Blending\cite{deepblend} & 21.09 & 18.13 & 23.61 & 24.08 & 20.80 & 27.20 & 26.28 & 25.02 & 27.08 \\
Instant-NGP\cite{instantngp} & 22.79 & 19.19 & 25.26 & 24.80 & 22.46 & 30.31 & 26.21 & 29.00 & 31.08  \\
MipNeRF-360\cite{miperf360} & 24.40 & 21.64 & 26.94 & 26.36 & 22.81 & 31.40 & 29.44 & 32.02 & 33.11  \\
ZipNeRF\cite{zipnerf} & 25.90 & 22.43 & 28.17 & 27.02 & 23.89 & 32.66 & 29.37 & 32.97 & 34.72 \\
ZipNeRF+NeRFLiX\cite{nerflix} & 26.07 & 22.40 & 27.77 & 27.26 & 24.06 & 32.96 & 29.70 & 33.15 & 35.20
\\
ZipNeRF+\ours & \textbf{27.18} & \textbf{23.00} & \textbf{28.87} & \textbf{28.22} & \textbf{24.80} & \textbf{33.09} & \textbf{30.26} & \textbf{33.77} & \textbf{35.50} \\
\bottomrule
\end{tabular}

\vspace{+3mm}

\begin{tabular}{@{}lccccc|cccc@{}}
\toprule 
\multirow{2}{*}{SSIM $\uparrow$} & \multicolumn{5}{c|}{Outdoor} & \multicolumn{4}{c}{Indoor}  \\
\cmidrule{2-6}
\cmidrule{7-10}
 &  bicycle & flowers & garden   
 &  stump & treehill & room & counter & kitchen & bonsai \\
\midrule
NeRF\cite{nerf} & 0.455 & 0.376 & 0.546 & 0.453 & 0.459 & 0.843 & 0.775 & 0.749 & 0.792 \\
MipNeRF\cite{mipnerf} & 0.454 & 0.373 & 0.543 & 0.517 & 0.466 & 0.851 & 0.779 & 0.745 & 0.818  \\
NeRF++\cite{nerf++} & 0.526 & 0.453 & 0.635 & 0.594 & 0.530 & 0.852 & 0.802 & 0.816 & 0.876   \\ 
Deep Blending\cite{deepblend} & 0.466 & 0.320 & 0.675 & 0.634 & 0.523 & 0.868 & 0.856 & 0.768 & 0.883 \\
Instant-NGP\cite{instantngp} & 0.540 & 0.378 & 0.709 & 0.654 & 0.547 & 0.893 & 0.845 & 0.857 & 0.924 \\
MipNeRF-360\cite{miperf360} & 0.693 & 0.583 & 0.816 & 0.746 & 0.632 & 0.913 & 0.895 &  0.920 & 0.939 \\
ZipNeRF\cite{zipnerf} & 0.780 & 0.657 & 0.874 & 0.791 &
0.657 & 0.929 & 0.907 & 0.941 & 0.956
 \\
ZipNeRF+NeRFLiX\cite{nerflix} & 0.778 & 0.645 & 0.854
& 0.799 & 0.665 & 0.931 & 0.915 & 0.942 & \textbf{0.961}
\\
ZipNeRF+\ours & \textbf{0.807} & \textbf{0.677} & \textbf{0.888} &
\textbf{0.827} & \textbf{0.693} & \textbf{0.933} & \textbf{0.916} & \textbf{0.953} &
0.956 \\
\bottomrule
\end{tabular}

\vspace{+3mm}

\begin{tabular}{@{}lccccc|cccc@{}}

\toprule 
\multirow{2}{*}{LPIPS $\downarrow$} & \multicolumn{5}{c|}{Outdoor} & \multicolumn{4}{c}{Indoor}  \\
\cmidrule{2-6}
\cmidrule{7-10}
 &  bicycle & flowers & garden   
 &  stump & treehill & room & counter & kitchen & bonsai \\
\midrule
NeRF\cite{nerf} & 0.536 & 0.529 & 0.415 & 0.551 & 0.546 & 0.353 & 0.394 & 0.335 & 0.398  \\
MipNeRF\cite{mipnerf} & 0.541 & 0.535 & 0.422 & 0.490 & 0.538 & 0.346 & 0.390 & 0.336 & 0.370	  \\
NeRF++\cite{nerf++} & 0.455 & 0.466 & 0.331 &  0.416 & 0.466 & 0.335 &  0.351 & 0.260 & 0.291  \\ 
Deep Blending\cite{deepblend} & 0.377 & 0.476 & 0.231 & 0.351 & 0.383 & 0.266 & 0.258 & 0.246 & 0.275 \\
Instant-NGP\cite{instantngp} & 0.398 & 0.441 & 0.255 & 0.339 & 0.420 & 0.242 & 0.255 & 0.170 & 0.198  \\
MipNeRF-360\cite{miperf360} & 0.289 & 0.345 & 0.164 & 0.254 & 0.338 & 0.211 & 0.203 & 0.126 & 0.177 \\
ZipNeRF\cite{zipnerf} & 0.232 & 0.307 & 0.128 & 0.247 & 0.293 & 0.244 & 0.231 & 0.137 & 0.200 \\
ZipNeRF+NeRFLiX\cite{nerflix} & 0.247 & 0.338 & 0.166
& 0.247 & 0.311 & 0.255 & 0.233 & 0.146 & 0.215
\\
ZipNeRF+\ours & \textbf{0.143} & \textbf{0.179} & \textbf{0.089}
& \textbf{0.141} & \textbf{0.186} & \textbf{0.142} & \textbf{0.136} & \textbf{0.092 }& \textbf{0.100} \\
\bottomrule
\end{tabular}

\vspace{-3mm}

\label{table:360_per_scene}
\vspace{-3mm}
\end{table}

\subsection{More Results for the Large-scale Urban Scenes}
\label{more results for matrixcity}
\begin{figure*}[t]
	\small
	\centering
        \begin{minipage}[t]{0.2\linewidth}
		\centering
		\centerline{\includegraphics[width=0.99\linewidth]{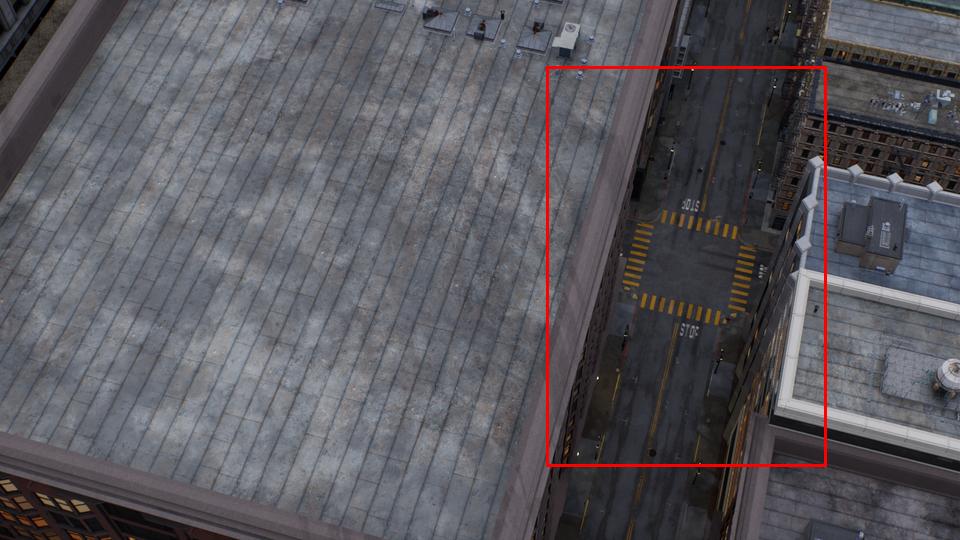}}
		\vspace{-0.3mm}
	\end{minipage}
	\begin{minipage}[t]{0.15\linewidth}
		\centering
		\centerline{\includegraphics[width=0.99\linewidth]{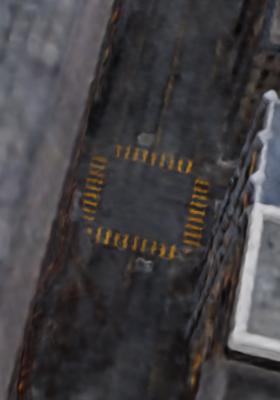}}
		\vspace{-0.3mm}
	\end{minipage}
	\begin{minipage}[t]{0.15\linewidth}
		\centering
		\centerline{\includegraphics[width=0.99\linewidth]{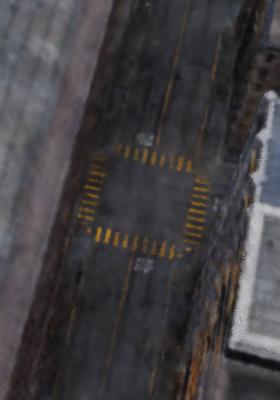}}
		\vspace{-0.3mm}
	\end{minipage}
	\begin{minipage}[t]{0.15\linewidth}
		\centering
		\centerline{\includegraphics[width=0.99\linewidth]{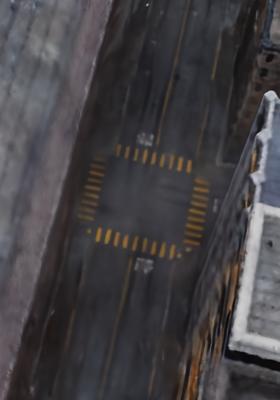}}
		\vspace{-0.3mm}
	\end{minipage}
    \begin{minipage}[t]{0.15\linewidth}
		\centering
		\centerline{\includegraphics[width=0.99\linewidth]{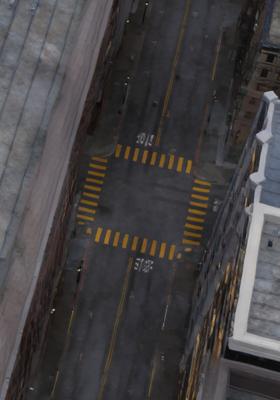}}
		\vspace{-0.3mm}
	\end{minipage}
	\begin{minipage}[t]{0.15\linewidth}
		\centering
		\centerline{\includegraphics[width=0.99\linewidth]{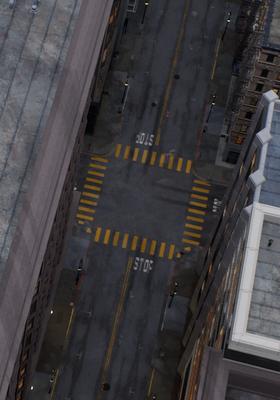}}
		\vspace{-0.3mm}
	\end{minipage}

     \begin{minipage}[t]{0.2\linewidth}
		\centering
		\centerline{\includegraphics[width=0.99\linewidth]{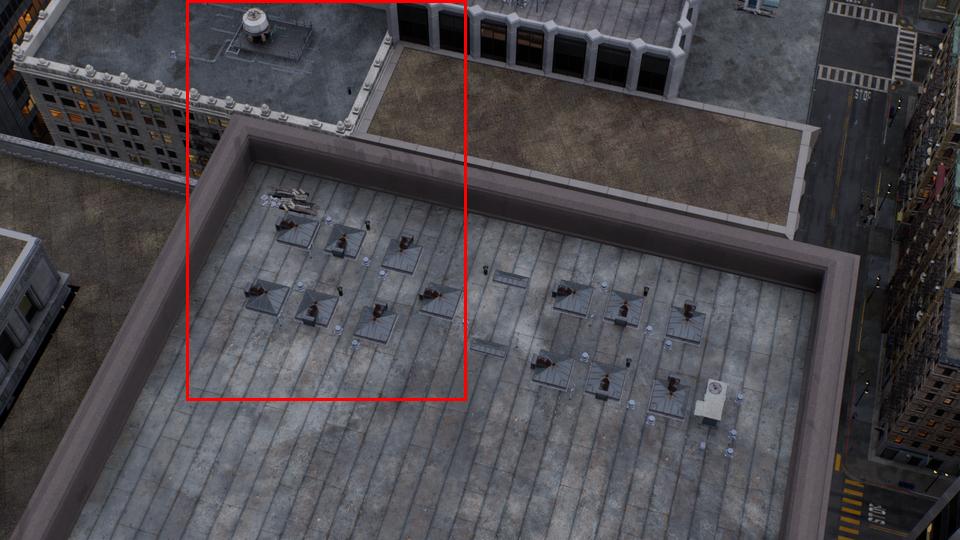}}
		\vspace{-0.3mm}
        \centerline{Block\_A}
	\end{minipage}
	\begin{minipage}[t]{0.15\linewidth}
		\centering
		\centerline{\includegraphics[width=0.99\linewidth]{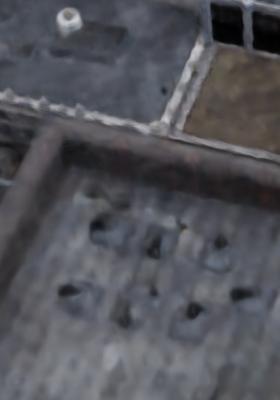}}
		\vspace{-0.3mm}
        \centerline{TensoRF}
	\end{minipage}
	\begin{minipage}[t]{0.15\linewidth}
		\centering
		\centerline{\includegraphics[width=0.99\linewidth]{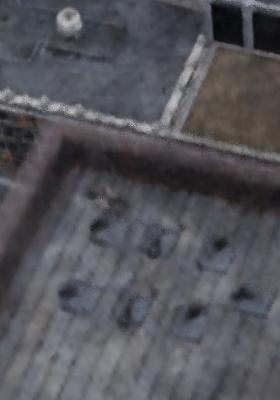}}
		\vspace{-0.3mm}
        \centerline{GridNeRF}
	\end{minipage}
	\begin{minipage}[t]{0.15\linewidth}
		\centering
		\centerline{\includegraphics[width=0.99\linewidth]{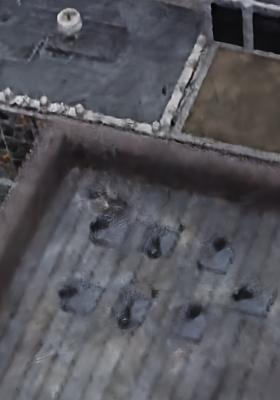}}
		\vspace{-0.3mm}
        \centerline{GridNeRF}
        \centerline{+NeRFLiX}
	\end{minipage}
    \begin{minipage}[t]{0.15\linewidth}
		\centering
		\centerline{\includegraphics[width=0.99\linewidth]{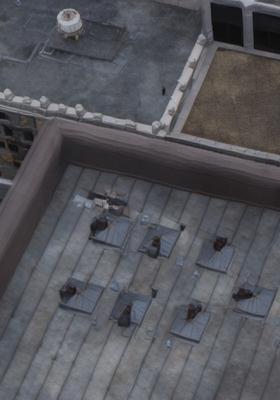}}
		\vspace{-0.3mm}
        \centerline{GridNeRF}
        \centerline{+\ours}
	\end{minipage}
	\begin{minipage}[t]{0.15\linewidth}
		\centering
		\centerline{\includegraphics[width=0.99\linewidth]{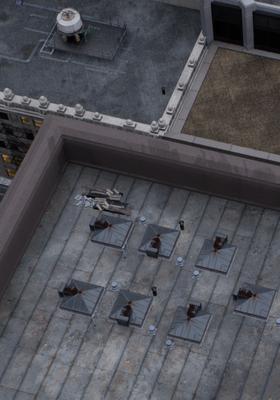}}
		\vspace{-0.3mm}
        \centerline{G.T.}
	\end{minipage}

    \vspace{+3mm}

        \begin{minipage}[t]{0.2\linewidth}
		\centering
		\centerline{\includegraphics[width=0.99\linewidth]{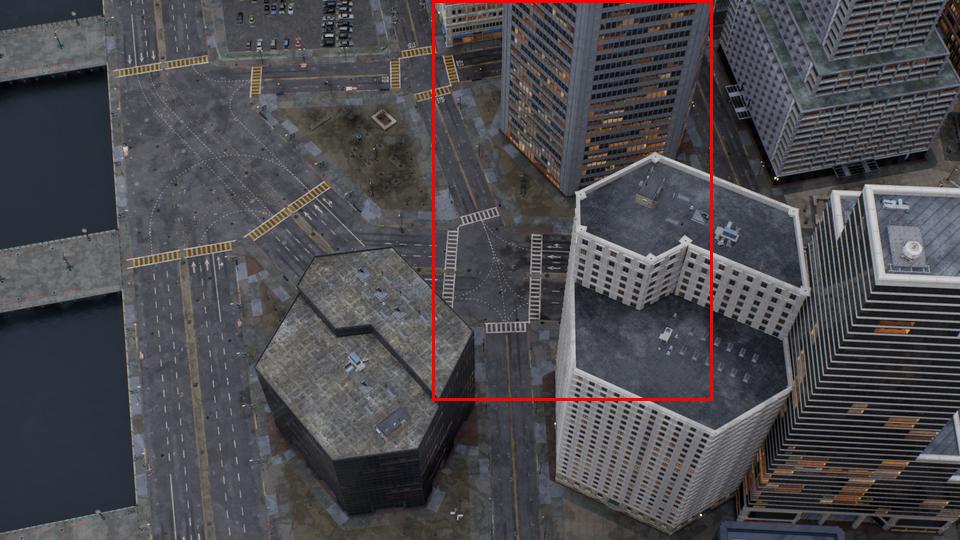}}
		\vspace{-0.3mm}
	\end{minipage}
	\begin{minipage}[t]{0.15\linewidth}
		\centering
		\centerline{\includegraphics[width=0.99\linewidth]{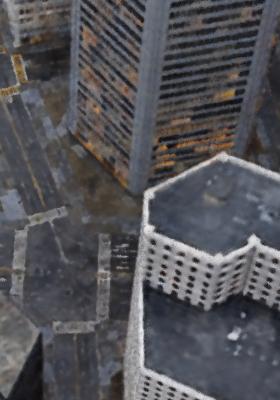}}
		\vspace{-0.3mm}
	\end{minipage}
	\begin{minipage}[t]{0.15\linewidth}
		\centering
		\centerline{\includegraphics[width=0.99\linewidth]{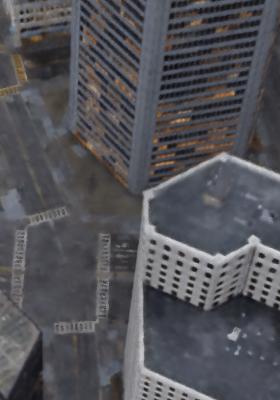}}
		\vspace{-0.3mm}
	\end{minipage}
	\begin{minipage}[t]{0.15\linewidth}
		\centering
		\centerline{\includegraphics[width=0.99\linewidth]{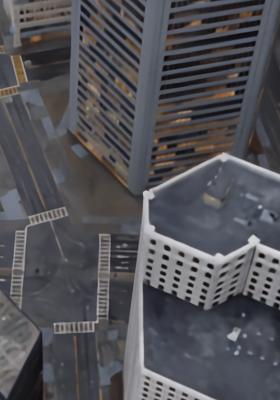}}
		\vspace{-0.3mm}
	\end{minipage}
    \begin{minipage}[t]{0.15\linewidth}
		\centering
		\centerline{\includegraphics[width=0.99\linewidth]{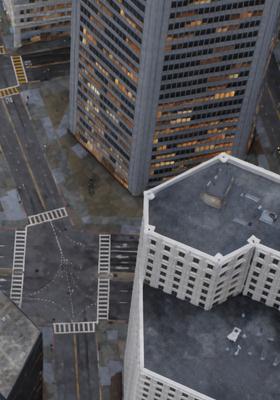}}
		\vspace{-0.3mm}
	\end{minipage}
	\begin{minipage}[t]{0.15\linewidth}
		\centering
		\centerline{\includegraphics[width=0.99\linewidth]{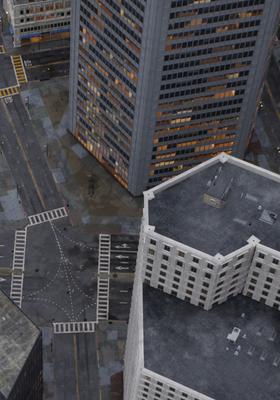}}
		\vspace{-0.3mm}
	\end{minipage}

     \begin{minipage}[t]{0.2\linewidth}
		\centering
		\centerline{\includegraphics[width=0.99\linewidth]{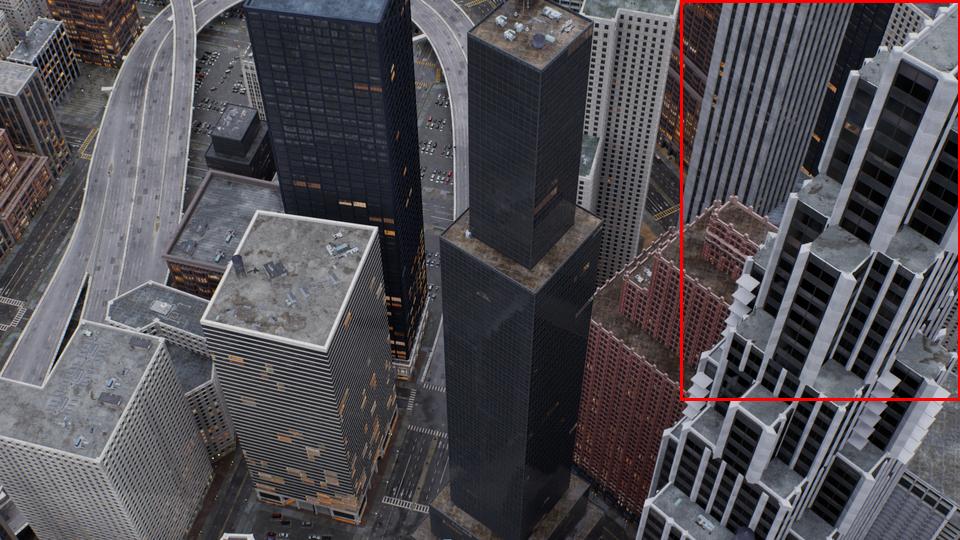}}
		\vspace{-0.3mm}
        \centerline{Block\_B}
	\end{minipage}
	\begin{minipage}[t]{0.15\linewidth}
		\centering
		\centerline{\includegraphics[width=0.99\linewidth]{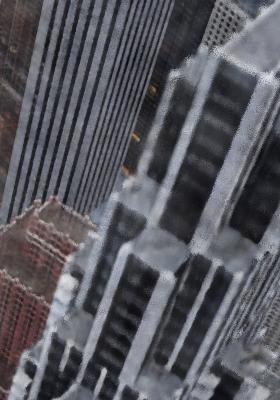}}
		\vspace{-0.3mm}
        \centerline{TensoRF}
	\end{minipage}
	\begin{minipage}[t]{0.15\linewidth}
		\centering
		\centerline{\includegraphics[width=0.99\linewidth]{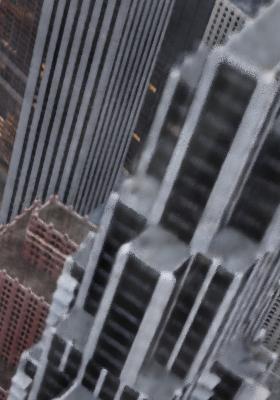}}
		\vspace{-0.3mm}
        \centerline{GridNeRF}
	\end{minipage}
	\begin{minipage}[t]{0.15\linewidth}
		\centering
		\centerline{\includegraphics[width=0.99\linewidth]{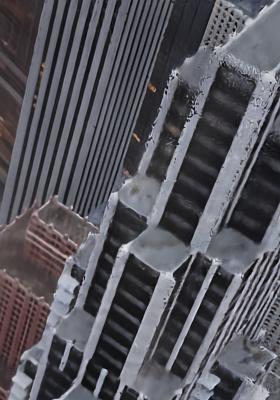}}
		\vspace{-0.3mm}
        \centerline{GridNeRF}
        \centerline{+NeRFLiX}
	\end{minipage}
    \begin{minipage}[t]{0.15\linewidth}
		\centering
		\centerline{\includegraphics[width=0.99\linewidth]{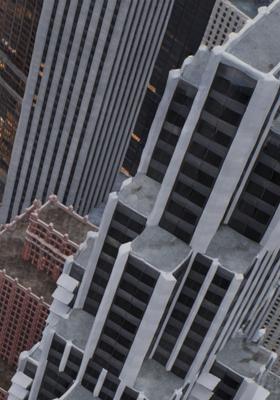}}
		\vspace{-0.3mm}
        \centerline{GridNeRF}
        \centerline{+\ours}
	\end{minipage}
	\begin{minipage}[t]{0.15\linewidth}
		\centering
		\centerline{\includegraphics[width=0.99\linewidth]{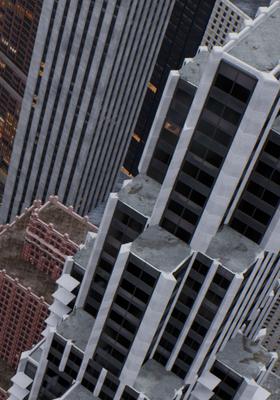}}
		\vspace{-0.3mm}
        \centerline{G.T.}
	\end{minipage}

    \vspace{+3mm}

        \begin{minipage}[t]{0.2\linewidth}
		\centering
		\centerline{\includegraphics[width=0.99\linewidth]{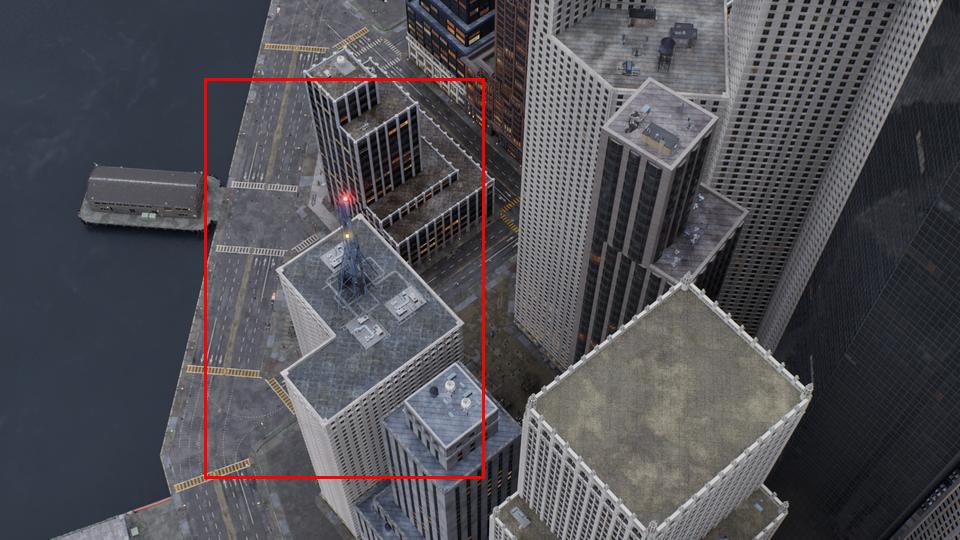}}
		\vspace{-0.3mm}
	\end{minipage}
	\begin{minipage}[t]{0.15\linewidth}
		\centering
		\centerline{\includegraphics[width=0.99\linewidth]{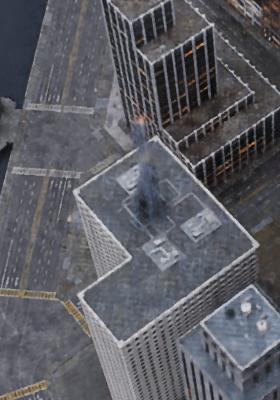}}
		\vspace{-0.3mm}
	\end{minipage}
	\begin{minipage}[t]{0.15\linewidth}
		\centering
		\centerline{\includegraphics[width=0.99\linewidth]{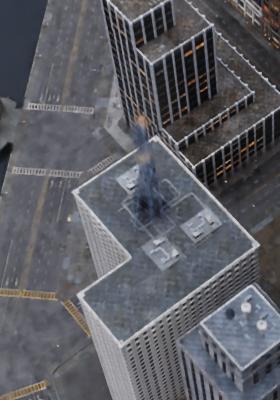}}
		\vspace{-0.3mm}
	\end{minipage}
	\begin{minipage}[t]{0.15\linewidth}
		\centering
		\centerline{\includegraphics[width=0.99\linewidth]{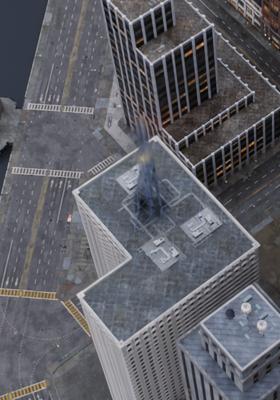}}
		\vspace{-0.3mm}
	\end{minipage}
    \begin{minipage}[t]{0.15\linewidth}
		\centering
		\centerline{\includegraphics[width=0.99\linewidth]{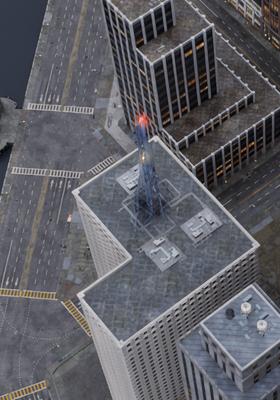}}
		\vspace{-0.3mm}
	\end{minipage}
	\begin{minipage}[t]{0.15\linewidth}
		\centering
		\centerline{\includegraphics[width=0.99\linewidth]{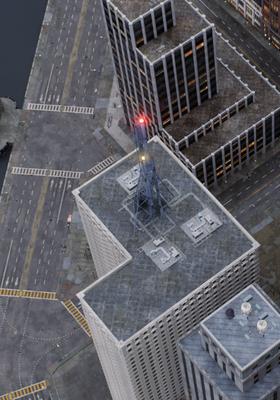}}
		\vspace{-0.3mm}
	\end{minipage}

     \begin{minipage}[t]{0.2\linewidth}
		\centering
		\centerline{\includegraphics[width=0.99\linewidth]{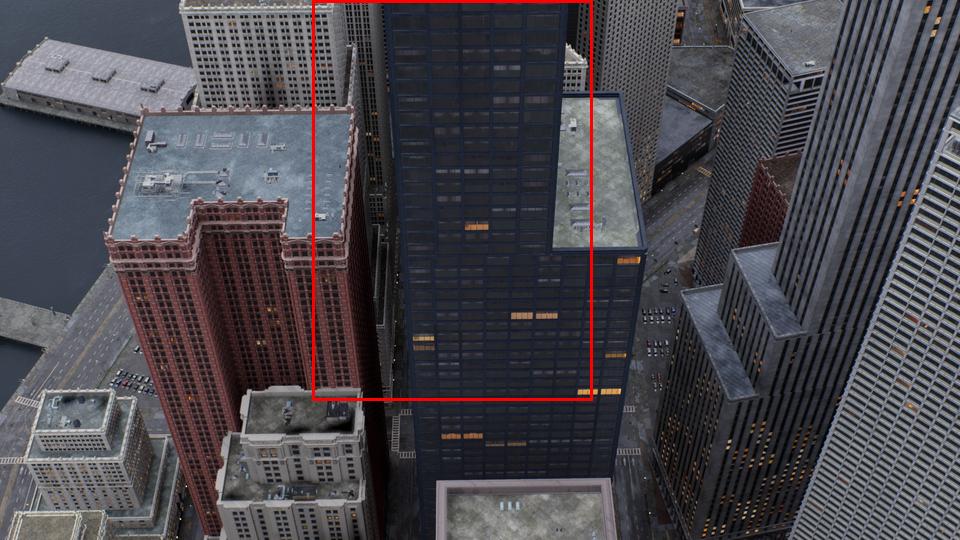}}
		\vspace{-0.3mm}
        \centerline{Block\_C}
	\end{minipage}
	\begin{minipage}[t]{0.15\linewidth}
		\centering
		\centerline{\includegraphics[width=0.99\linewidth]{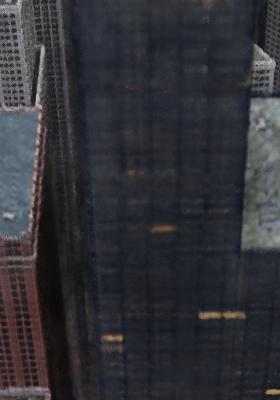}}
		\vspace{-0.3mm}
        \centerline{TensoRF}
	\end{minipage}
	\begin{minipage}[t]{0.15\linewidth}
		\centering
		\centerline{\includegraphics[width=0.99\linewidth]{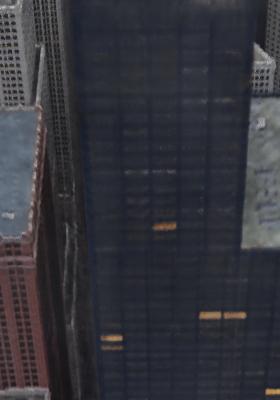}}
		\vspace{-0.3mm}
        \centerline{GridNeRF}
	\end{minipage}
	\begin{minipage}[t]{0.15\linewidth}
		\centering
		\centerline{\includegraphics[width=0.99\linewidth]{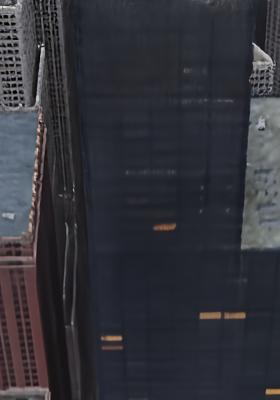}}
		\vspace{-0.3mm}
        \centerline{GridNeRF}
        \centerline{+NeRFLiX}
	\end{minipage}
    \begin{minipage}[t]{0.15\linewidth}
		\centering
		\centerline{\includegraphics[width=0.99\linewidth]{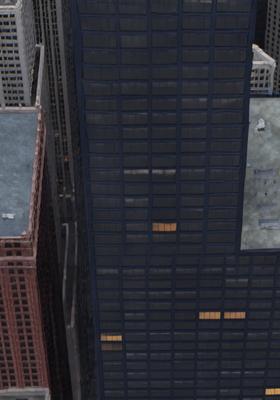}}
		\vspace{-0.3mm}
        \centerline{GridNeRF}
        \centerline{+\ours}
	\end{minipage}
	\begin{minipage}[t]{0.15\linewidth}
		\centering
		\centerline{\includegraphics[width=0.99\linewidth]{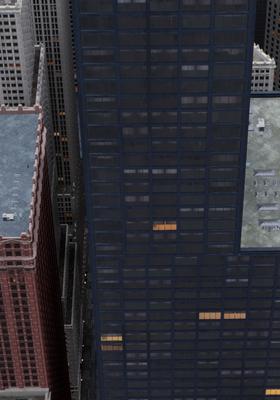}}
		\vspace{-0.3mm}
        \centerline{G.T.}
	\end{minipage}

\caption{More qualitative comparisons for aerial scenes in MatrixCity~\cite{matrixcity} dataset (Part 1). From top to down: \textit{Block\_A}, \textit{Block\_B} and \textit{Block\_C}.}
\label{fig:morematrix1}
\vspace{-2mm}
\end{figure*}

\begin{figure*}[t]
	\small
	\centering
    \begin{minipage}[t]{0.2\linewidth}
		\centering
		\centerline{\includegraphics[width=0.99\linewidth]{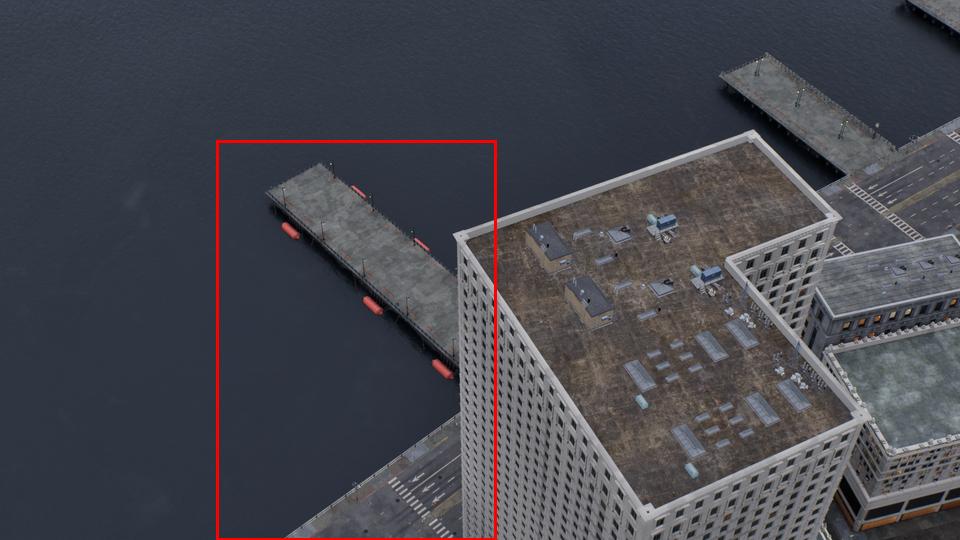}}
		\vspace{-0.3mm}
	\end{minipage}
	\begin{minipage}[t]{0.15\linewidth}
		\centering
		\centerline{\includegraphics[width=0.99\linewidth]{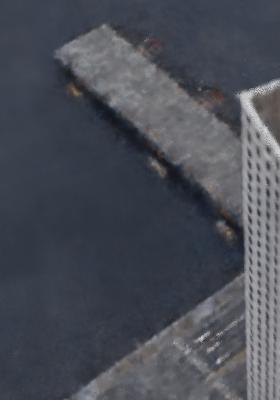}}
		\vspace{-0.3mm}
	\end{minipage}
	\begin{minipage}[t]{0.15\linewidth}
		\centering
		\centerline{\includegraphics[width=0.99\linewidth]{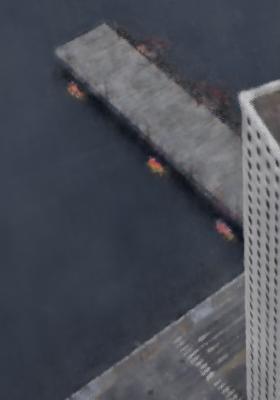}}
		\vspace{-0.3mm}
	\end{minipage}
	\begin{minipage}[t]{0.15\linewidth}
		\centering
		\centerline{\includegraphics[width=0.99\linewidth]{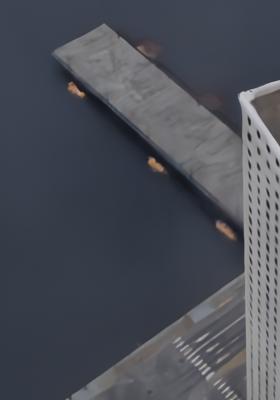}}
		\vspace{-0.3mm}
	\end{minipage}
    \begin{minipage}[t]{0.15\linewidth}
		\centering
		\centerline{\includegraphics[width=0.99\linewidth]{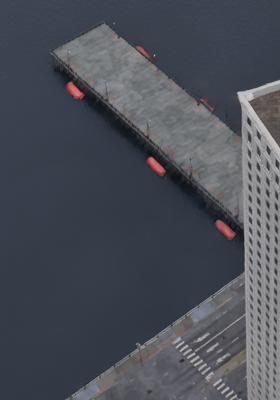}}
		\vspace{-0.3mm}
	\end{minipage}
	\begin{minipage}[t]{0.15\linewidth}
		\centering
		\centerline{\includegraphics[width=0.99\linewidth]{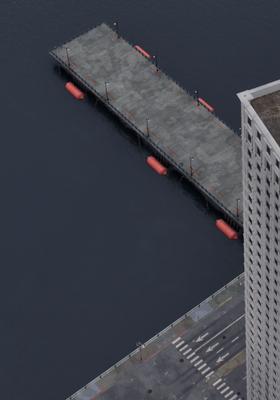}}
		\vspace{-0.3mm}
	\end{minipage}

     \begin{minipage}[t]{0.2\linewidth}
		\centering
		\centerline{\includegraphics[width=0.99\linewidth]{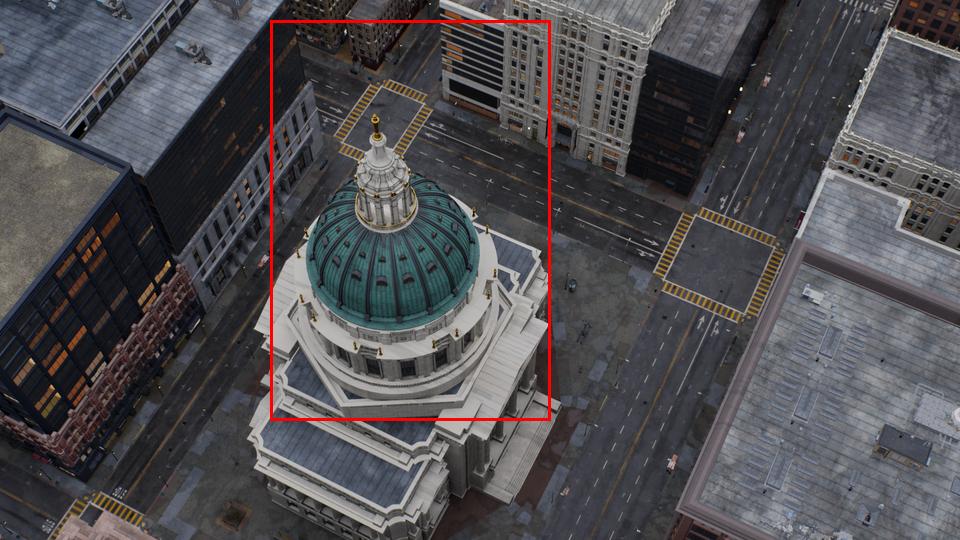}}
		\vspace{-0.3mm}
        \centerline{Block\_D}
	\end{minipage}
	\begin{minipage}[t]{0.15\linewidth}
		\centering
		\centerline{\includegraphics[width=0.99\linewidth]{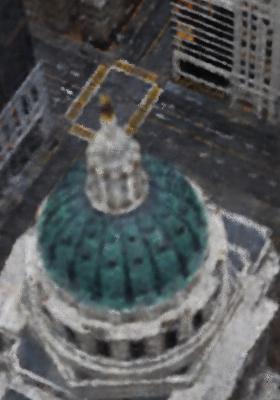}}
		\vspace{-0.3mm}
        \centerline{TensoRF}
	\end{minipage}
	\begin{minipage}[t]{0.15\linewidth}
		\centering
		\centerline{\includegraphics[width=0.99\linewidth]{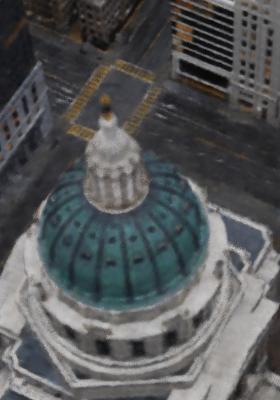}}
		\vspace{-0.3mm}
        \centerline{GridNeRF}
	\end{minipage}
	\begin{minipage}[t]{0.15\linewidth}
		\centering
		\centerline{\includegraphics[width=0.99\linewidth]{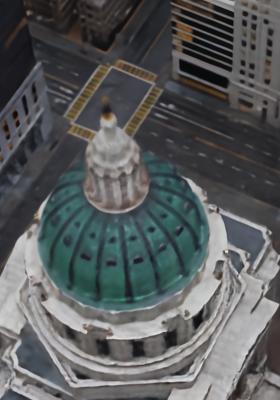}}
		\vspace{-0.3mm}
        \centerline{GridNeRF}
        \centerline{+NeRFLiX}
	\end{minipage}
    \begin{minipage}[t]{0.15\linewidth}
		\centering
		\centerline{\includegraphics[width=0.99\linewidth]{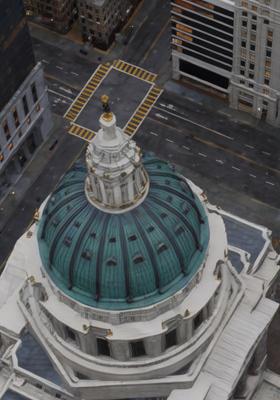}}
		\vspace{-0.3mm}
        \centerline{GridNeRF}
        \centerline{+\ours}
	\end{minipage}
	\begin{minipage}[t]{0.15\linewidth}
		\centering
		\centerline{\includegraphics[width=0.99\linewidth]{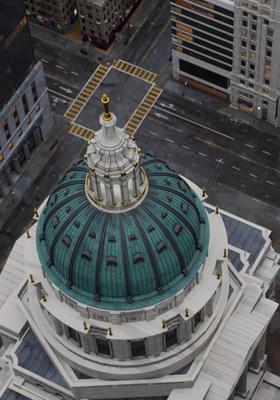}}
		\vspace{-0.3mm}
        \centerline{G.T.}
	\end{minipage}

    \vspace{+3mm}

        \begin{minipage}[t]{0.2\linewidth}
		\centering
		\centerline{\includegraphics[width=0.99\linewidth]{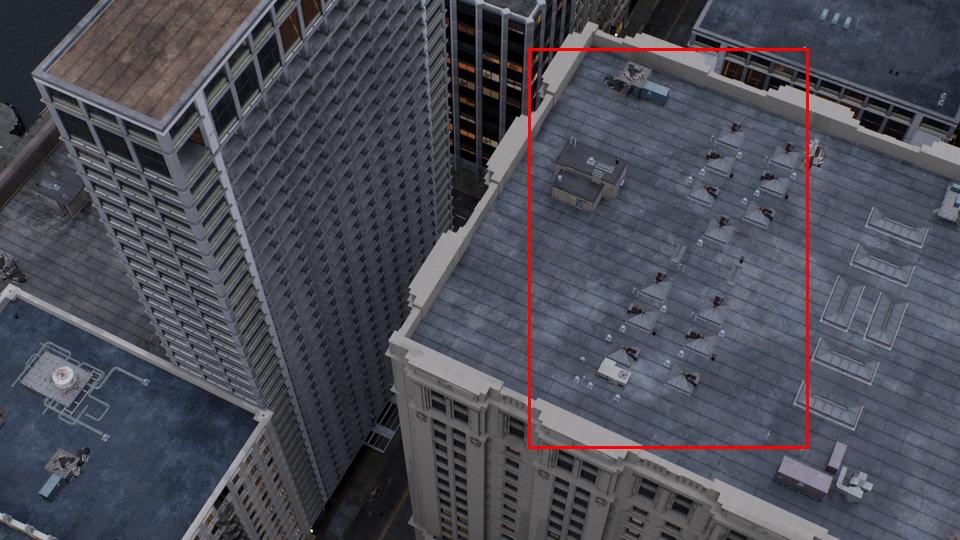}}
		\vspace{-0.3mm}
	\end{minipage}
	\begin{minipage}[t]{0.15\linewidth}
		\centering
		\centerline{\includegraphics[width=0.99\linewidth]{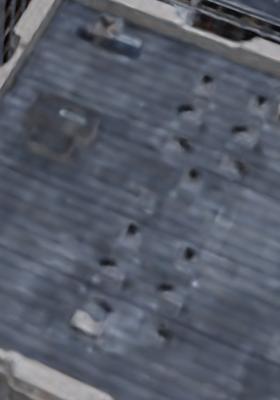}}
		\vspace{-0.3mm}
	\end{minipage}
	\begin{minipage}[t]{0.15\linewidth}
		\centering
		\centerline{\includegraphics[width=0.99\linewidth]{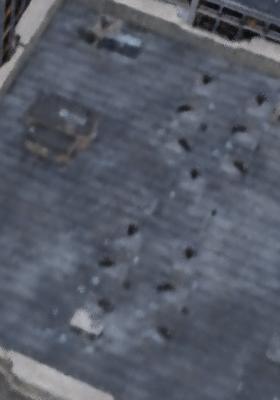}}
		\vspace{-0.3mm}
	\end{minipage}
	\begin{minipage}[t]{0.15\linewidth}
		\centering
		\centerline{\includegraphics[width=0.99\linewidth]{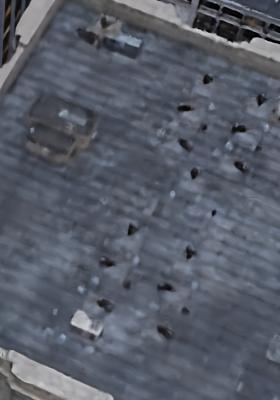}}
		\vspace{-0.3mm}
	\end{minipage}
    \begin{minipage}[t]{0.15\linewidth}
		\centering
		\centerline{\includegraphics[width=0.99\linewidth]{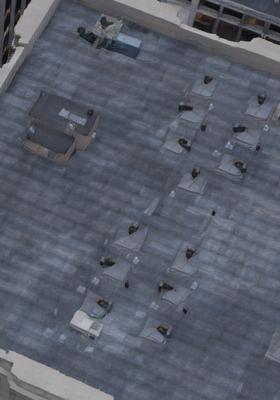}}
		\vspace{-0.3mm}
	\end{minipage}
	\begin{minipage}[t]{0.15\linewidth}
		\centering
		\centerline{\includegraphics[width=0.99\linewidth]{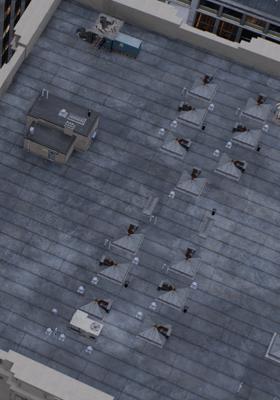}}
		\vspace{-0.3mm}
	\end{minipage}

     \begin{minipage}[t]{0.2\linewidth}
		\centering
		\centerline{\includegraphics[width=0.99\linewidth]{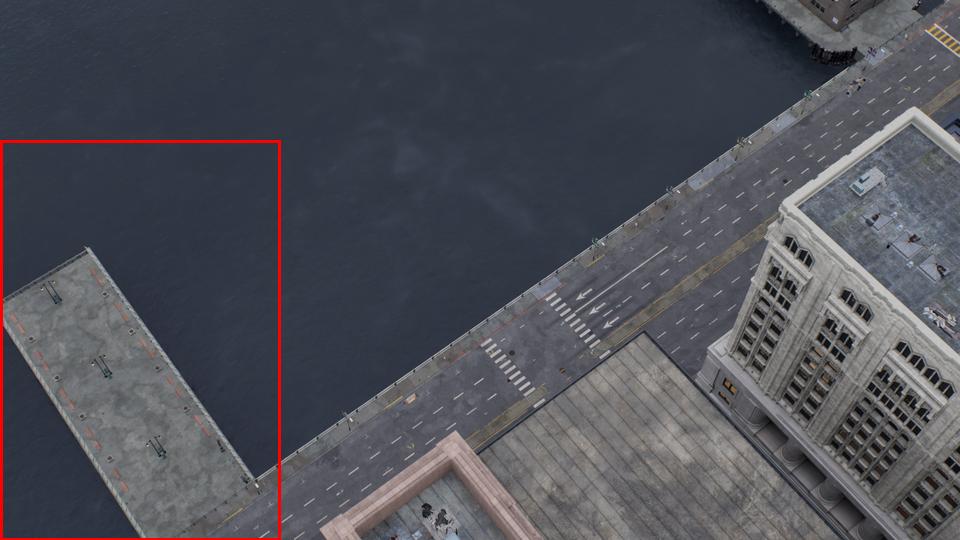}}
		\vspace{-0.3mm}
        \centerline{Block\_E}
	\end{minipage}
	\begin{minipage}[t]{0.15\linewidth}
		\centering
		\centerline{\includegraphics[width=0.99\linewidth]{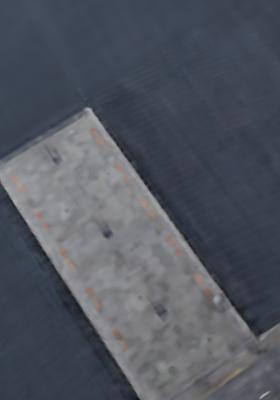}}
		\vspace{-0.3mm}
        \centerline{TensoRF}
	\end{minipage}
	\begin{minipage}[t]{0.15\linewidth}
		\centering
		\centerline{\includegraphics[width=0.99\linewidth]{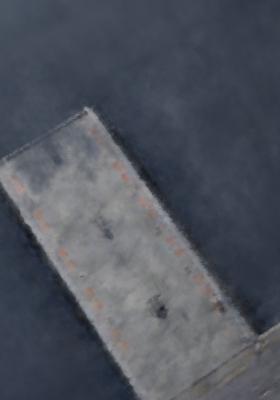}}
		\vspace{-0.3mm}
        \centerline{GridNeRF}
	\end{minipage}
	\begin{minipage}[t]{0.15\linewidth}
		\centering
		\centerline{\includegraphics[width=0.99\linewidth]{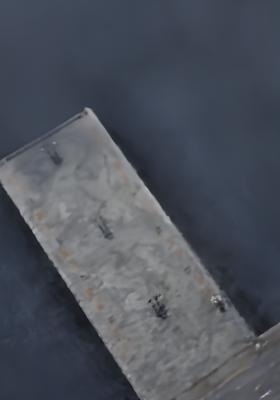}}
		\vspace{-0.3mm}
        \centerline{GridNeRF}
        \centerline{+NeRFLiX}
	\end{minipage}
    \begin{minipage}[t]{0.15\linewidth}
		\centering
		\centerline{\includegraphics[width=0.99\linewidth]{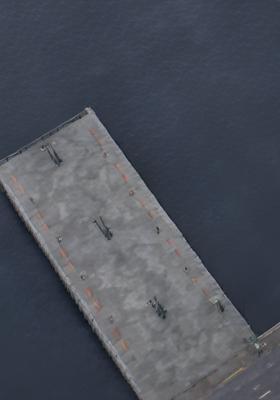}}
		\vspace{-0.3mm}
        \centerline{GridNeRF}
        \centerline{+\ours}
	\end{minipage}
	\begin{minipage}[t]{0.15\linewidth}
		\centering
		\centerline{\includegraphics[width=0.99\linewidth]{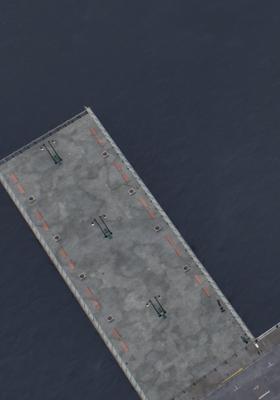}}
		\vspace{-0.3mm}
        \centerline{G.T.}
	\end{minipage}

    \vspace{+3mm}
     
    \begin{minipage}[t]{0.2\linewidth}
		\centering
		\centerline{\includegraphics[width=0.99\linewidth]{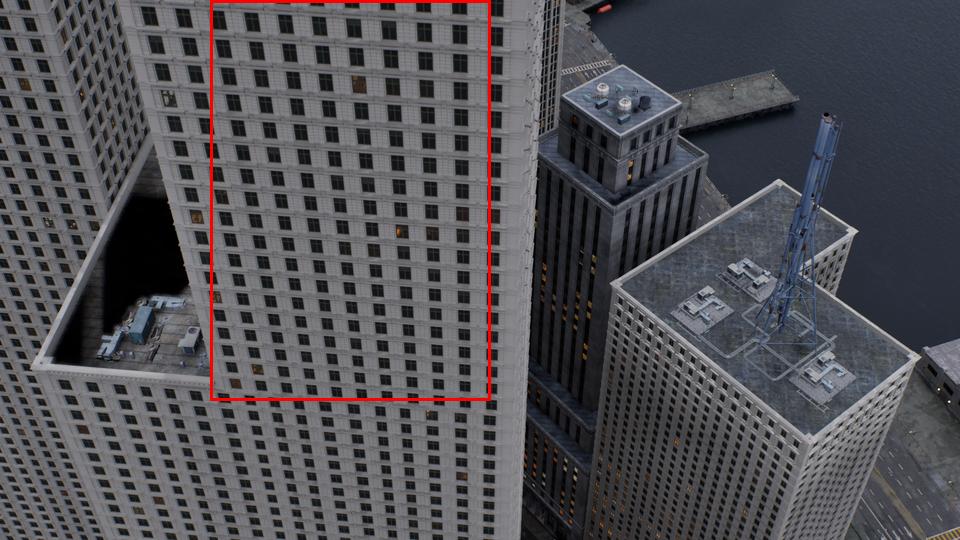}}
		\vspace{-0.3mm}
	\end{minipage}
	\begin{minipage}[t]{0.15\linewidth}
		\centering
		\centerline{\includegraphics[width=0.99\linewidth]{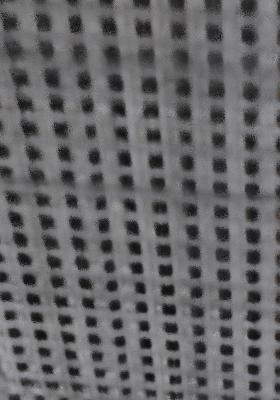}}
		\vspace{-0.3mm}
	\end{minipage}
	\begin{minipage}[t]{0.15\linewidth}
		\centering
		\centerline{\includegraphics[width=0.99\linewidth]{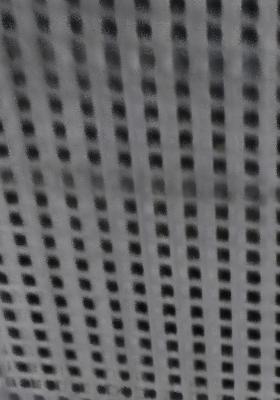}}
		\vspace{-0.3mm}
	\end{minipage}
	\begin{minipage}[t]{0.15\linewidth}
		\centering
		\centerline{\includegraphics[width=0.99\linewidth]{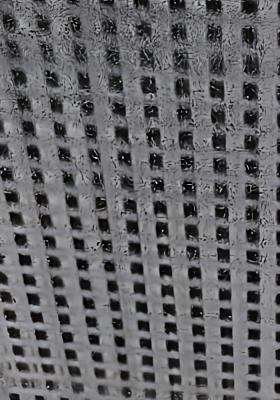}}
		\vspace{-0.3mm}
	\end{minipage}
    \begin{minipage}[t]{0.15\linewidth}
		\centering
		\centerline{\includegraphics[width=0.99\linewidth]{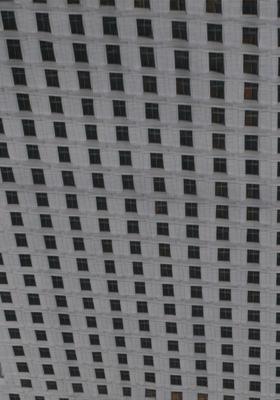}}
		\vspace{-0.3mm}
	\end{minipage}
	\begin{minipage}[t]{0.15\linewidth}
		\centering
		\centerline{\includegraphics[width=0.99\linewidth]{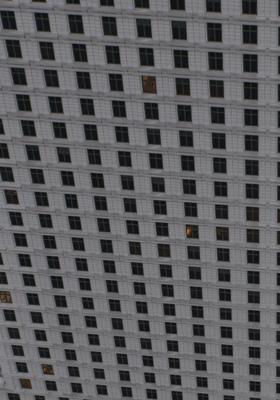}}
		\vspace{-0.3mm}
	\end{minipage}

     \begin{minipage}[t]{0.2\linewidth}
		\centering
		\centerline{\includegraphics[width=0.99\linewidth]{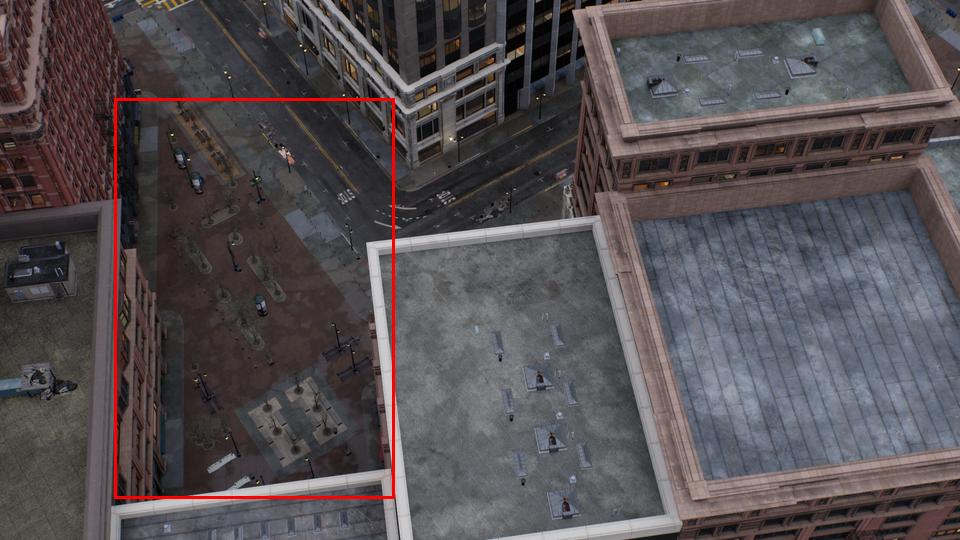}}
		\vspace{-0.3mm}
        \centerline{Block\_ALL}
	\end{minipage}
	\begin{minipage}[t]{0.15\linewidth}
		\centering
		\centerline{\includegraphics[width=0.99\linewidth]{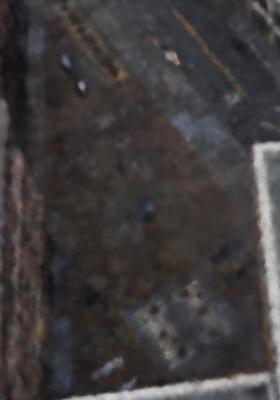}}
		\vspace{-0.3mm}
        \centerline{TensoRF}
	\end{minipage}
	\begin{minipage}[t]{0.15\linewidth}
		\centering
		\centerline{\includegraphics[width=0.99\linewidth]{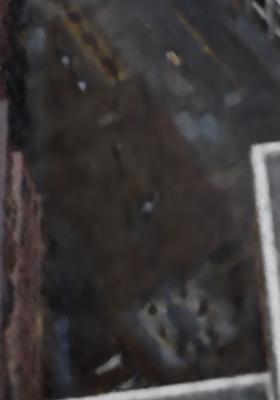}}
		\vspace{-0.3mm}
        \centerline{GridNeRF}
	\end{minipage}
	\begin{minipage}[t]{0.15\linewidth}
		\centering
		\centerline{\includegraphics[width=0.99\linewidth]{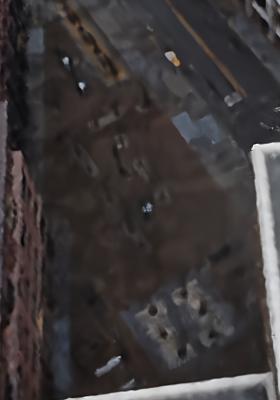}}
		\vspace{-0.3mm}
        \centerline{GridNeRF}
        \centerline{+NeRFLiX}
	\end{minipage}
    \begin{minipage}[t]{0.15\linewidth}
		\centering
		\centerline{\includegraphics[width=0.99\linewidth]{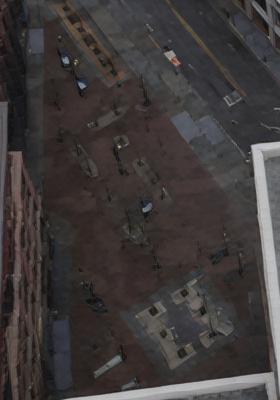}}
		\vspace{-0.3mm}
        \centerline{GridNeRF}
        \centerline{+\ours}
	\end{minipage}
	\begin{minipage}[t]{0.15\linewidth}
		\centering
		\centerline{\includegraphics[width=0.99\linewidth]{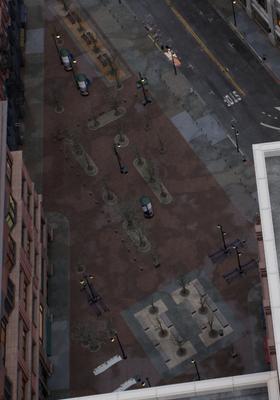}}
		\vspace{-0.3mm}
        \centerline{G.T.}
	\end{minipage}

\caption{More qualitative comparisons for aerial scenes in MatrixCity~\cite{matrixcity} dataset (Part 2). From top to down: \textit{Block\_D}, \textit{Block\_E} and \textit{Block\_ALL}.}
\label{fig:morematrix2}
\vspace{-2mm}
\end{figure*}

The breakdown numerical results for each aerial scene in MatrixCity~\cite{matrixcity} dataset corresponding to average results in Table \ref{table:matrixcity} is presented in Table \ref{table:matrix_per_scene}. And the additional visual comparisons for each city block are located in Figure \ref{fig:morematrix1} and Figure \ref{fig:morematrix2}. We can observe that the proposed \ours~consistently produces much more visually appealing finer details, such as clearer street lines, buildings and roofs, while significantly reduces aliasing artifacts, blurs and floaters. For a more intuitive visual comparison, we also include a video demo in the supplementary material.

In Figure \ref{fig:genvsspe}, we show some visual comparisons between \ours$_{gen}$ and \ours$_{spe}$. We can observe that \ours$_{gen}$ could achieve comparable visual enhancements as \ours$_{spe}$ in most of the scene scenarios (row 1-2). For a few rare scene-specific degradations, applying a scene-specific diffusion model \ours$_{spe}$ performs better restoration effects (row 3).

\subsection{More Results for the Unbounded 360-degree Scenes}
\label{more results for 360}
The detailed numerical results for each scene in MipNeRF-360~\cite{miperf360} dataset corresponding to average results in Table \ref{table:360} could be found in Table \ref{table:360_per_scene}. We compare against NeRF~\cite{nerf}, MipNeRF~\cite{mipnerf}, NeRF++~\cite{nerf++}, Deep Blending~\cite{deepblend},
Instant-NGP~\cite{instantngp}, MipNeRF-360~\cite{miperf360}, ZipNeRF~\cite{zipnerf} and ZipNeRF+NeRFLiX~\cite{nerflix}. Regardless of the specific contents within the scene, \ours~is significantly beneficial for improving distortion-based metric like PSNR, structure-based metric like SSIM, as well as perceptual-based metric like LPIPS.

\begin{figure*}[t]
	\small
	\centering

     \begin{minipage}[t]{0.2\linewidth}
		\centering
		\centerline{\includegraphics[width=0.99\linewidth]{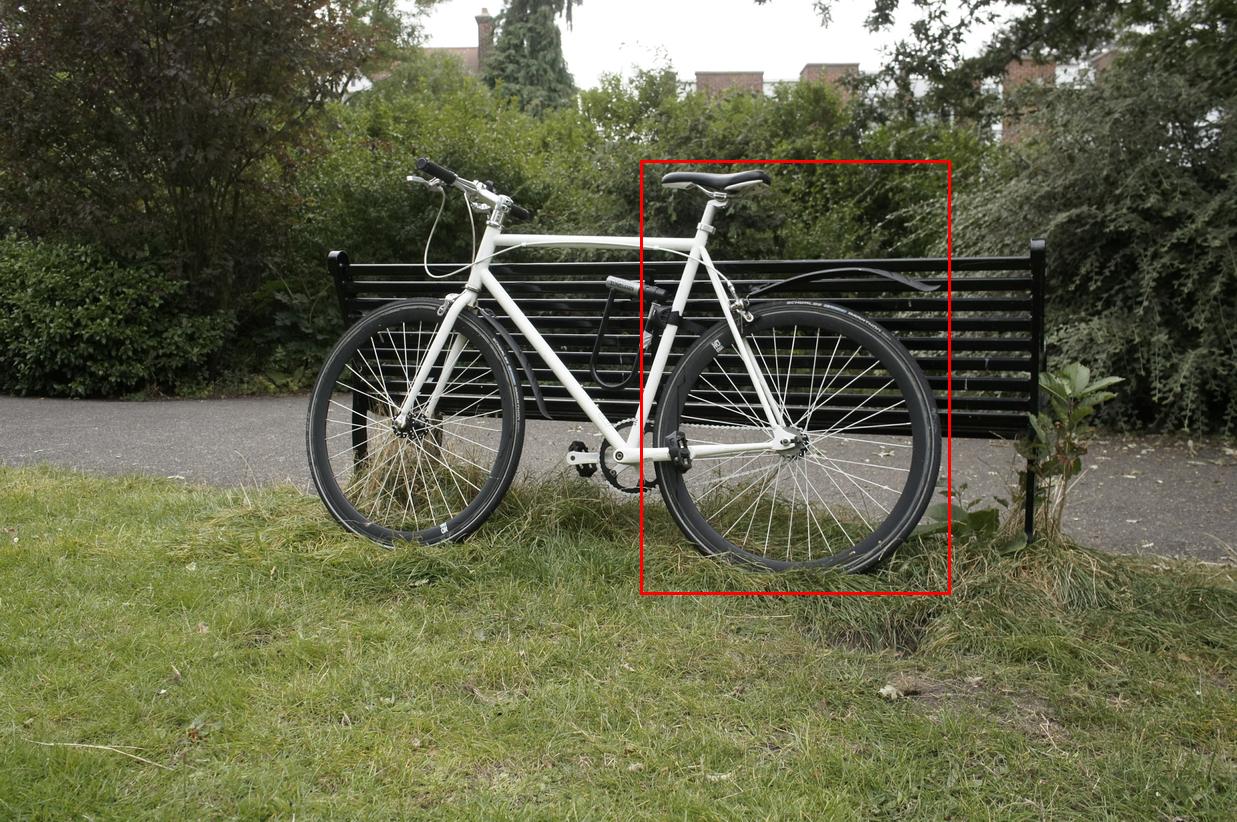}}
		\vspace{-0.3mm}
	\end{minipage}
	\begin{minipage}[t]{0.15\linewidth}
		\centering
		\centerline{\includegraphics[width=0.99\linewidth]{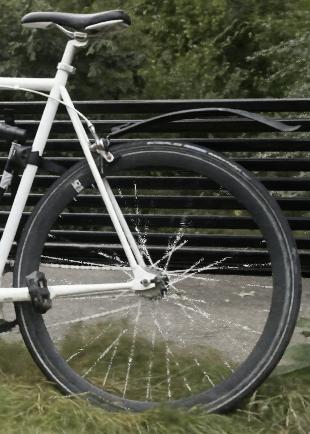}}
		\vspace{-0.3mm}
	\end{minipage}
	\begin{minipage}[t]{0.15\linewidth}
		\centering
		\centerline{\includegraphics[width=0.99\linewidth]{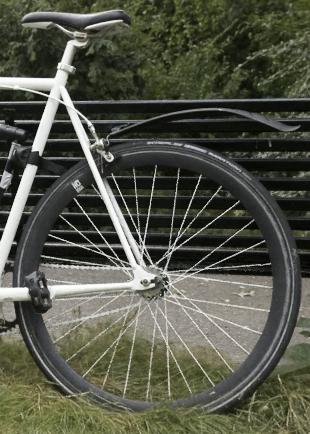}}
		\vspace{-0.3mm}
	\end{minipage}
    \begin{minipage}[t]{0.15\linewidth}
		\centering
		\centerline{\includegraphics[width=0.99\linewidth]{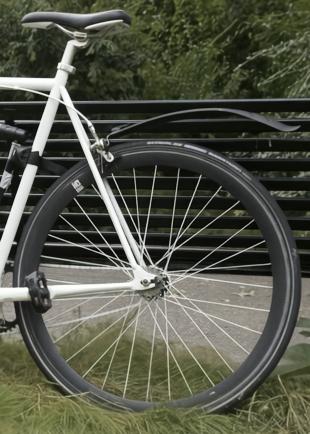}}
		\vspace{-0.3mm}
	\end{minipage}
	\begin{minipage}[t]{0.15\linewidth}
		\centering
		\centerline{\includegraphics[width=0.99\linewidth]{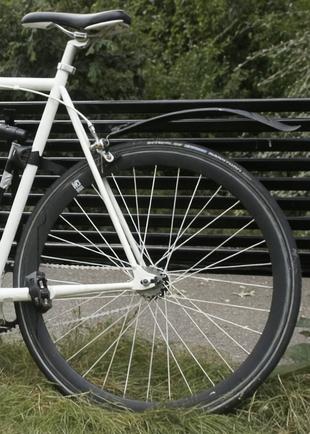}}
		\vspace{-0.3mm}
	\end{minipage}
	\begin{minipage}[t]{0.15\linewidth}
		\centering
		\centerline{\includegraphics[width=0.99\linewidth]{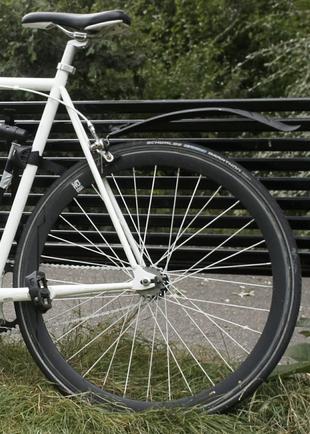}}
		\vspace{-0.3mm}
	\end{minipage}

    \begin{minipage}[t]{0.2\linewidth}
	\centering
		\centerline{\includegraphics[width=0.99\linewidth]{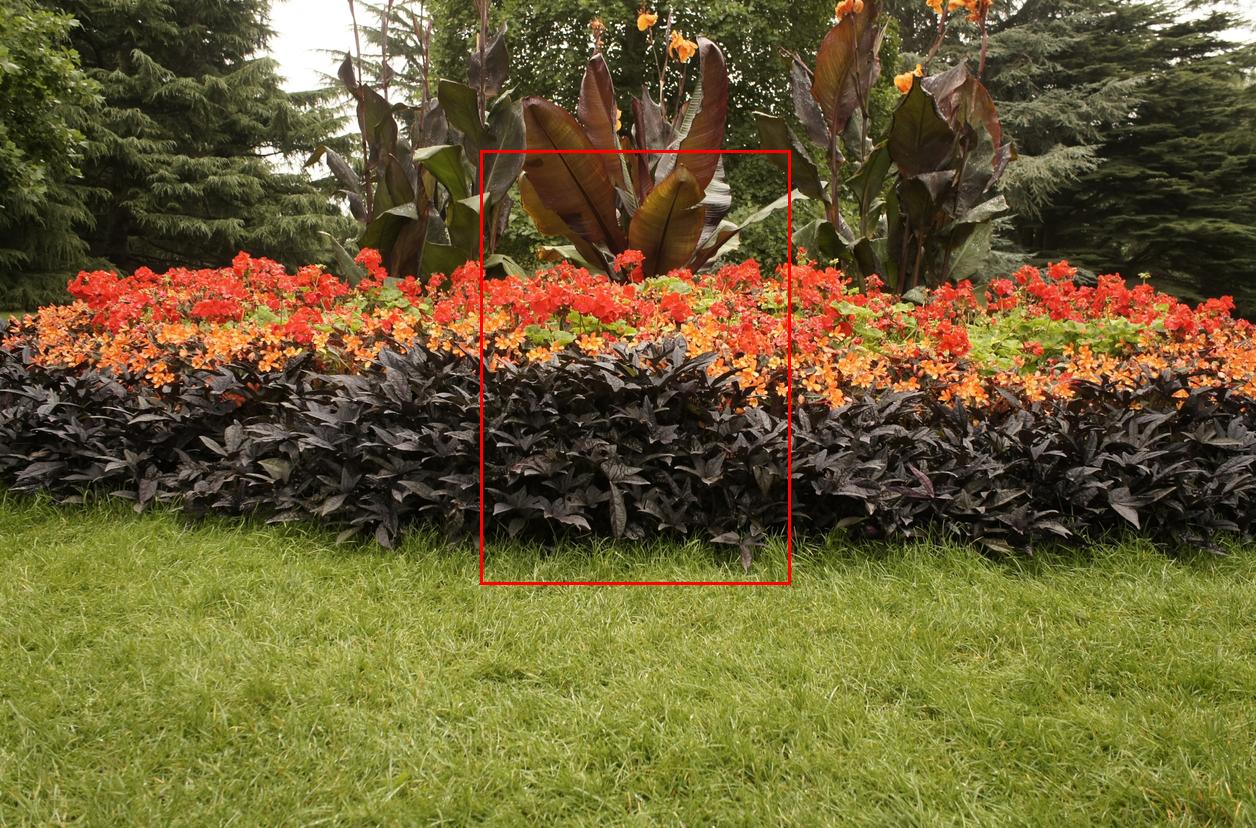}}
		\vspace{-0.3mm}
	\end{minipage}
	\begin{minipage}[t]{0.15\linewidth}
		\centering
		\centerline{\includegraphics[width=0.99\linewidth]{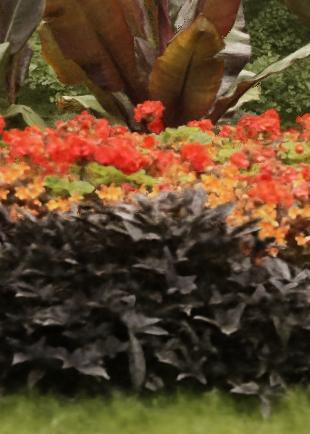}}
		\vspace{-0.3mm}
	\end{minipage}
	\begin{minipage}[t]{0.15\linewidth}
		\centering
		\centerline{\includegraphics[width=0.99\linewidth]{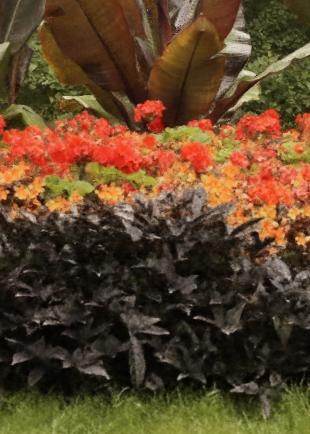}}
		\vspace{-0.3mm}
	\end{minipage}
	\begin{minipage}[t]{0.15\linewidth}
		\centering
		\centerline{\includegraphics[width=0.99\linewidth]{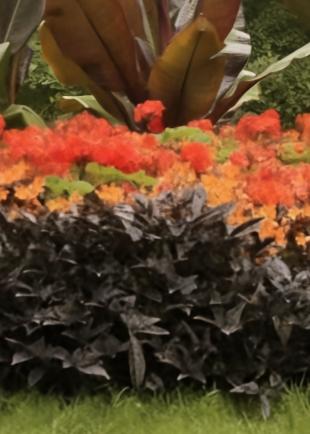}}
		\vspace{-0.3mm}
	\end{minipage}
    \begin{minipage}[t]{0.15\linewidth}
		\centering
		\centerline{\includegraphics[width=0.99\linewidth]{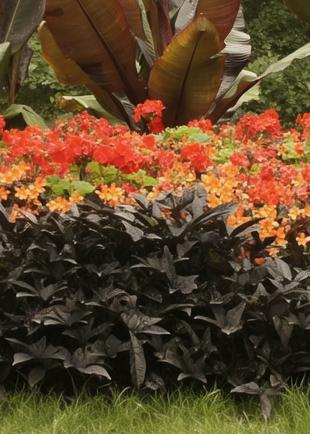}}
		\vspace{-0.3mm}
	\end{minipage}
	\begin{minipage}[t]{0.15\linewidth}
		\centering
		\centerline{\includegraphics[width=0.99\linewidth]{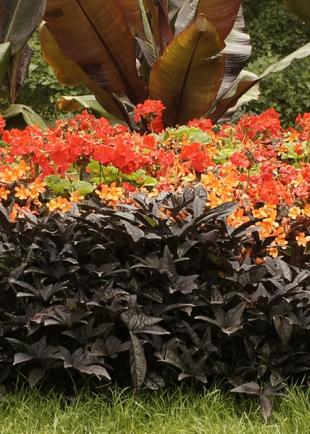}}
		\vspace{-0.3mm}
	\end{minipage}

	\begin{minipage}[t]{0.2\linewidth}
	\centering
		\centerline{\includegraphics[width=0.99\linewidth]{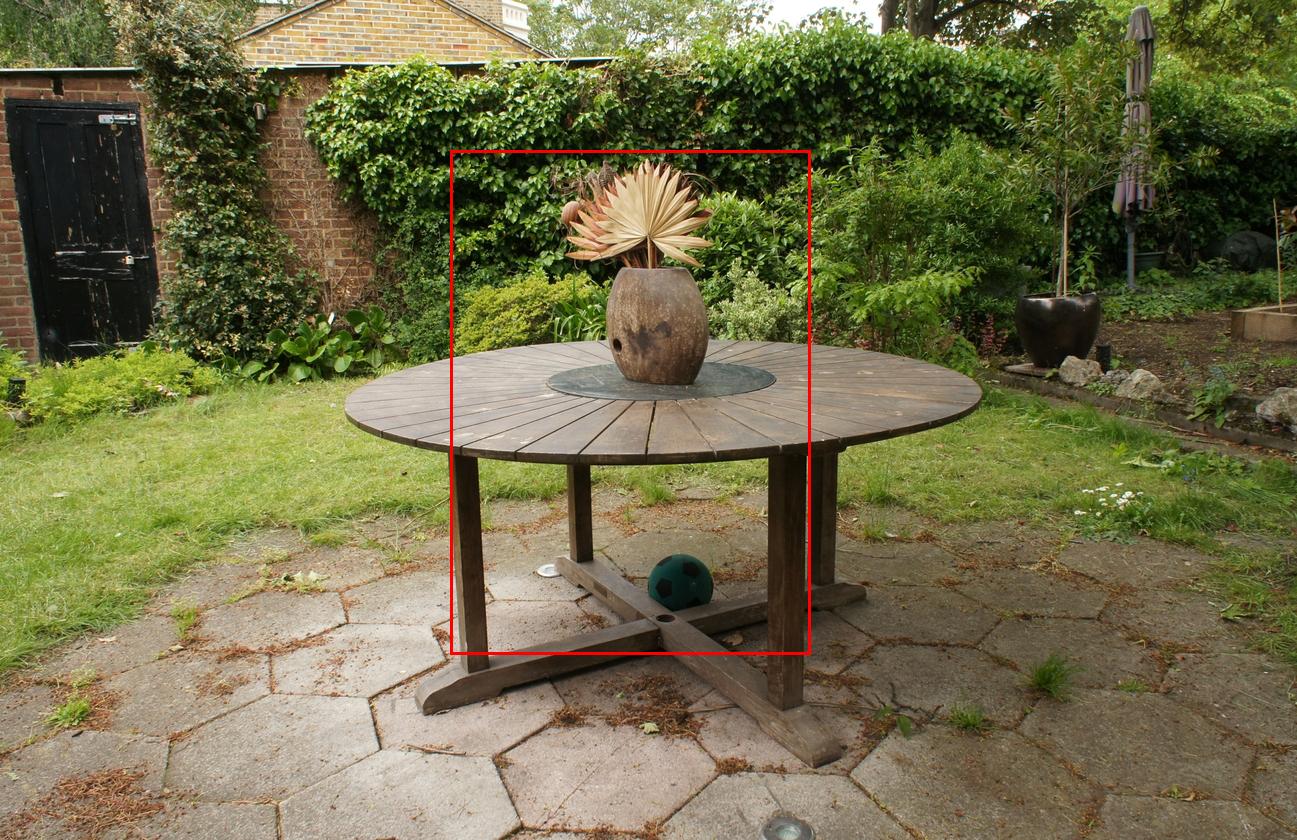}}
		\vspace{-0.3mm}
	\end{minipage}
	\begin{minipage}[t]{0.15\linewidth}
		\centering
		\centerline{\includegraphics[width=0.99\linewidth]{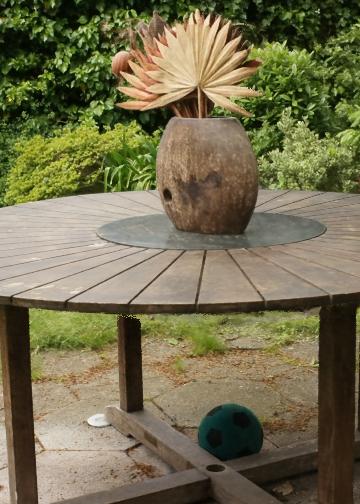}}
		\vspace{-0.3mm}
	\end{minipage}
	\begin{minipage}[t]{0.15\linewidth}
		\centering
		\centerline{\includegraphics[width=0.99\linewidth]{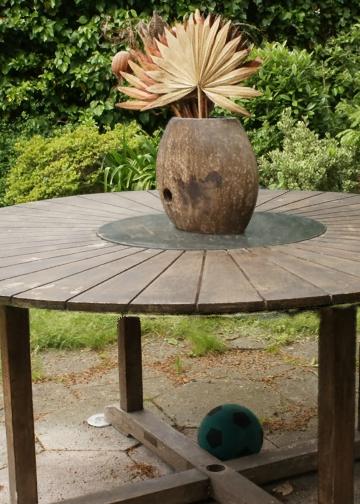}}
		\vspace{-0.3mm}
	\end{minipage}
    \begin{minipage}[t]{0.15\linewidth}
		\centering
		\centerline{\includegraphics[width=0.99\linewidth]{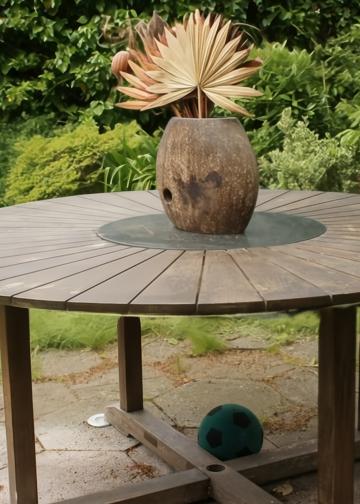}}
		\vspace{-0.3mm}
	\end{minipage}
	\begin{minipage}[t]{0.15\linewidth}
		\centering
		\centerline{\includegraphics[width=0.99\linewidth]{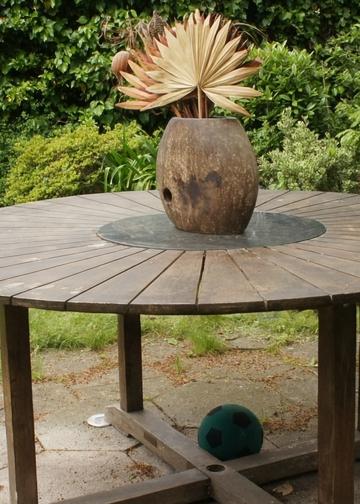}}
		\vspace{-0.3mm}
	\end{minipage}
	\begin{minipage}[t]{0.15\linewidth}
		\centering
    \centerline{\includegraphics[width=0.99\linewidth]{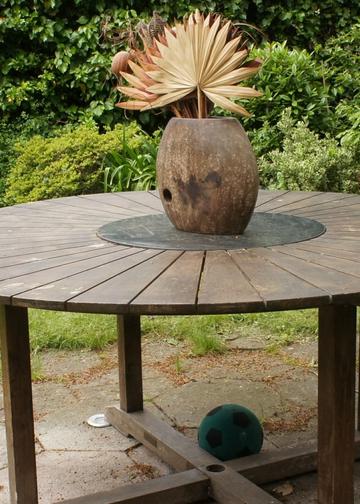}}
		\vspace{-0.3mm}
	\end{minipage}

    \begin{minipage}[t]{0.2\linewidth}
	\centering
    \centerline{\includegraphics[width=0.99\linewidth]{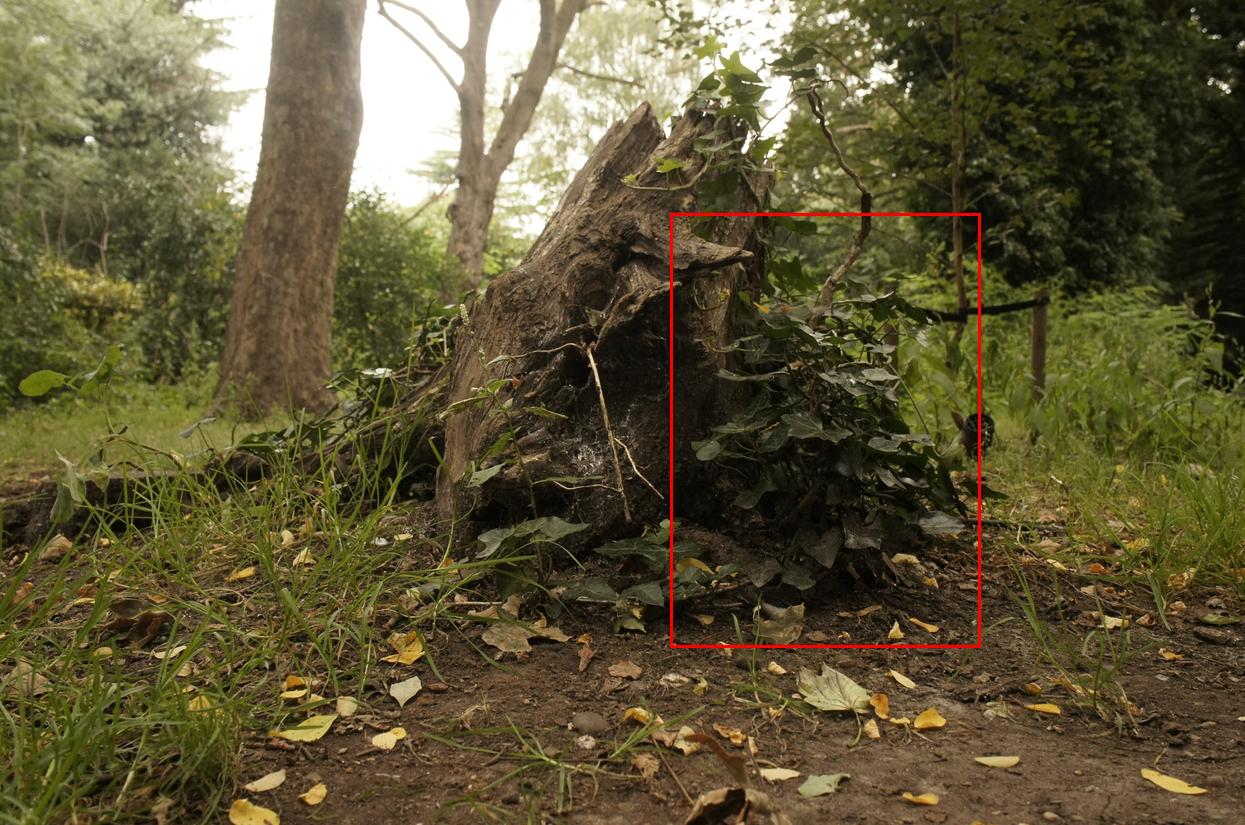}}
		\vspace{-0.3mm}
	\end{minipage}
	\begin{minipage}[t]{0.15\linewidth}
		\centering
    \centerline{\includegraphics[width=0.99\linewidth]{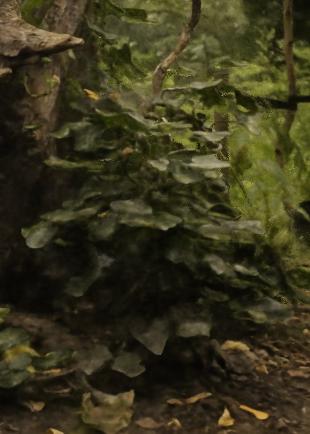}}
		\vspace{-0.3mm}
	\end{minipage}
	\begin{minipage}[t]{0.15\linewidth}
		\centering
    \centerline{\includegraphics[width=0.99\linewidth]{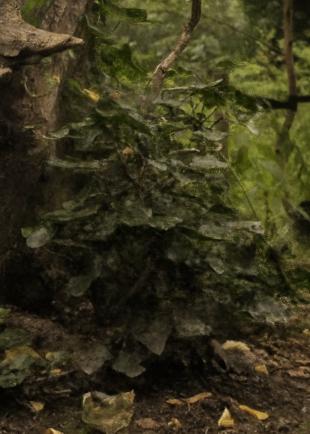}}
		\vspace{-0.3mm}
	\end{minipage}
	\begin{minipage}[t]{0.15\linewidth}
		\centering
    \centerline{\includegraphics[width=0.99\linewidth]{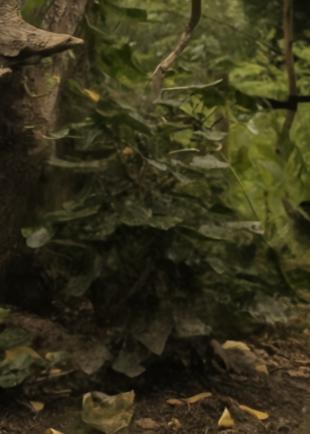}}
		\vspace{-0.3mm}
	\end{minipage}
   \begin{minipage}[t]{0.15\linewidth}
		\centering
    \centerline{\includegraphics[width=0.99\linewidth]{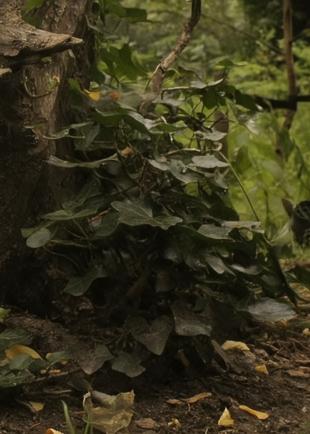}}
		\vspace{-0.3mm}
	\end{minipage}
	\begin{minipage}[t]{0.15\linewidth}
		\centering
	\centerline{\includegraphics[width=0.99\linewidth]{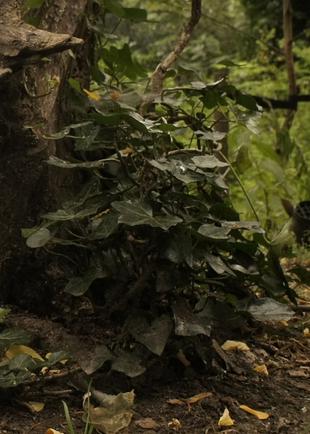}}
		\vspace{-0.3mm}
	\end{minipage}

    \begin{minipage}[t]{0.2\linewidth}
		\centering
	\centerline{\includegraphics[width=0.99\linewidth]{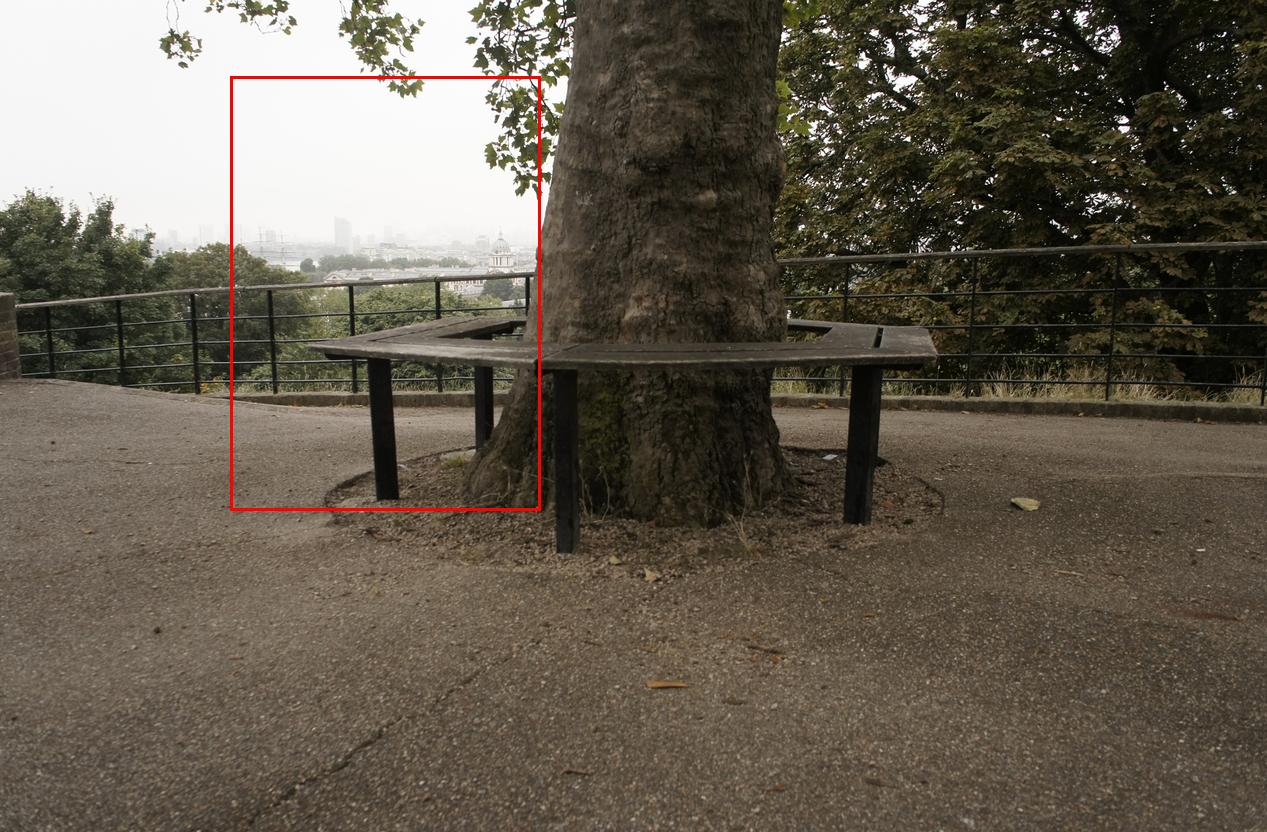}}
		\vspace{-0.3mm}
		\centerline{Full Image}
	\end{minipage}
	\begin{minipage}[t]{0.15\linewidth}
		\centering
	\centerline{\includegraphics[width=0.99\linewidth]{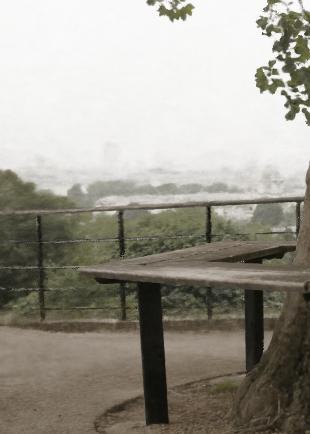}}
		\vspace{-0.3mm}
		\centerline{MipNeRF-360}
	\end{minipage}
	\begin{minipage}[t]{0.15\linewidth}
		\centering
	\centerline{\includegraphics[width=0.99\linewidth]{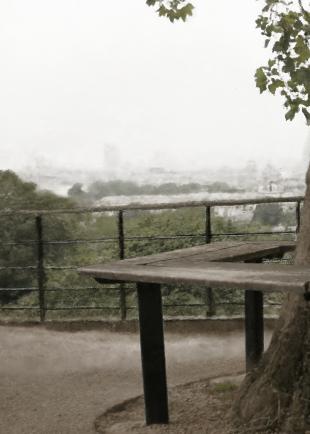}}
		\vspace{-0.3mm}
		\centerline{ZipNeRF}
	\end{minipage}
    \begin{minipage}[t]{0.15\linewidth}
		\centering
	\centerline{\includegraphics[width=0.99\linewidth]{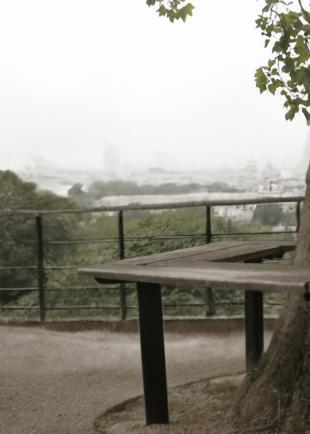}}
		\vspace{-0.3mm}
		\centerline{ZipNeRF}
        \centerline{+NeRFLiX}
	\end{minipage}
	\begin{minipage}[t]{0.15\linewidth}
		\centering
	\centerline{\includegraphics[width=0.99\linewidth]{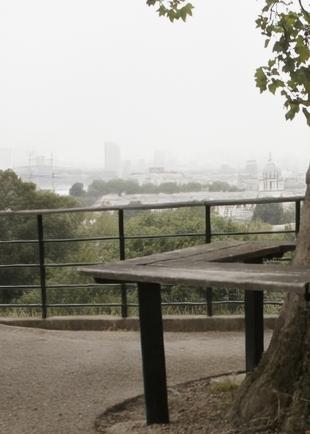}}
		\vspace{-0.3mm}
		\centerline{ZipNeRF}
        \centerline{+\ours}
	\end{minipage}
	\begin{minipage}[t]{0.15\linewidth}
		\centering
	\centerline{\includegraphics[width=0.99\linewidth]{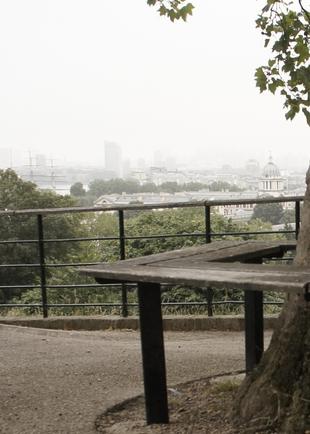}}
		\vspace{-0.3mm}
		\centerline{G.T.}
	\end{minipage}
	
	\caption{More qualitative comparisons for outdoor scenes in MipNeRF-360~\cite{miperf360} dataset. From top to down: \textit{bicycle}, \textit{flowers}, \textit{garden}, \textit{stump} and \textit{treehill}.}
	\label{fig:360outdoors}
	\vspace{-2mm}
\end{figure*}

\begin{figure*}[t]
	\small
	\centering

	\begin{minipage}[t]{0.2\linewidth}
	\centering
		\centerline{\includegraphics[width=0.99\linewidth]{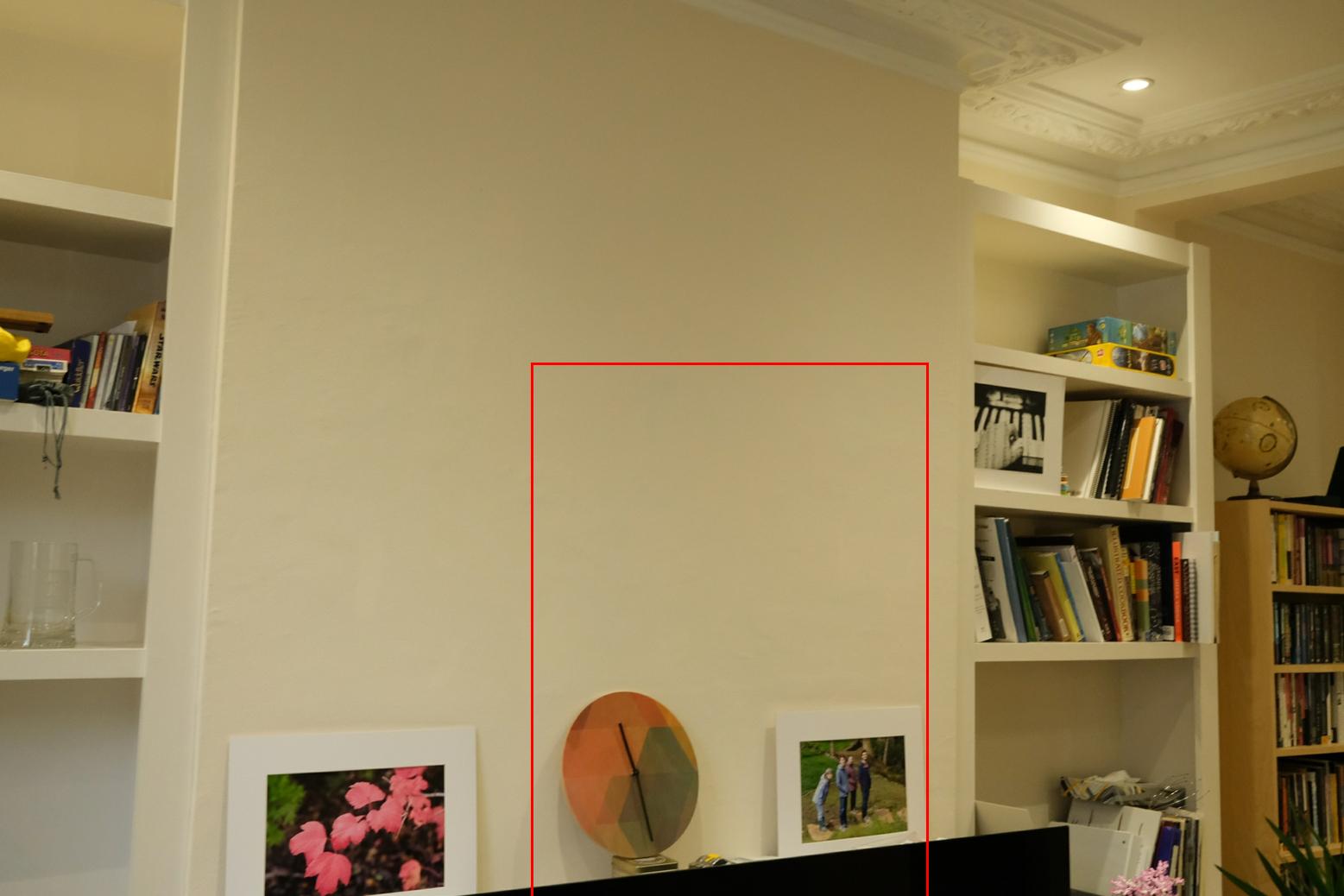}}
		\vspace{-0.3mm}
	\end{minipage}
	\begin{minipage}[t]{0.15\linewidth}
		\centering
		\centerline{\includegraphics[width=0.99\linewidth]{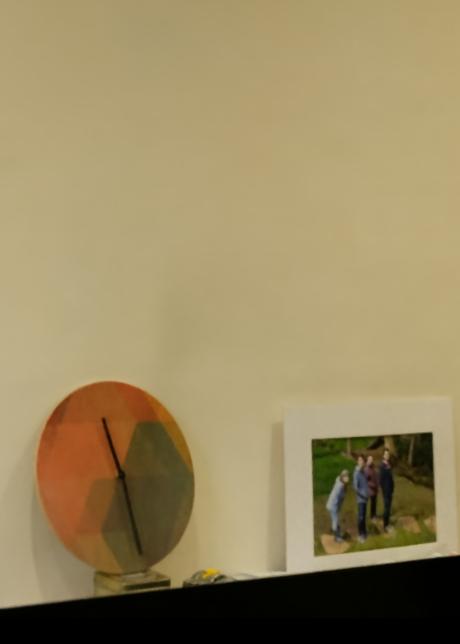}}
		\vspace{-0.3mm}
	\end{minipage}
	\begin{minipage}[t]{0.15\linewidth}
		\centering
		\centerline{\includegraphics[width=0.99\linewidth]{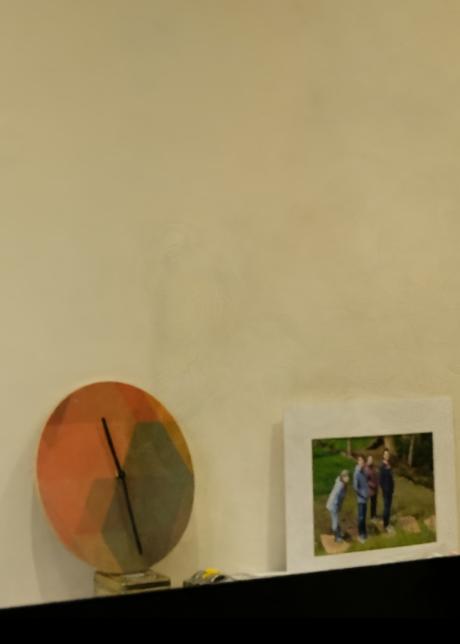}}
		\vspace{-0.3mm}
	\end{minipage}
    \begin{minipage}[t]{0.15\linewidth}
		\centering
		\centerline{\includegraphics[width=0.99\linewidth]{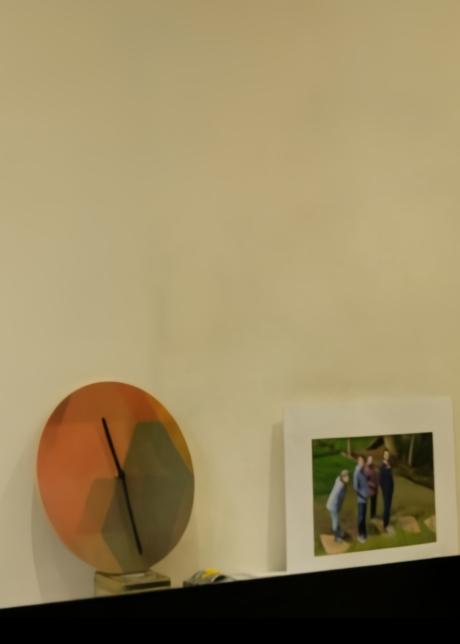}}
		\vspace{-0.3mm}
	\end{minipage}
	\begin{minipage}[t]{0.15\linewidth}
		\centering
		\centerline{\includegraphics[width=0.99\linewidth]{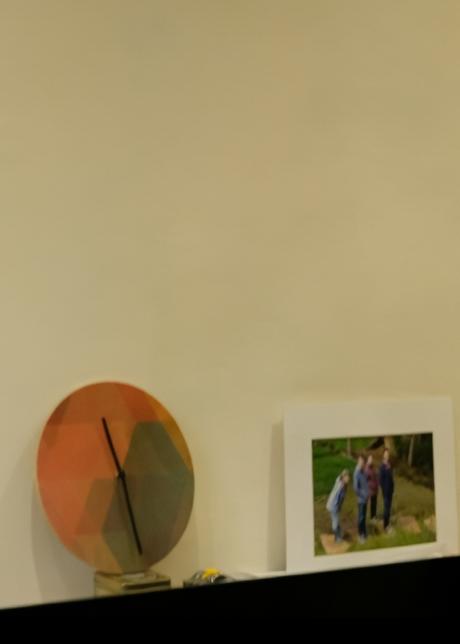}}
		\vspace{-0.3mm}
	\end{minipage}
	\begin{minipage}[t]{0.15\linewidth}
		\centering
		\centerline{\includegraphics[width=0.99\linewidth]{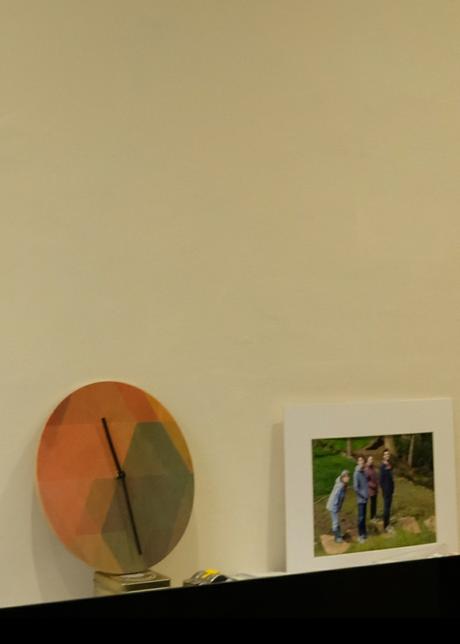}}
		\vspace{-0.3mm}
	\end{minipage}
 
    \begin{minipage}[t]{0.2\linewidth}
	\centering
		\centerline{\includegraphics[width=0.99\linewidth]{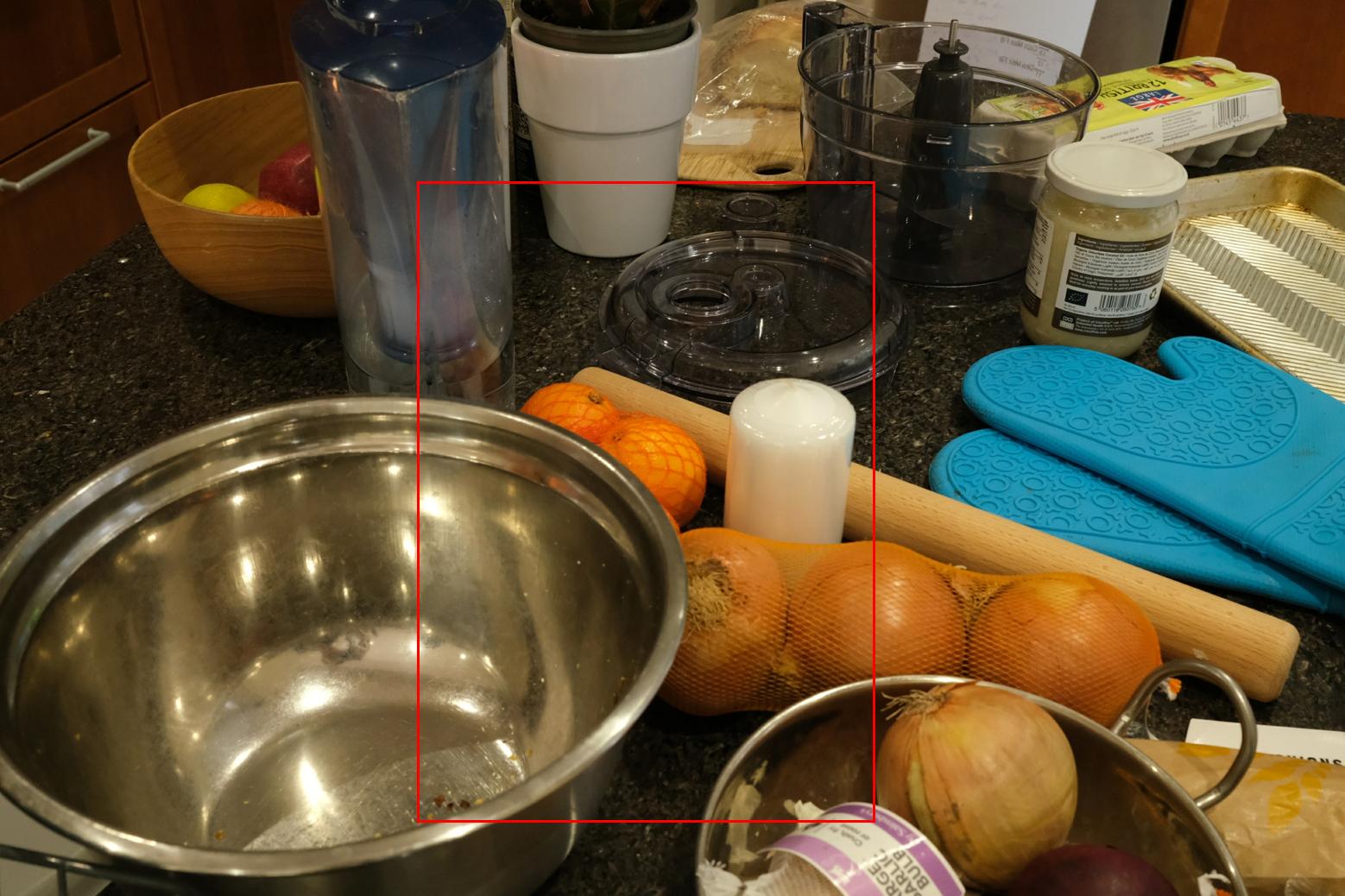}}
		\vspace{-0.3mm}
	\end{minipage}
	\begin{minipage}[t]{0.15\linewidth}
		\centering
		\centerline{\includegraphics[width=0.99\linewidth]{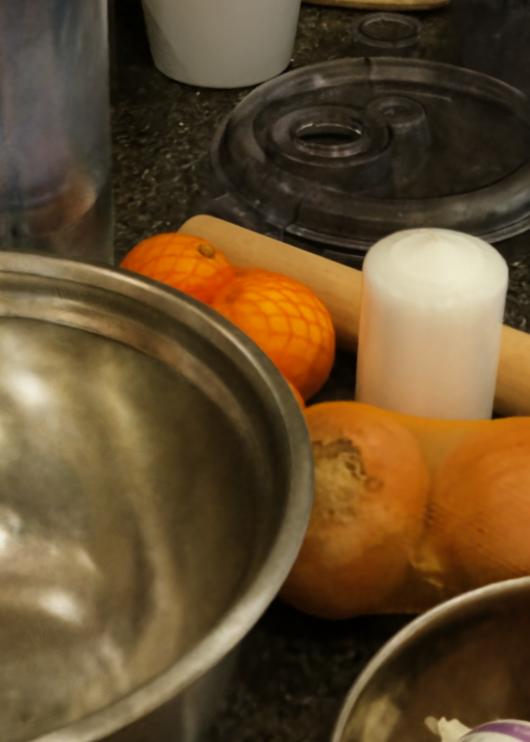}}
		\vspace{-0.3mm}
	\end{minipage}
	\begin{minipage}[t]{0.15\linewidth}
		\centering
		\centerline{\includegraphics[width=0.99\linewidth]{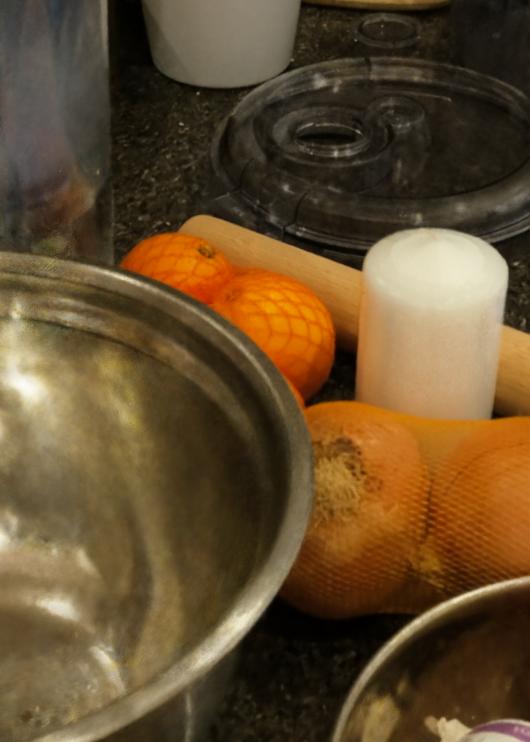}}
		\vspace{-0.3mm}
	\end{minipage}
	\begin{minipage}[t]{0.15\linewidth}
		\centering
		\centerline{\includegraphics[width=0.99\linewidth]{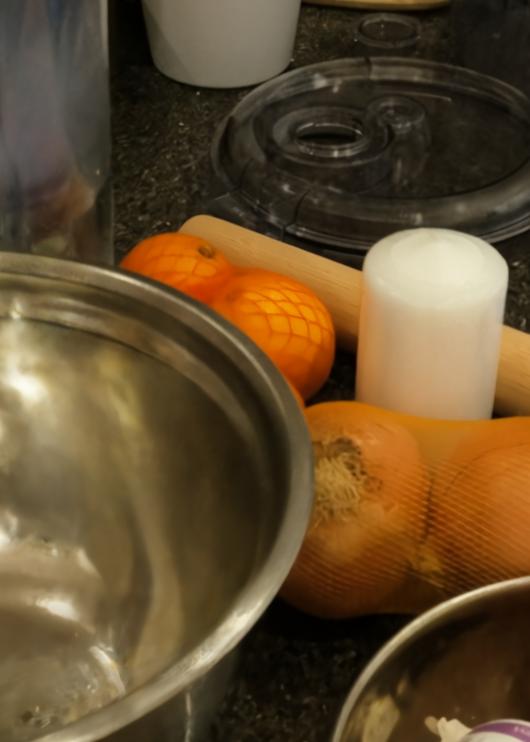}}
		\vspace{-0.3mm}
	\end{minipage}
    \begin{minipage}[t]{0.15\linewidth}
		\centering
		\centerline{\includegraphics[width=0.99\linewidth]{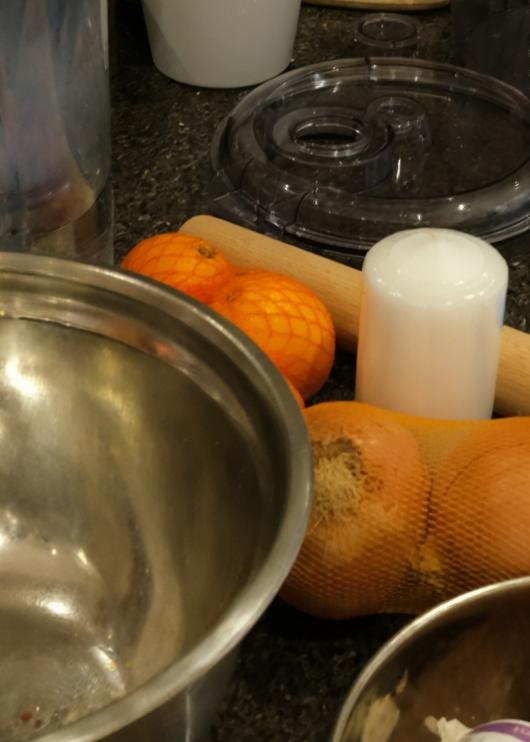}}
		\vspace{-0.3mm}
	\end{minipage}
	\begin{minipage}[t]{0.15\linewidth}
		\centering
		\centerline{\includegraphics[width=0.99\linewidth]{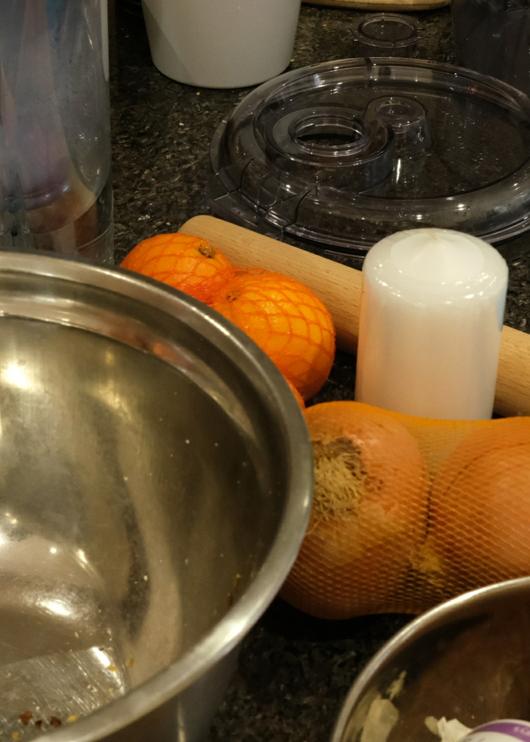}}
		\vspace{-0.3mm}
	\end{minipage}
 
    \begin{minipage}[t]{0.2\linewidth}
	\centering
		\centerline{\includegraphics[width=0.99\linewidth]{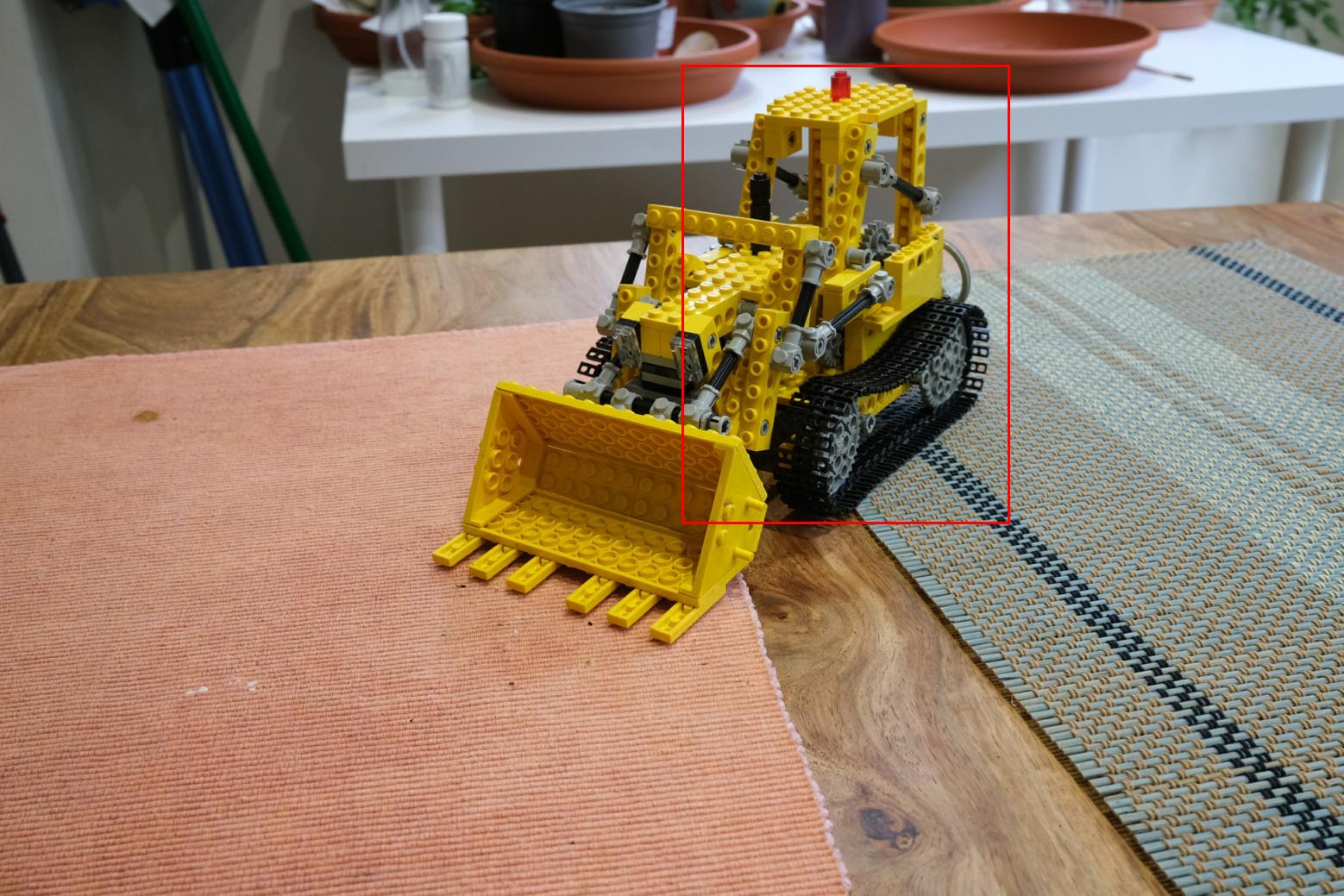}}
		\vspace{-0.3mm}
	\end{minipage}
	\begin{minipage}[t]{0.15\linewidth}
		\centering
		\centerline{\includegraphics[width=0.99\linewidth]{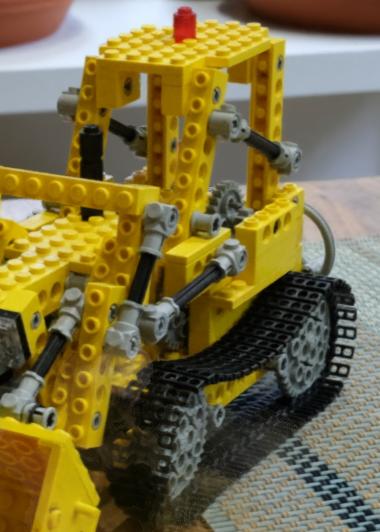}}
		\vspace{-0.3mm}
	\end{minipage}
	\begin{minipage}[t]{0.15\linewidth}
		\centering
		\centerline{\includegraphics[width=0.99\linewidth]{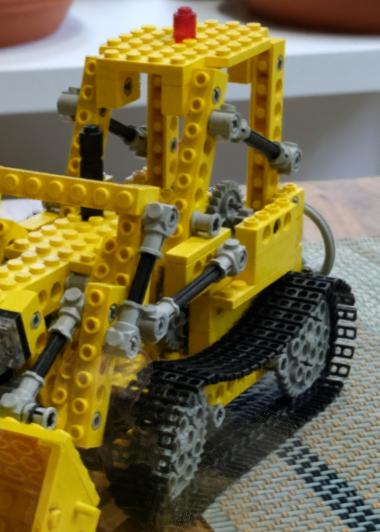}}
		\vspace{-0.3mm}
	\end{minipage}
    \begin{minipage}[t]{0.15\linewidth}
		\centering
		\centerline{\includegraphics[width=0.99\linewidth]{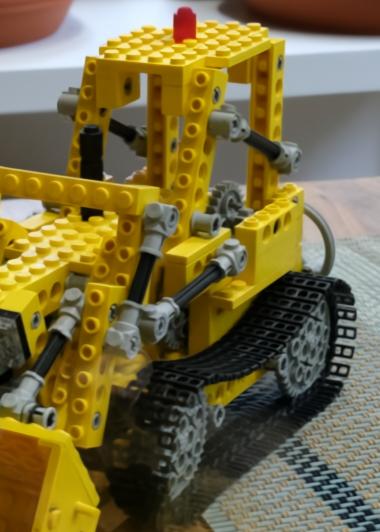}}
		\vspace{-0.3mm}
	\end{minipage}
	\begin{minipage}[t]{0.15\linewidth}
		\centering
		\centerline{\includegraphics[width=0.99\linewidth]{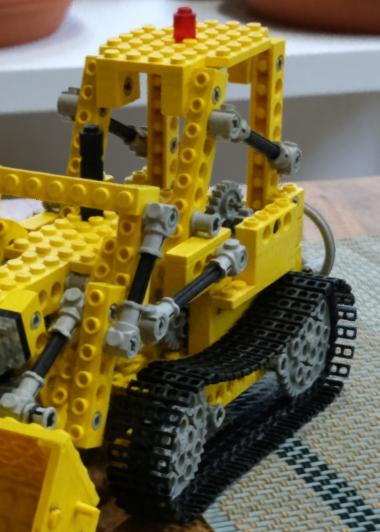}}
		\vspace{-0.3mm}
	\end{minipage}
	\begin{minipage}[t]{0.15\linewidth}
		\centering
		\centerline{\includegraphics[width=0.99\linewidth]{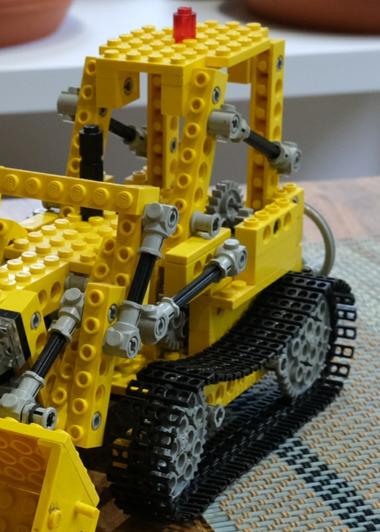}}
		\vspace{-0.3mm}
	\end{minipage}

	\begin{minipage}[t]{0.2\linewidth}
	\centering
		\centerline{\includegraphics[width=0.99\linewidth]{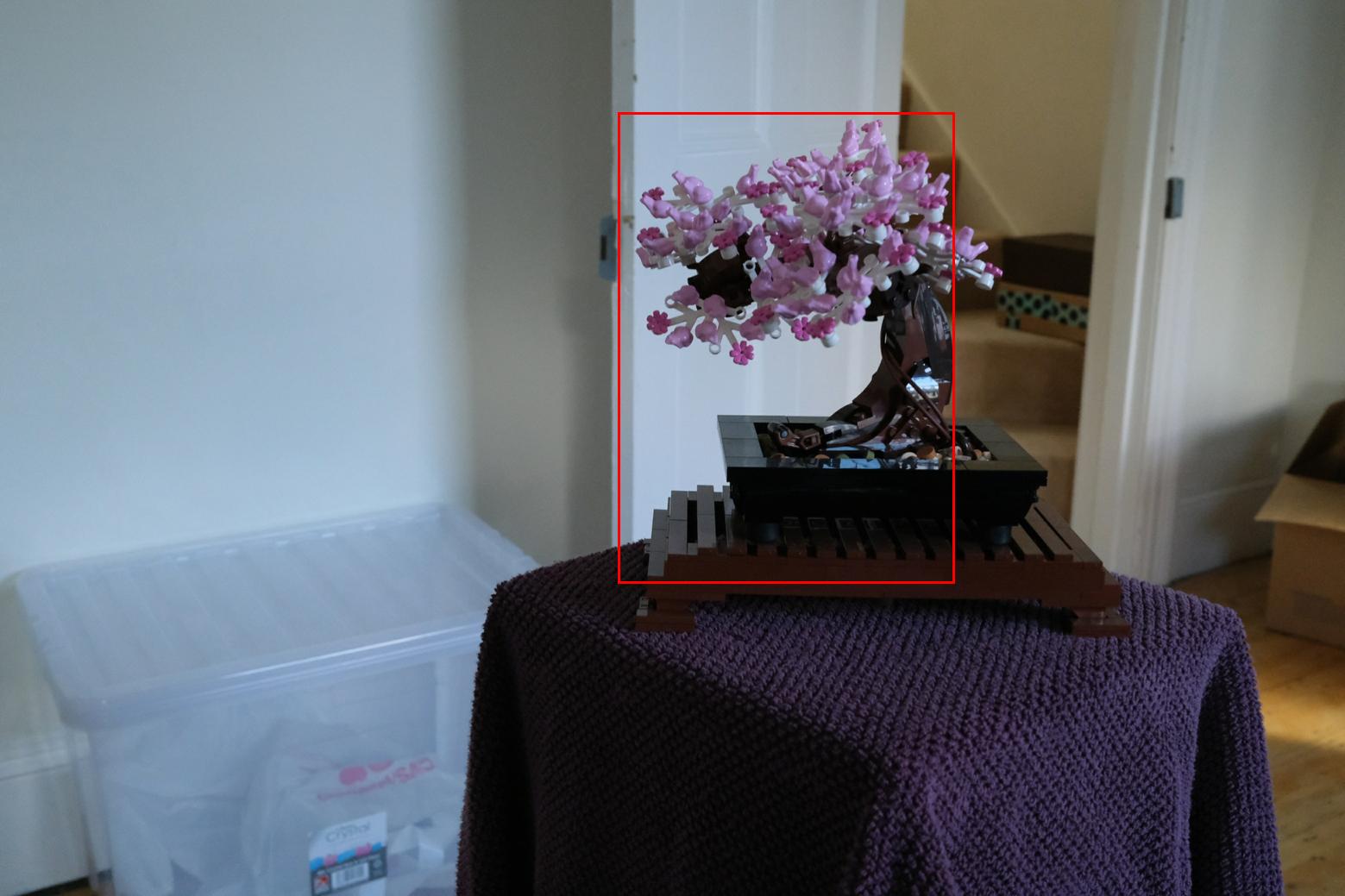}}
		\vspace{-0.3mm}
		\centerline{Full Image}
	\end{minipage}
	\begin{minipage}[t]{0.15\linewidth}
		\centering
		\centerline{\includegraphics[width=0.99\linewidth]{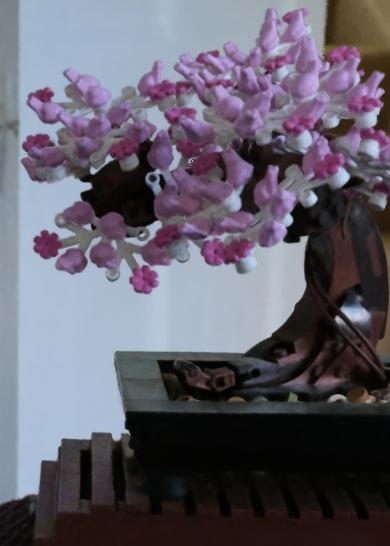}}
		\vspace{-0.3mm}
		\centerline{MipNeRF-360}
	\end{minipage}
	\begin{minipage}[t]{0.15\linewidth}
		\centering
		\centerline{\includegraphics[width=0.99\linewidth]{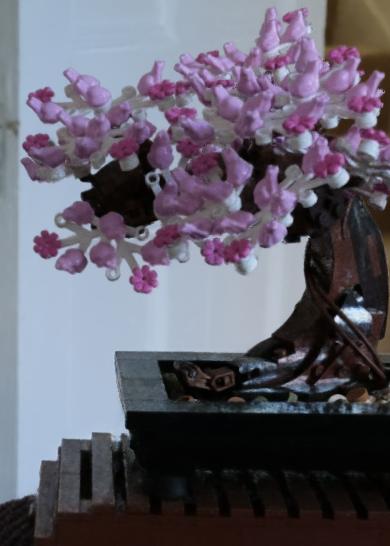}}
		\vspace{-0.3mm}
		\centerline{ZipNeRF}
	\end{minipage}
    \begin{minipage}[t]{0.15\linewidth}
		\centering
		\centerline{\includegraphics[width=0.99\linewidth]{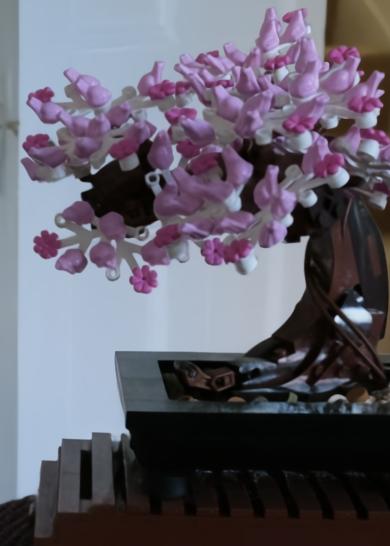}}
		\vspace{-0.3mm}
		\centerline{ZipNeRF}
        \centerline{+NeRFLiX}
    \end{minipage}
	\begin{minipage}[t]{0.15\linewidth}
		\centering
		\centerline{\includegraphics[width=0.99\linewidth]{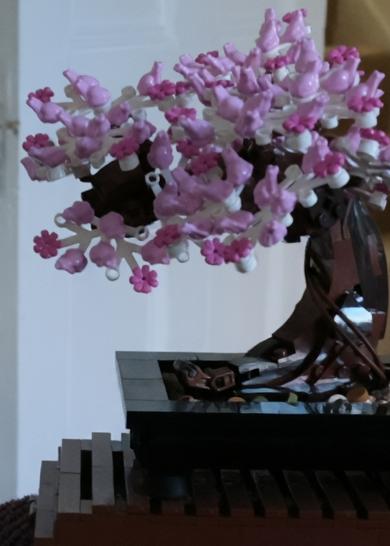}}
		\vspace{-0.3mm}
		\centerline{ZipNeRF}
        \centerline{+\ours}
	\end{minipage}
	\begin{minipage}[t]{0.15\linewidth}
		\centering
		\centerline{\includegraphics[width=0.99\linewidth]{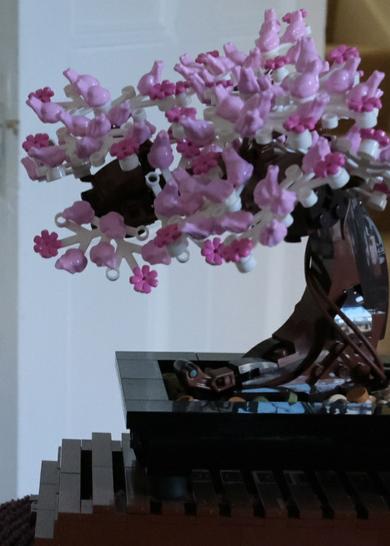}}
		\vspace{-0.3mm}
		\centerline{G.T.}
	\end{minipage}

	\caption{More qualitative comparisons for indoor scenes in MipNeRF-360~\cite{miperf360} dataset. From top to down: \textit{room}, \textit{counter}, \textit{kitchen} and \textit{bonsai}.}
	\label{fig:360indoors}
	\vspace{-2mm}
\end{figure*}

More qualitative comparisons for each scene are located in Figure \ref{fig:360outdoors} (five outdoor scenes) and Figure \ref{fig:360indoors} (four indoor scenes). From the visualizations, we prove that \ours~is capable of restoring from various types of aliasing degradations: 
1) Reducing various kinds of rendering artifacts, like removing the dirts on the lego track in the \textit{kitchen} (row-3 in Figure \ref{fig:360indoors}), wiping off the noise specks on the leaves in the \textit{stump} (row-4 in Figure \ref{fig:360outdoors}) and \textit{flowers} (row-2 in Figure \ref{fig:360outdoors}). 
2) Generating more realistic appearance details and sharpening the edges, like the surfaces of \textit{flowers} (row-2 in Figure \ref{fig:360outdoors}) and the leaves on the \textit{stump} (row-4 in Figure \ref{fig:360outdoors}). 
3) Restoring more reasonable geometry structures, like the spokes of the \textit{bicycle} (row-1 in Figure \ref{fig:360outdoors}), the table leg in the \textit{garden} (row-3 in Figure \ref{fig:360outdoors}), and the fences 
in the \textit{treehill} (row-5 in Figure \ref{fig:360outdoors}).
4) Rectifying the illuminations and reflections, like the wall in the \textit{room} (row-1 in Figure \ref{fig:360indoors}), the plastic and metal surfaces in the \textit{counter} (row-2 in Figure \ref{fig:360indoors}), and the base of the \textit{bonsai} (row-4 in Figure \ref{fig:360indoors}). For a more intuitive visual comparison, we also include a video demo in the supplementary material.

\subsection{More Results for the Ablation Studies}
\label{appendix:ablation}


\begin{wrapfigure}{r}{0.5\textwidth}  
  \centering  
  \vspace{-2mm}
  \includegraphics[width=0.5\textwidth]{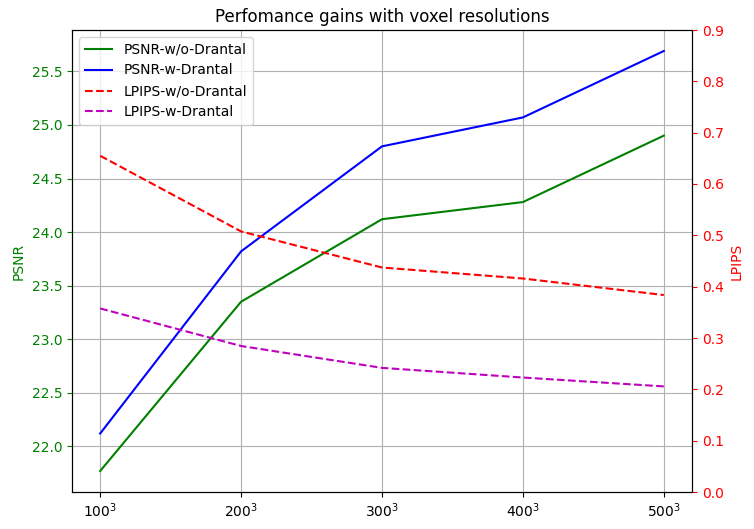}  
  \vspace{-5mm}
  \caption{Line chart plotting the trend for performance gains with the change of voxel resolutions. Solid line: PSNR. Dash line: LPIPS.}
  \label{fig:ablation} 
\end{wrapfigure}  

Figure \ref{fig:ablation} depicts the trend of performance improvements in relation to the changes in voxel resolutions. For the \textit{Block\_ALL} scene in MatrixCity, we incrementally increase the voxel resolutions for GridNeRF+\ours~from $100^3$ to $500^3$ along x-axis, and plot the values of PSNR and LPIPS along y-axis. We conclude that as the voxel resolution decreases, the performance gain of LPIPS increases while the gain of PSNR shrinks. We reckon that when the voxel resolutions are relatively small, \ours~is still able to generate realistic scene contents, thereby leading to high perceptual quality improvement. However, due to the lack of structural details in the low-quality renderings produced by low-capacity NeRF itself, the conditional generation process may suffer from lower fidelity, therefore the reconstruction metric like PSNR shrinks.

\begin{figure*}[t]
	\small
	\centering

	\begin{minipage}[t]{0.28\linewidth}
		\centering
        \centerline{\includegraphics[width=0.99\linewidth]{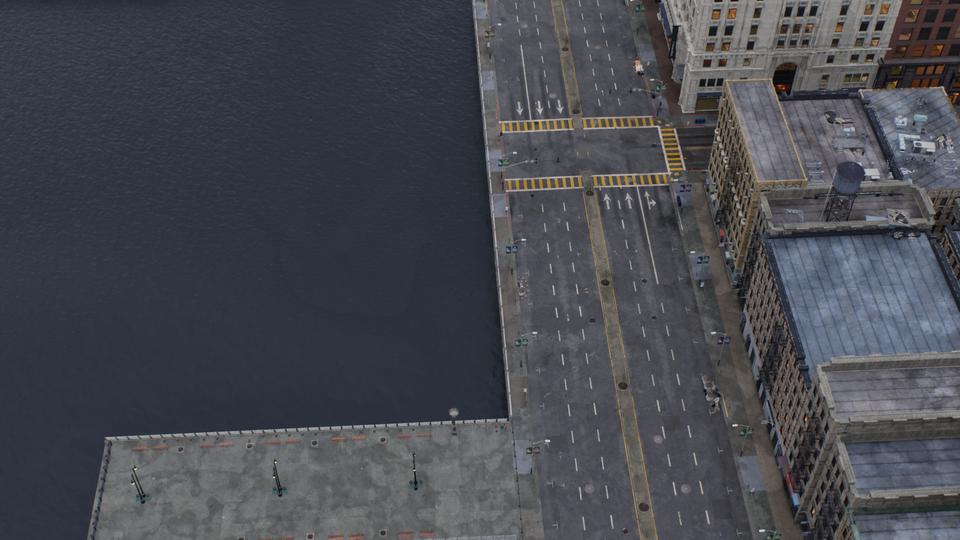}}
		\vspace{-0.3mm}
		\centerline{Ground Truth}
	\end{minipage}
	\begin{minipage}[t]{0.28\linewidth}
		\centering
		\centerline{\includegraphics[width=0.99\linewidth]{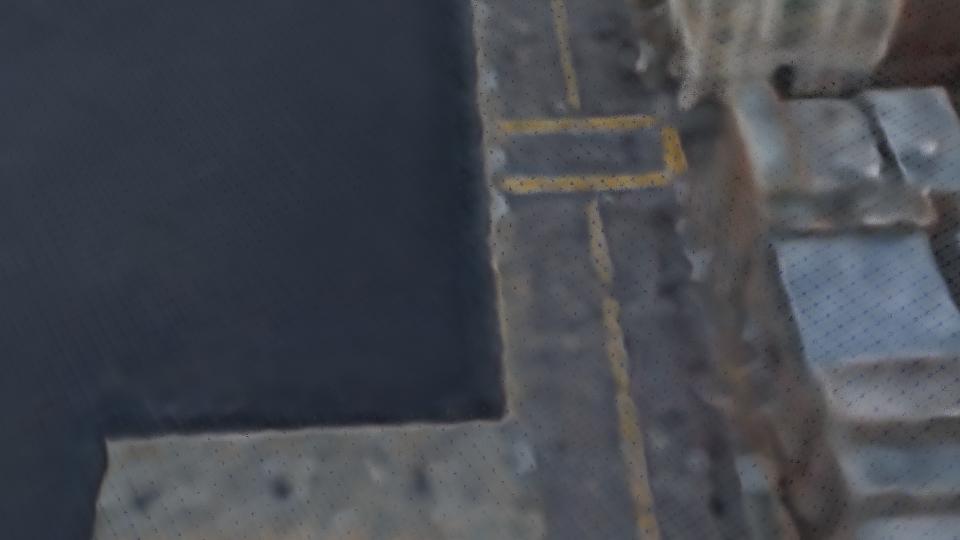}}
		\vspace{-0.3mm}
		\centerline{GridNeRF ($100^3$)}
	\end{minipage}
	\begin{minipage}[t]{0.28\linewidth}
		\centering
		\centerline{\includegraphics[width=0.99\linewidth]{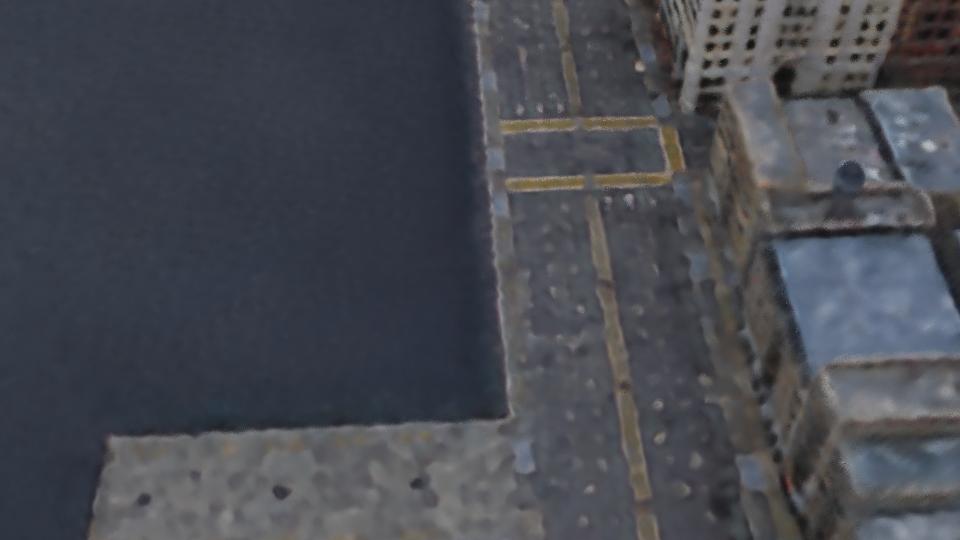}}
		\vspace{-0.3mm}
		\centerline{GridNeRF ($200^3$)}
	\end{minipage}

        \begin{minipage}[t]{0.28\linewidth}
		\centering
        \hspace{+1mm}
	\end{minipage}
        \begin{minipage}[t]{0.28\linewidth}
		\centering
		\centerline{\includegraphics[width=0.99\linewidth]{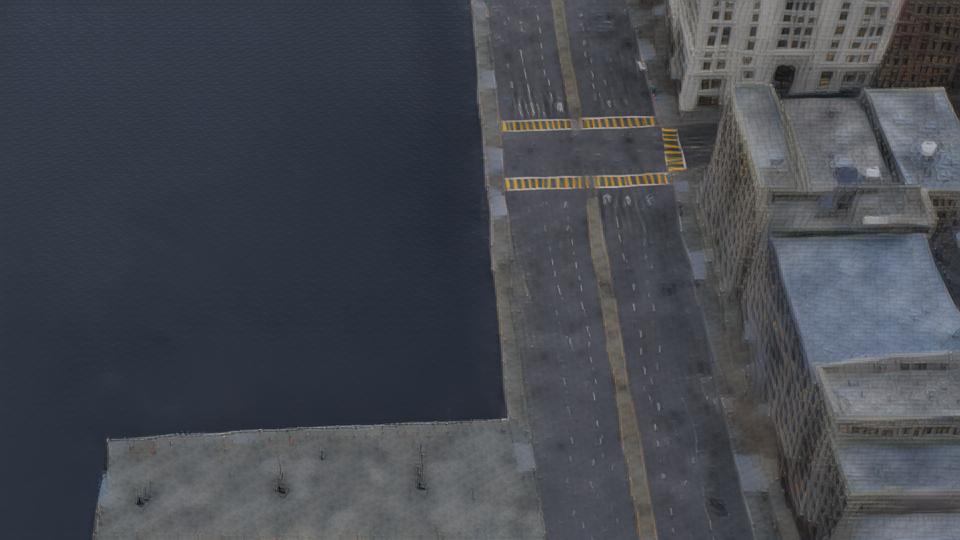}}
		\vspace{-0.3mm}
		\centerline{GridNeRF}
        \centerline{+\ours~($100^3$)}
	\end{minipage}
        \begin{minipage}[t]{0.28\linewidth}
		\centering
		\centerline{\includegraphics[width=0.99\linewidth]{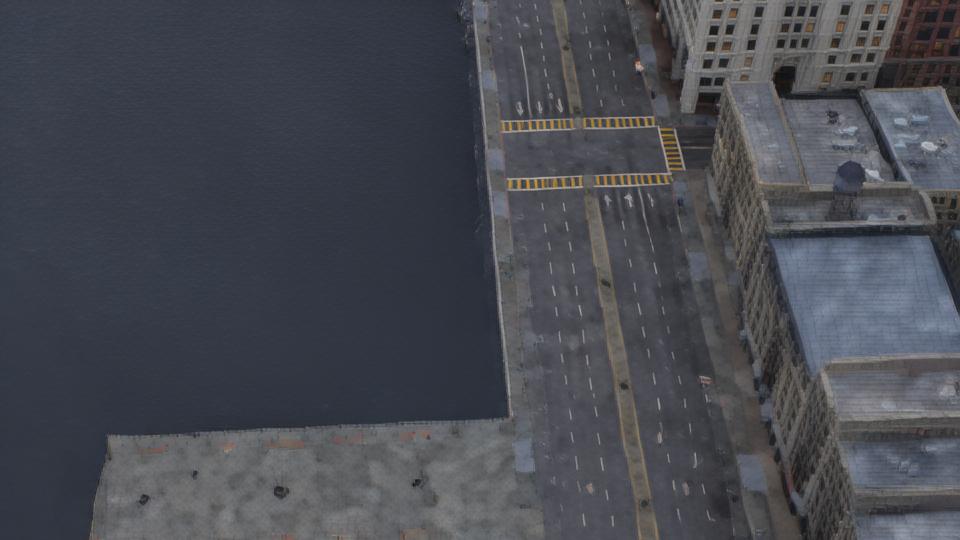}}
		\vspace{-0.3mm}
		\centerline{GridNeRF}
        \centerline{+\ours~($200^3$)}
  	\end{minipage} 
        
        \vspace{+5mm}

	\begin{minipage}[t]{0.28\linewidth}
		\centering
		\centerline{\includegraphics[width=0.99\linewidth]{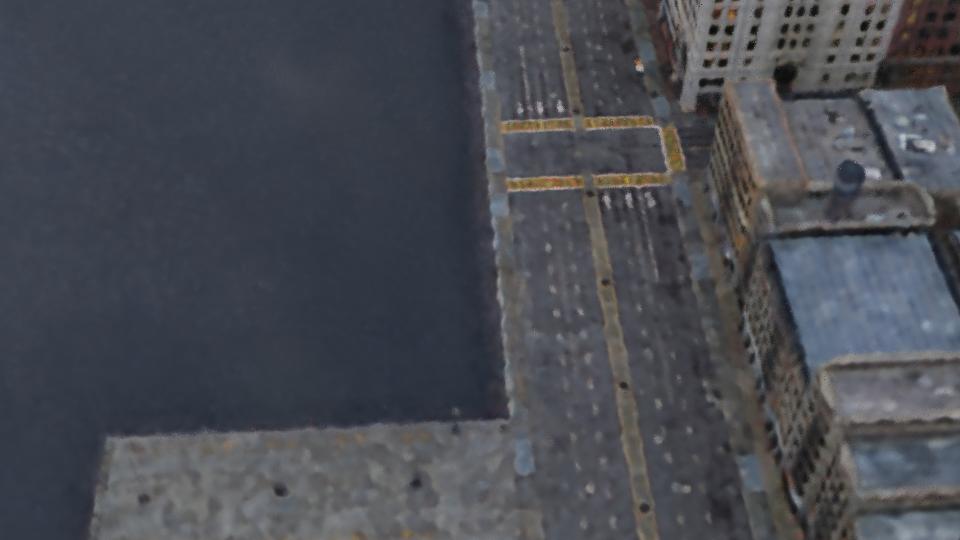}}
		\vspace{-0.3mm}
		\centerline{GridNeRF ($300^3$)}
	\end{minipage}
	\begin{minipage}[t]{0.28\linewidth}
		\centering
		\centerline{\includegraphics[width=0.99\linewidth]{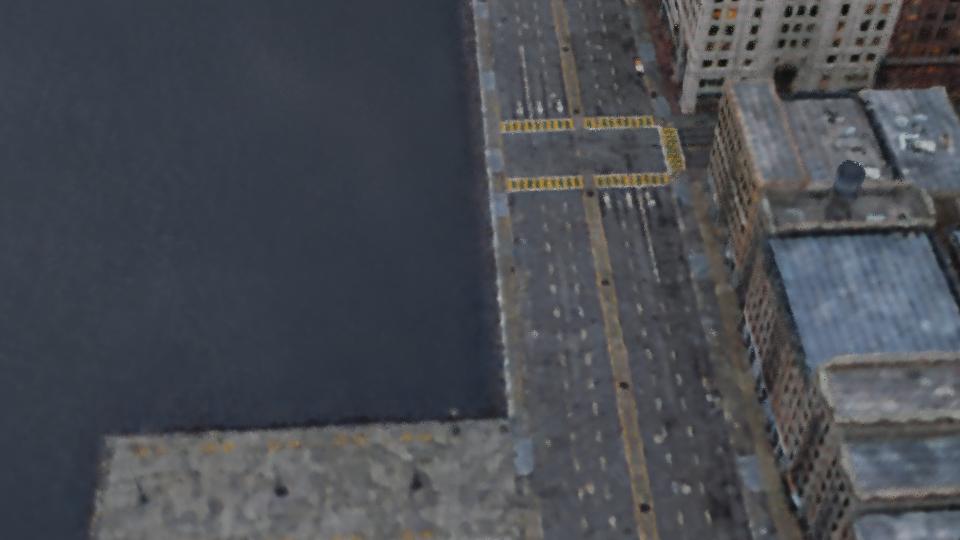}}
		\vspace{-0.3mm}
		\centerline{GridNeRF ($400^3$)}
	\end{minipage}
        \begin{minipage}[t]{0.28\linewidth}
		\centering
		\centerline{\includegraphics[width=0.99\linewidth]{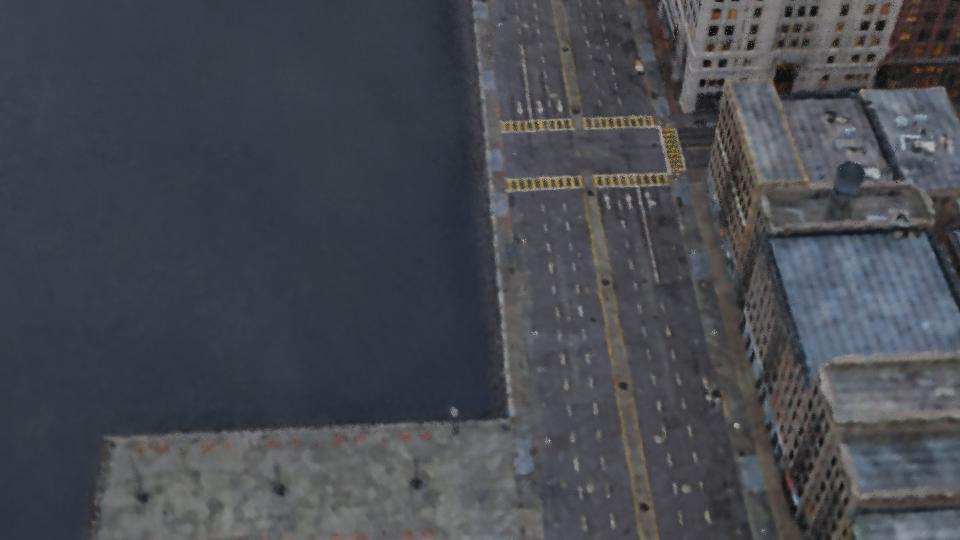}}
		\vspace{-0.3mm}
		\centerline{GridNeRF ($500^3$)}
	\end{minipage}

        \begin{minipage}[t]{0.28\linewidth}
		\centering
		\centerline{\includegraphics[width=0.99\linewidth]{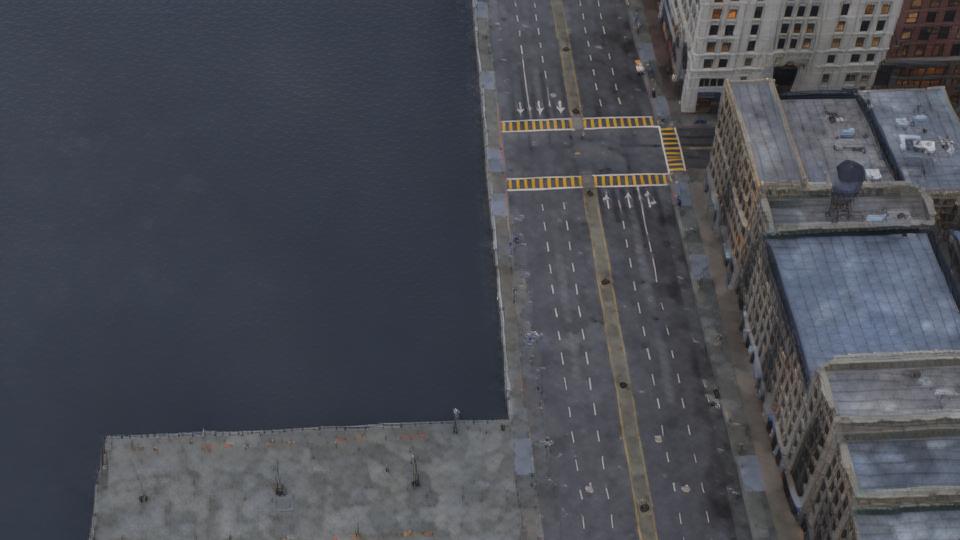}}
		\vspace{-0.3mm}
		\centerline{GridNeRF}
        \centerline{+\ours~($300^3$)}
	\end{minipage}
	\begin{minipage}[t]{0.28\linewidth}
		\centering
		\centerline{\includegraphics[width=0.99\linewidth]{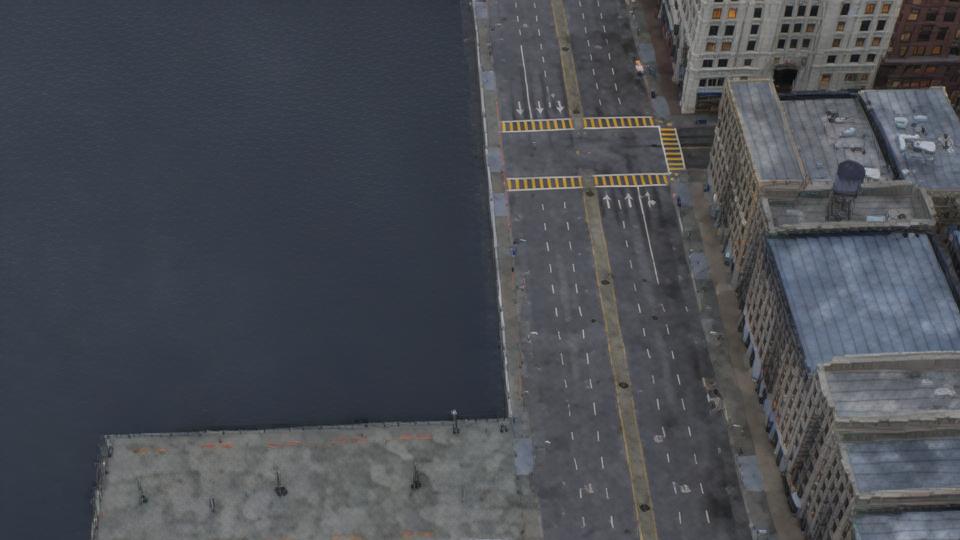}}
		\vspace{-0.3mm}
		\centerline{GridNeRF}
        \centerline{+\ours~($400^3$)}
	\end{minipage}
        \begin{minipage}[t]{0.28\linewidth}
		\centering
		\centerline{\includegraphics[width=0.99\linewidth]{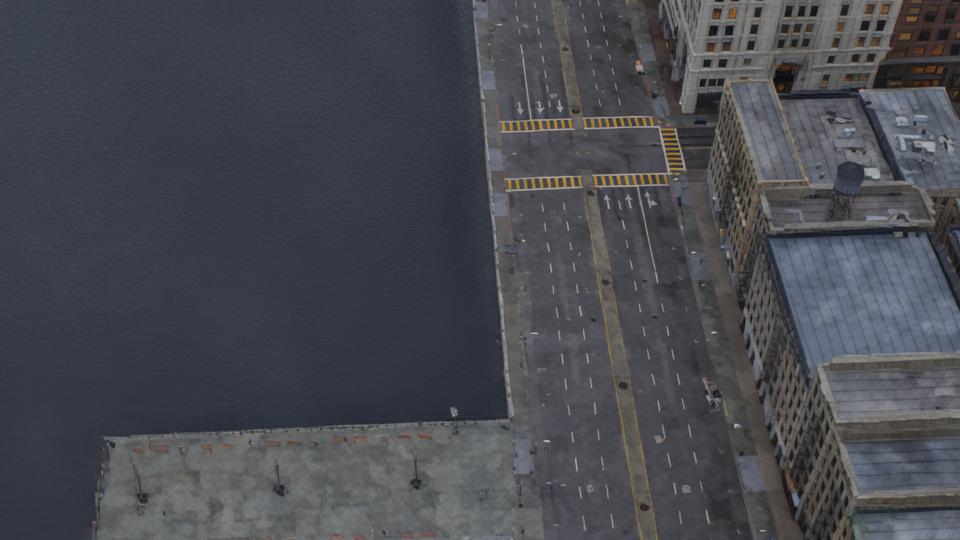}}
		\vspace{-0.3mm}
		\centerline{GridNeRF}
        \centerline{+\ours~($500^3$)}
	\end{minipage}
    \caption{The visual comparison of GridNeRF and GridNeRF+\ours~for \textit{Block\_ALL} scene as the voxel resolution changes from $100^3$ to $500^3$ (Part 1).}
    \label{fig:matrixcityablapart1}
\end{figure*}
\begin{figure*}[t]
	\small
	\centering

	\begin{minipage}[t]{0.28\linewidth}
		\centering
        \centerline{\includegraphics[width=0.99\linewidth]{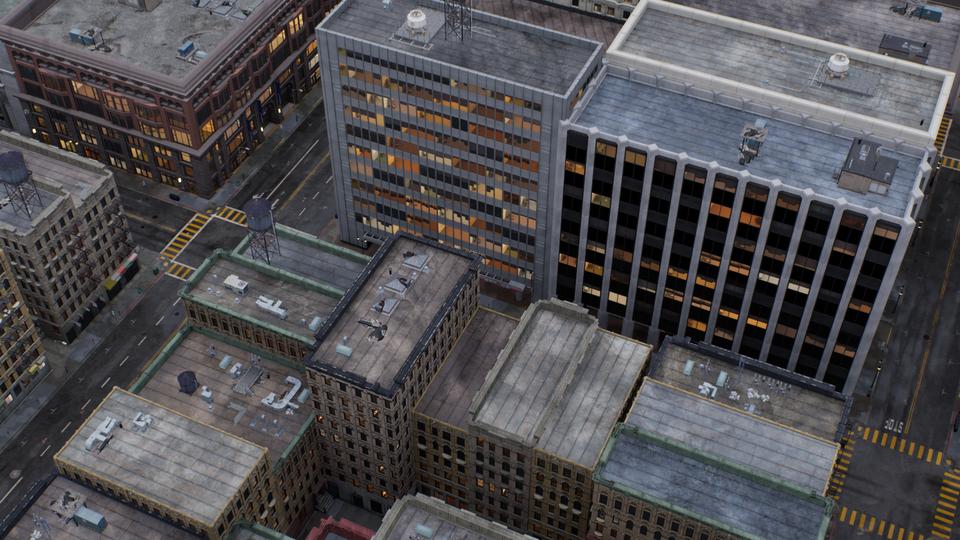}}
		\vspace{-0.3mm}
		\centerline{Ground Truth}
	\end{minipage}
	\begin{minipage}[t]{0.28\linewidth}
		\centering
		\centerline{\includegraphics[width=0.99\linewidth]{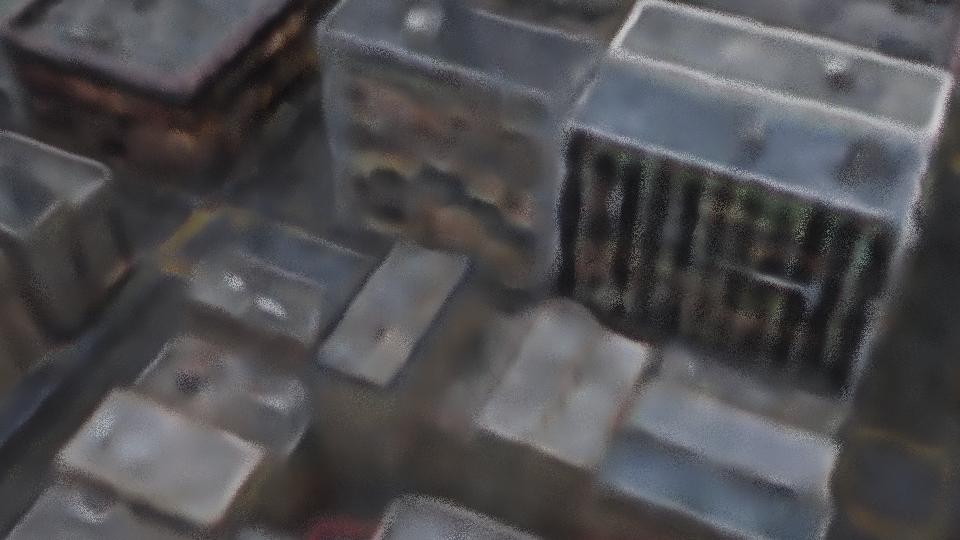}}
		\vspace{-0.3mm}
		\centerline{GridNeRF ($100^3$)}
	\end{minipage}
	\begin{minipage}[t]{0.28\linewidth}
		\centering
		\centerline{\includegraphics[width=0.99\linewidth]{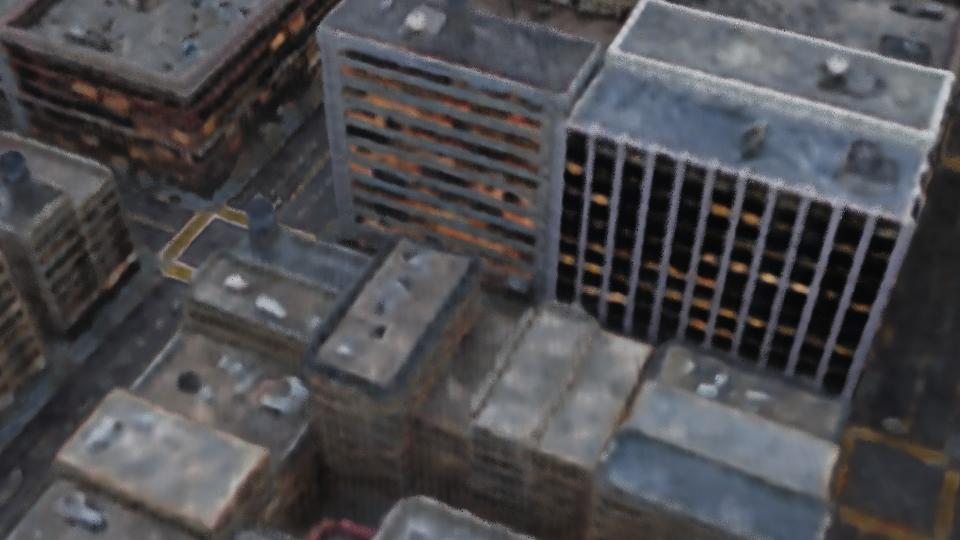}}
		\vspace{-0.3mm}
		\centerline{GridNeRF ($200^3$)}
	\end{minipage}

        \begin{minipage}[t]{0.28\linewidth}
		\centering
        \hspace{+1mm}
	\end{minipage}
        \begin{minipage}[t]{0.28\linewidth}
		\centering
		\centerline{\includegraphics[width=0.99\linewidth]{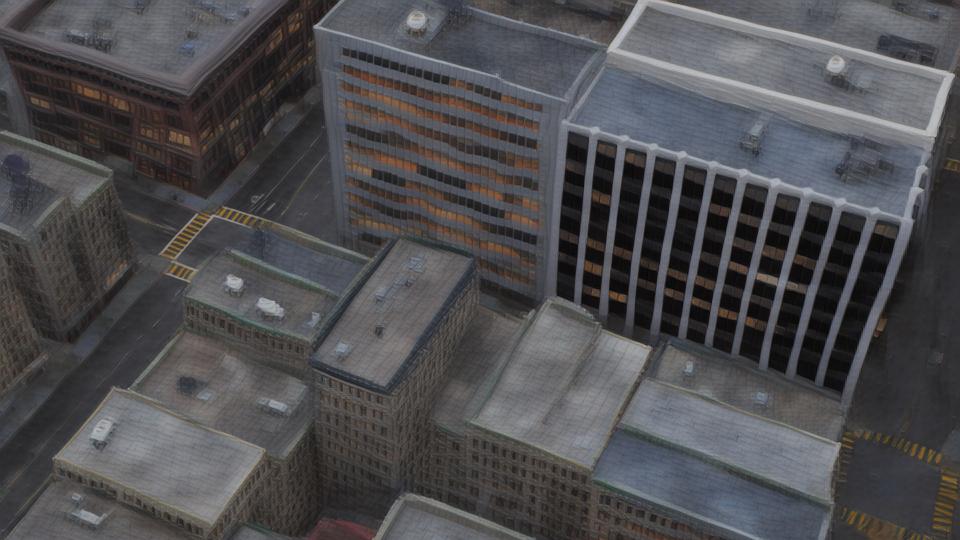}}
		\vspace{-0.3mm}
		\centerline{GridNeRF}
        \centerline{+\ours~($100^3$)}
	\end{minipage}
        \begin{minipage}[t]{0.28\linewidth}
		\centering
		\centerline{\includegraphics[width=0.99\linewidth]{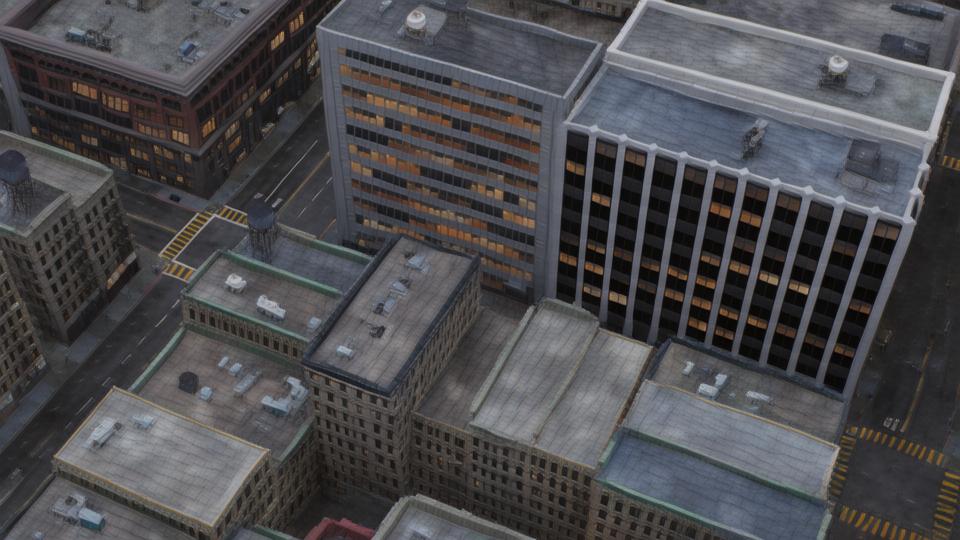}}
		\vspace{-0.3mm}
		\centerline{GridNeRF}
        \centerline{+\ours~($200^3$)}
  	\end{minipage} 
        
        \vspace{+5mm}

	\begin{minipage}[t]{0.28\linewidth}
		\centering
		\centerline{\includegraphics[width=0.99\linewidth]{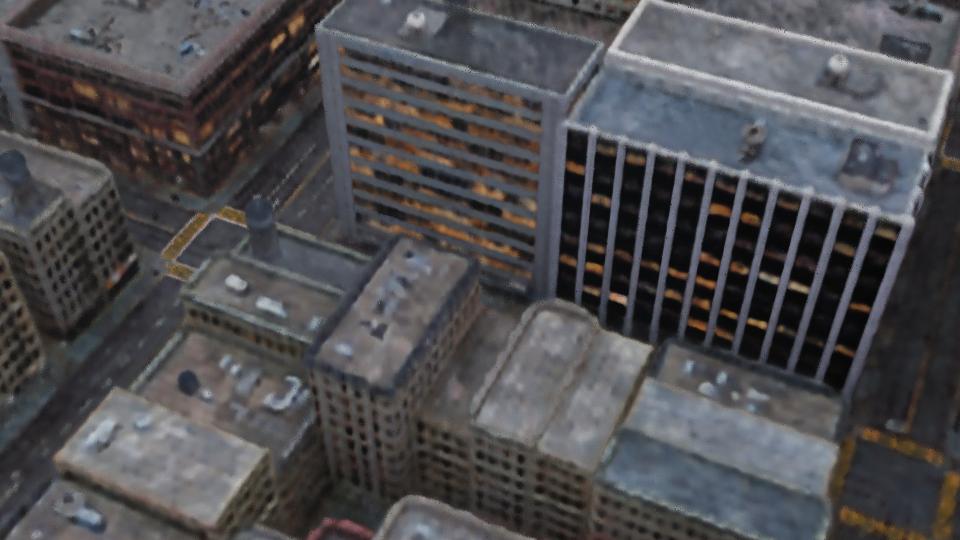}}
		\vspace{-0.3mm}
		\centerline{GridNeRF ($300^3$)}
	\end{minipage}
	\begin{minipage}[t]{0.28\linewidth}
		\centering
		\centerline{\includegraphics[width=0.99\linewidth]{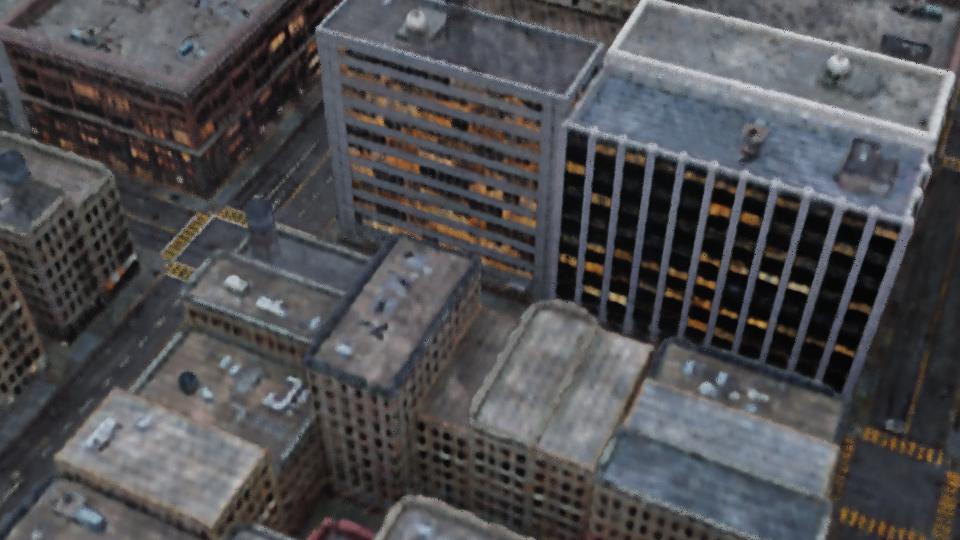}}
		\vspace{-0.3mm}
		\centerline{GridNeRF ($400^3$)}
	\end{minipage}
        \begin{minipage}[t]{0.28\linewidth}
		\centering
		\centerline{\includegraphics[width=0.99\linewidth]{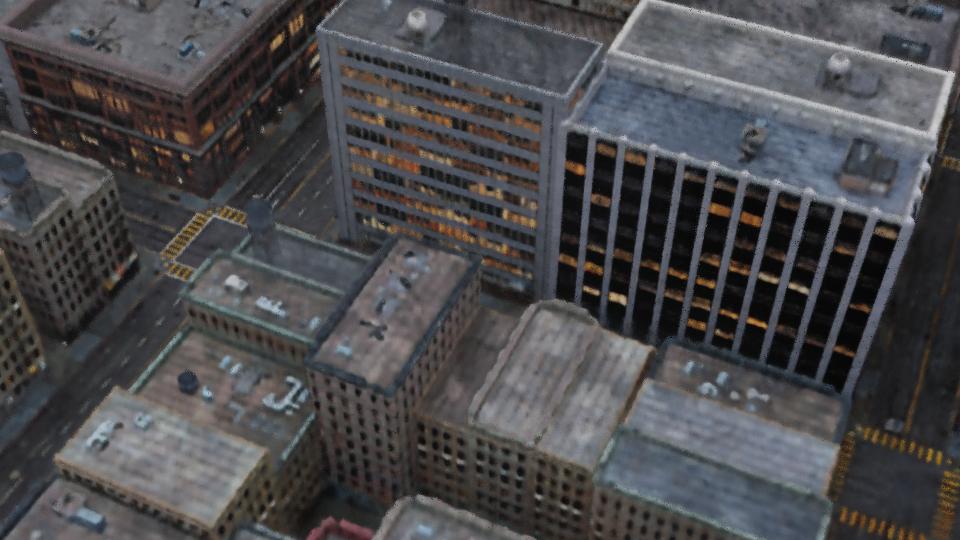}}
		\vspace{-0.3mm}
		\centerline{GridNeRF ($500^3$)}
	\end{minipage}

        \begin{minipage}[t]{0.28\linewidth}
		\centering
		\centerline{\includegraphics[width=0.99\linewidth]{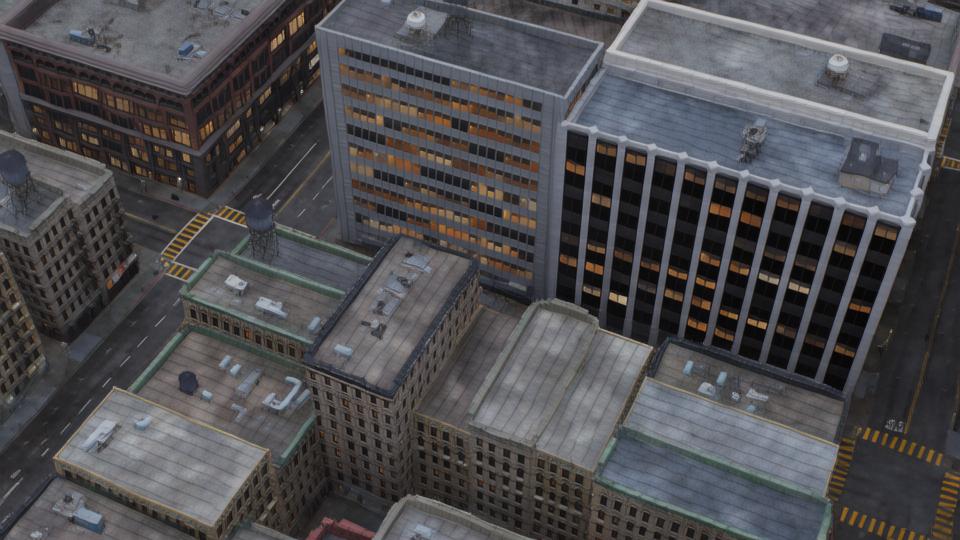}}
		\vspace{-0.3mm}
		\centerline{GridNeRF}
        \centerline{+\ours~($300^3$)}
	\end{minipage}
	\begin{minipage}[t]{0.28\linewidth}
		\centering
		\centerline{\includegraphics[width=0.99\linewidth]{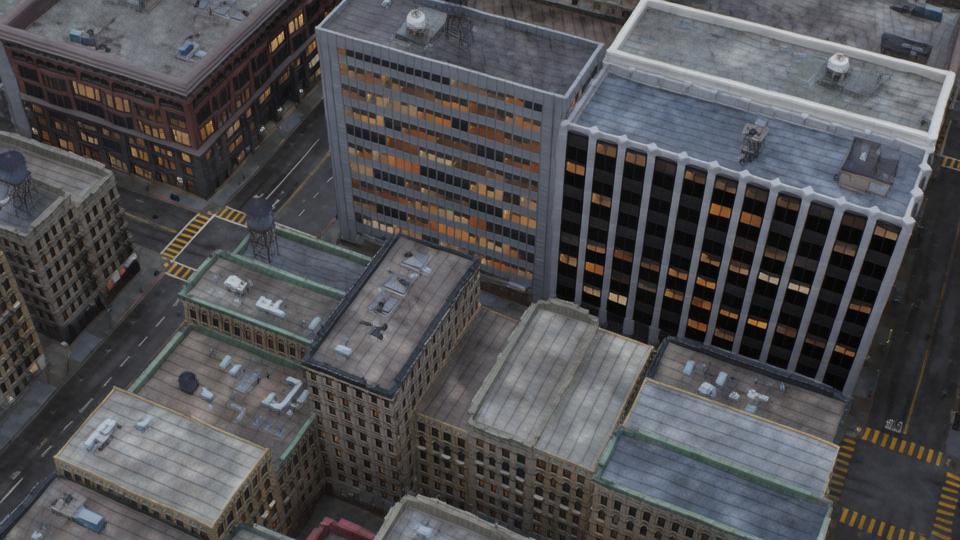}}
		\vspace{-0.3mm}
		\centerline{GridNeRF}
        \centerline{+\ours~($400^3$)}
	\end{minipage}
        \begin{minipage}[t]{0.28\linewidth}
		\centering
		\centerline{\includegraphics[width=0.99\linewidth]{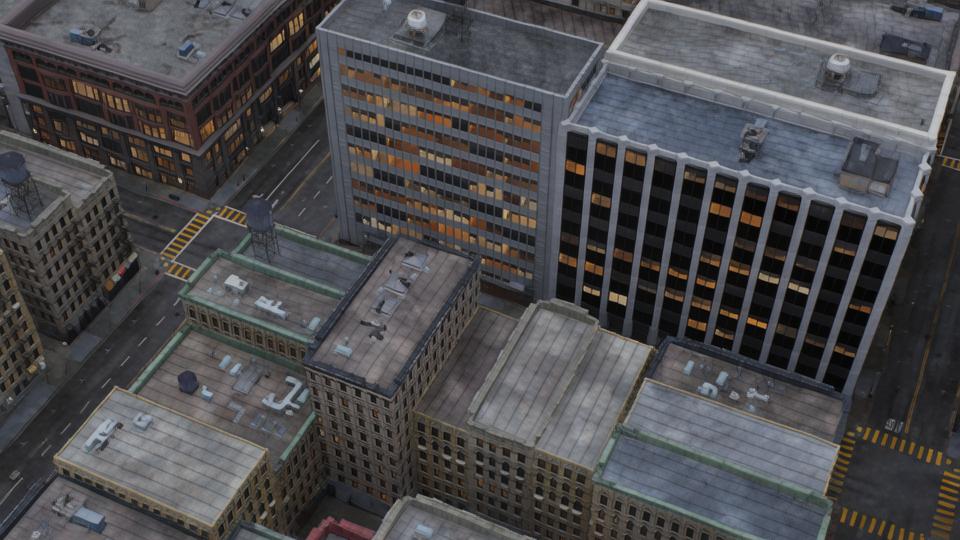}}
		\vspace{-0.3mm}
		\centerline{GridNeRF}
        \centerline{+\ours~($500^3$)}
	\end{minipage}
    \caption{The visual comparison of GridNeRF and GridNeRF+\ours~for \textit{Block\_ALL} scene as the voxel resolution changes from $100^3$ to $500^3$ (Part 2).}
    \label{fig:matrixcityablapart2}
\end{figure*}

In Figure \ref{fig:matrixcityablapart1} and Figure \ref{fig:matrixcityablapart2}, we provide the visual comparisons of GridNeRF and GridNeRF+\ours, as we increase the voxel resolutions incrementally from $100^3$ to $500^3$. From the qualitative results, we conclude that: 1) Under all the voxel resolutions, \ours~exhibits strong capability in generating much more visually appealing renderings with much more finer details and much less blurs and artifacts, which demonstrates the robustness and generalizability of our proposed method. 2) Even in the case of a very low voxel resolution---$100^3$, where the renderings produced by GridNeRF suffer from severe blurs and miss a lot of texture details, \ours~is still capable of restoring most of the appearances with high realism. And with the introduction of \ours, a relatively small voxel resolution like $200^3$ is already able to yield visually-acceptable results in the large-scale scenes. It indicates that the proposed \ours~has great potential in mitigating the memory usage issue which is proportional cubically to the voxel resolutions.

\section{More Implementation Details}
\label{appendix: more imple details}

\subsection{Structure for the Controllable Feature Wrapping Module}
\label{cfw-structure}

\begin{figure}[t]
  \centering
  \includegraphics[width=0.96\textwidth]{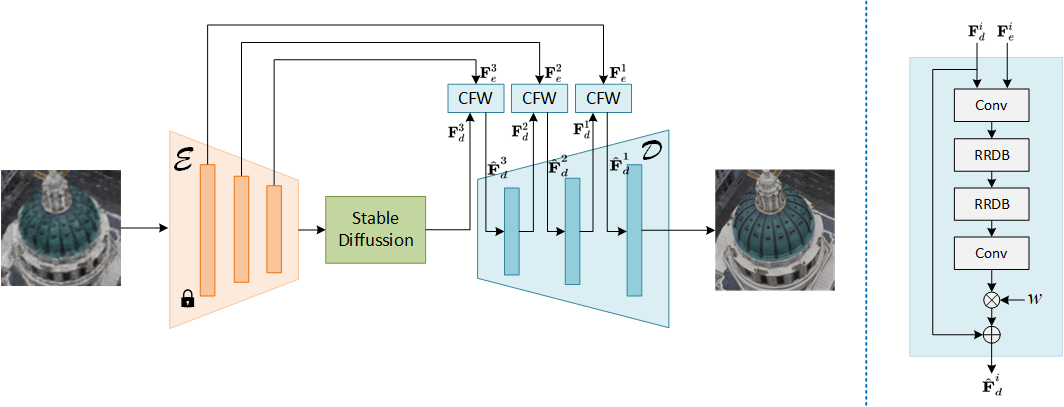}
  \caption{
   The detailed structure for controllable feature wrapping (CFW) module, which is depicted on the right side. We omit the inner structure of Stable Diffusion model for brevity. 
  }
  \label{fig:cfwmodule}
\end{figure}

As we mention in Section \ref{sec3.2}, controllable feature wrapping (CFW) modules are utilized to fuse the intermediate latent representations $\mathbf{F}_e$ and $\mathbf{F}_d$ from the VAE encoder $\mathcal{E}$ and the VAE decoder $\mathcal{D}$ respectively, which enhances the fidelity of the generation process. We illustrate the detailed structure for CFW module in Figure \ref{fig:cfwmodule}. As shown in Figure \ref{fig:cfwmodule}, CFW contains two convolution layers and two Residual in Residual Dense Block(RRDB) layers~\cite{esrgan}. The controllable coefficient $w \in [0,1]$ controls how much information from the low-quality aliased inputs is integrated to the generated anti-aliasing high-quality outputs, while in our experiments we set $w=1$ during both training and inference for better fidelity, thus better multi-view consistency.

\subsection{Patch Aggregation Reverse Sampling Procedure}
\label{patch reverse sampling pseducode}
In this section, we provide a detailed elucidation on the process of patch aggregation reverse sampling, which corresponds to Section \ref{sec3.3}.

We denote the aliased low-quality image rendered by Neural Radiance Field as $x_{lq} \in \mathcal{R}^{h \times w}$ and the corresponding latent feature map after encoded by VAE encoder $\mathcal{E}$ as $\boldsymbol{F} \in \mathcal{R}^{(h/8)\times(w/8)}$. Accordingly, $\boldsymbol{Z}^{(t)} \in \mathcal{R}^{(h/8)\times(w/8)}$ represents the intermediate latent feature map at timestep $t$. We denoise recursively step-by-step until obtain $\boldsymbol{Z}^{(0)}$, which is finally decoded by VAE decoder $\mathcal{D}$ to generate a high-quality anti-aliasing rendering $x_{hq} \in \mathcal{R}^{h \times w}$.

As mentioned in Section \ref{sec3.3} in the main paper, we divide $\boldsymbol{F}$ and $\boldsymbol{Z}^{(t)}$ into $M$ overlapping patches, dubbed as ${\{\boldsymbol{F}_{\Omega_n}}\}_{n=1}^M$ and ${\{\boldsymbol{Z}^{(t)}_{\Omega_n}}\}_{n=1}^M$ respectively, each with a resolution of $64 \times 64$. Here $\Omega_n$ is the coordinate set of the $nth$ patch. During each timestep $t$ in the reverse sampling, individual patches ($\boldsymbol{F}_{\Omega_n}$ and $\boldsymbol{Z}^{(t)}_{\Omega_n}$) are processed through the diffusion model $\epsilon_{\theta}$ independently, with the processed patches $\epsilon_{\theta}(\boldsymbol{Z}_{\Omega_n}^{(t)}, \boldsymbol{F}_{\Omega_n}, t) \in \mathcal{R}^{64 \times 64}$ subsequently aggregated. We define a set of Gaussian filters ${\{\boldsymbol{\omega}_{\Omega_n}}\}_{n=1}^M$, while each $\boldsymbol{\omega}_{\Omega_n} \in \mathcal{R}^{(h/8)\times(w/8)}$ follows up a Gaussian filter format in the region of $\Omega_n$ and equals to zero elsewhere. We also define a padding function $f(\cdot)$ that expands any patch of size $64 \times 64$ to the resolution of $(h/8)\times(w/8)$ by filling zeros outside the region $\Omega_n$. In this way, we can derive the expression for full-resolution noise-prediction item $\epsilon_{\theta}(\boldsymbol{Z}^{(t)}, \boldsymbol{F}, t) \in \mathcal{R}^{(h/8)\times(w/8)}$ as follows:
\begin{equation}
    \epsilon_{\theta}(\boldsymbol{Z}^{(t)}, \boldsymbol{F}, t)
    = \sum\limits_{n=1}^{M} \frac{\boldsymbol{\omega}_{\Omega_n}}{\hat{\boldsymbol{\omega}}} \odot
    f(\epsilon_{\theta}(\boldsymbol{Z}_{\Omega_n}^{(t)}, \boldsymbol{F}_{\Omega_n}, t))
    \label{epison_compute}
\end{equation}
where $\hat{\boldsymbol{\omega}}=\sum_n \boldsymbol{\omega}_{\Omega_n}$. Given $\epsilon_{\theta}(\boldsymbol{Z}^{(t)}, \boldsymbol{F}, t)$, we are able to obtain $\boldsymbol{Z}^{(t-1)}$ according to corresponding sampling procedure~\cite{ddim, ddpm}, which are denoted as Sampler$()$ function in Algorithm \ref{pseudocode}. We further re-split $\boldsymbol{Z}^{(t-1)}$ and repeat the above process until we obtain $\boldsymbol{Z}^{(0)}$. The overall algorithm for patch aggregation reverse sampling is located in Algorithm \ref{pseudocode}.

\begin{algorithm}  
\caption{Patch Aggregation Reverse Sampling}  
\begin{algorithmic}[1] 
\Require Cropped Regions ${\{\Omega_n}\}_{n=1}^M$, diffusion steps $T$, low-quality aliased feature map $\boldsymbol{F}$.

\State Initialize $\boldsymbol{\omega}_{\Omega_n}$ and $\hat{\boldsymbol{\omega}}$  
\State $\boldsymbol{Z}^{(T)} \sim \mathcal{N}(0, \mathbb{I})$

\For{$t \in [T, \cdots, 0]$}  
   \For{$n \in [1, \cdots, M]$}    
        \State Compute $\epsilon_{\theta}(\boldsymbol{Z}_{\Omega_n}^{(t)}, \boldsymbol{F}_{\Omega_n}, t)$
\EndFor  

    \State Compute $\epsilon_{\theta}(\boldsymbol{Z}^{(t)}, \boldsymbol{F}, t)$ following Equation (\ref{epison_compute})
    \State $\boldsymbol{Z}^{(t-1)}$ = Sampler$(\boldsymbol{Z}^{(t)}, \epsilon_{\theta}(\boldsymbol{Z}^{(t)}, \boldsymbol{F}, t))$
\EndFor
\State \Return $\boldsymbol{Z}^{(0)}$  
\end{algorithmic}  
\label{pseudocode}
\end{algorithm}

\subsection{Additional Details}
\label{additional training details}
As pointed out in Equation \ref{equation1} in the main paper, along with traditional reconstruction loss, training a Neural Radiance Field perfectly usually requires some additional regularization terms $\mathcal{L}_{reg}$ to avoid overfitting and local minima. As we train GridNeRF~\cite{gridnerf}, we adopt the standard regularization
terms that are commonly used in compressive sensing, including a $L1$ sparsity loss and a TV (total variation) loss on the decomposed vectors and matrix factors, expressed in Equation \ref{regloss:gridnerf}. 

\begin{equation}
    \mathcal{L}_{reg} = \mathcal{L}_{L1} + \mathcal{L}_{TV}
    \label{regloss:gridnerf}
\end{equation}

As we optimize ZipNeRF~\cite{zipnerf}, we use the anti-aliased interlevel loss to supervise the learning of the proposal branch network and anti $z$-aliasing, expressed in Equation \ref{regloss:zipnerf}.
\begin{equation}
    \mathcal{L}_{reg} = \mathcal{L}_{prop} 
    \label{regloss:zipnerf}
\end{equation}
Please refer to TensoRF~\cite{tensorf} and ZipNeRF~\cite{zipnerf} for the precise expressions of the loss terms in Equation \ref{regloss:gridnerf} and Equation \ref{regloss:zipnerf}, respectively.

The stage-one training for GridNeRF+\ours~and ZipNeRF+\ours~is conducted on one 80G A100 GPU, which takes approximately 12 hours. The stage-two training is conducted on four 32G V100 GPUs, which takes around 30 hours. 

Our proposed method is built on Stable Diffusion~\cite{stablediff} with resolution $512 \times 512$, however it is impractical to render $512 \times 512 = 262144$ pixels every training iteration due to the limited memory. Instead we render a sub-patch with smaller resolution at each iteration and piece them together every $n$ iterations to form a $512 \times 512$ large patch. Specifically, for GridNeRF~\cite{gridnerf} we render a $64 \times 128$ patch each iteration so $n=32$. For ZipNeRF~\cite{zipnerf} we render a $256 \times 256$ patch to form a large patch every $n=4$ iterations.

\clearpage
\begin{figure*}[t]
	\small
	\centering

       \begin{minipage}[t]{0.25\linewidth}
		\centering
		\centerline{\includegraphics[width=0.99\linewidth]{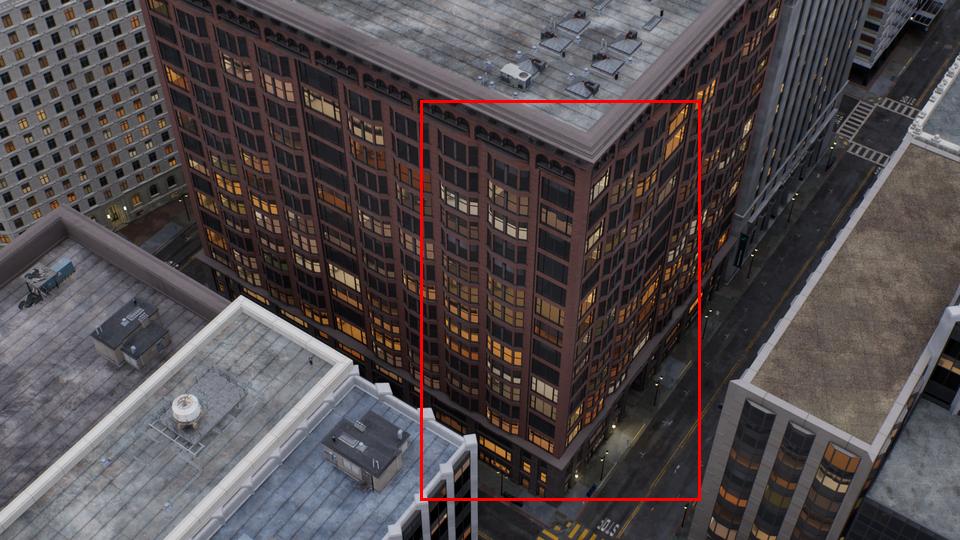}}
		\vspace{-0.3mm}
	\end{minipage}
	\begin{minipage}[t]{0.18\linewidth}
		\centering
		\centerline{\includegraphics[width=0.99\linewidth]{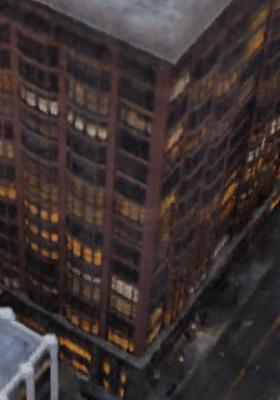}}
		\vspace{-0.3mm}
	\end{minipage}
	\begin{minipage}[t]{0.18\linewidth}
		\centering
		\centerline{\includegraphics[width=0.99\linewidth]{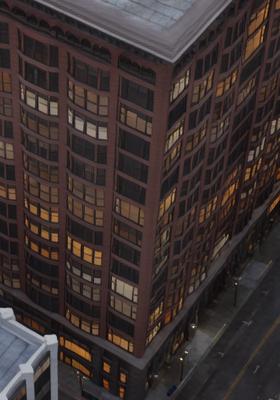}}
		\vspace{-0.3mm}
	\end{minipage}
    \begin{minipage}[t]{0.18\linewidth}
		\centering
		\centerline{\includegraphics[width=0.99\linewidth]{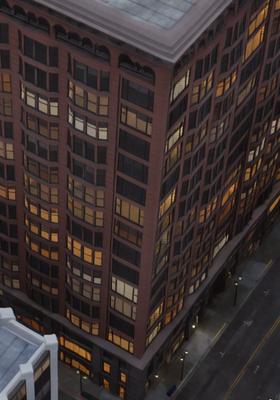}}
		\vspace{-0.3mm}
	\end{minipage}
	\begin{minipage}[t]{0.18\linewidth}
		\centering
		\centerline{\includegraphics[width=0.99\linewidth]{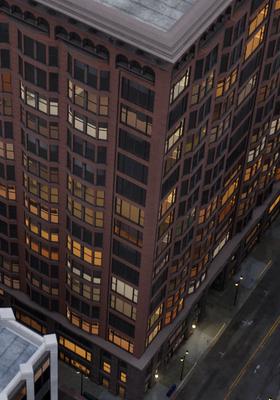}}
		\vspace{-0.3mm}
	\end{minipage}

   \begin{minipage}[t]{0.25\linewidth}
		\centering
		\centerline{\includegraphics[width=0.99\linewidth]{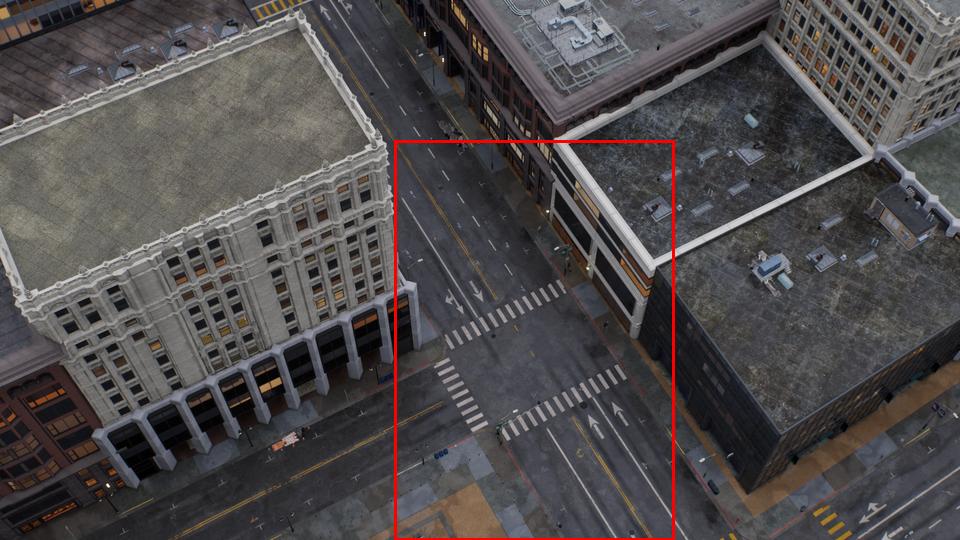}}
		\vspace{-0.3mm}
	\end{minipage}
	\begin{minipage}[t]{0.18\linewidth}
		\centering
		\centerline{\includegraphics[width=0.99\linewidth]{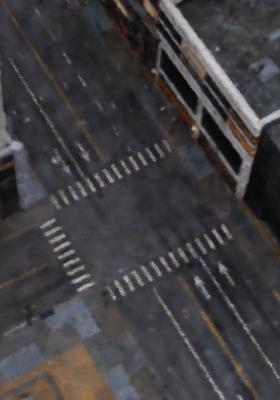}}
		\vspace{-0.3mm}
	\end{minipage}
	\begin{minipage}[t]{0.18\linewidth}
		\centering
		\centerline{\includegraphics[width=0.99\linewidth]{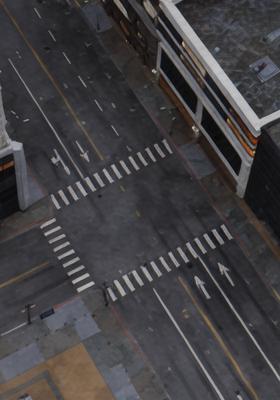}}
		\vspace{-0.3mm}
	\end{minipage}
    \begin{minipage}[t]{0.18\linewidth}
		\centering
		\centerline{\includegraphics[width=0.99\linewidth]{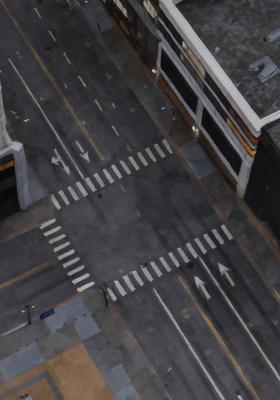}}
		\vspace{-0.3mm}
	\end{minipage}
	\begin{minipage}[t]{0.18\linewidth}
		\centering
		\centerline{\includegraphics[width=0.99\linewidth]{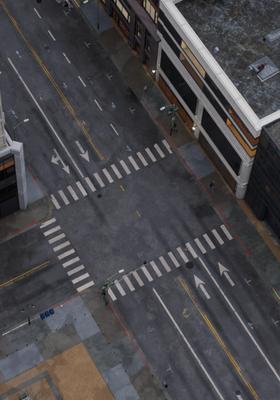}}
		\vspace{-0.3mm}
	\end{minipage}

    \begin{minipage}[t]{0.25\linewidth}
		\centering
		\centerline{\includegraphics[width=0.99\linewidth]{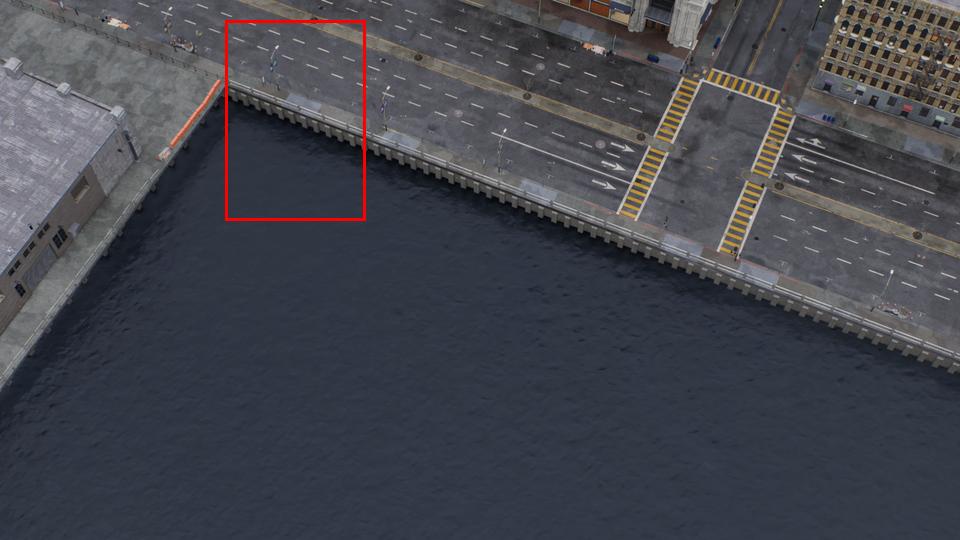}}
		\vspace{-0.3mm}
       \centerline{Full Image}
	\end{minipage}
	\begin{minipage}[t]{0.18\linewidth}
		\centering
		\centerline{\includegraphics[width=0.99\linewidth]{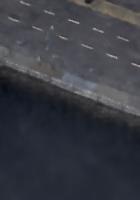}}
		\vspace{-0.3mm}
        \centerline{GridNeRF}
	\end{minipage}
	\begin{minipage}[t]{0.18\linewidth}
		\centering
		\centerline{\includegraphics[width=0.99\linewidth]{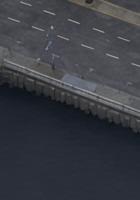}}
		\vspace{-0.3mm}
        \centerline{GridNeRF}
        \centerline{+\ours$_{gen}$}
	\end{minipage}
    \begin{minipage}[t]{0.18\linewidth}
		\centering
		\centerline{\includegraphics[width=0.99\linewidth]{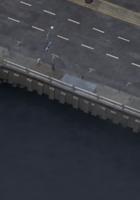}}
		\vspace{-0.3mm}
        \centerline{GridNeRF}
        \centerline{+\ours$_{spe}$}
	\end{minipage}
	\begin{minipage}[t]{0.18\linewidth}
		\centering
		\centerline{\includegraphics[width=0.99\linewidth]{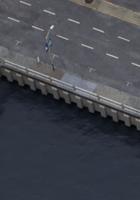}}
		\vspace{-0.3mm}
        \centerline{G.T.}
	\end{minipage}
	
	\vspace{-1mm}
	\caption{Visual comparisons of adding cross-scene \textbf{gen}eralizable model \ours$_{gen}$ and per-scene \textbf{spe}cific model \ours$_{spe}$ upon GridNeRF~\cite{gridnerf}.}
	\label{fig:genvsspe}
	\vspace{-6mm}
\end{figure*}

\clearpage

\clearpage

\clearpage

\clearpage

\end{document}